%% file: 0_main.tex
\pdfoutput=1

\documentclass[11pt]{article}

\usepackage{acl}

\usepackage{times}
\usepackage{latexsym}

\usepackage[T1]{fontenc}

\usepackage[utf8]{inputenc}

\usepackage{microtype}

\usepackage{inconsolata}
\usepackage{multirow}
\usepackage{arydshln} 
\usepackage{color, soul, colortbl}
\usepackage{graphicx}
\usepackage{booktabs}
\usepackage[most]{tcolorbox}
\usepackage[export]{adjustbox} 
\usepackage{pifont}
\usepackage{booktabs}
\usepackage{multirow}
\usepackage{subcaption}
\usepackage{graphicx}
\usepackage{tikz}
\usetikzlibrary{shapes.geometric, arrows}
\usetikzlibrary{decorations.markings}
\usepackage{soul}
\usepackage{wrapfig,graphicx,lipsum}
\usepackage[T1]{fontenc}
\usepackage[utf8]{inputenc}
\usepackage{lmodern}
\usepackage{extsizes}
\usepackage{mathptmx}
\usepackage{cuted}
\usepackage{flushend}
\usepackage{float}
\usepackage{multicol}
\usepackage{anyfontsize}
\usepackage[draft,textsize=footnotesize,textwidth=15mm]{todonotes}
\usepackage{enumitem}
\setlist{leftmargin=5.5mm}
\usepackage{setspace}
\usepackage{cleveref}

\usepackage{algorithm}
\usepackage{algpseudocode}
\usepackage{color}

\usepackage{siunitx,tabularx,ragged2e,booktabs,caption}

\makeatletter
\newcommand{\DrawLine}{%
  \begin{tikzpicture}
  \path[use as bounding box] (0,0) -- (\linewidth,0);
  \draw[color=blue!75!black,dashed,dash phase=.5pt]
        (0-\kvtcb@leftlower-\kvtcb@boxsep,0)--
        (\linewidth+\kvtcb@rightlower+\kvtcb@boxsep,0);
  \end{tikzpicture}%
  }
\makeatother

\usepackage[draft,textsize=footnotesize,textwidth=15mm]{todonotes}
\usepackage{color}
\usepackage{soul}

\title{
\vspace{-20mm}
\begin{center}
\vspace{-3mm}
\includegraphics[width=0.75\textwidth]{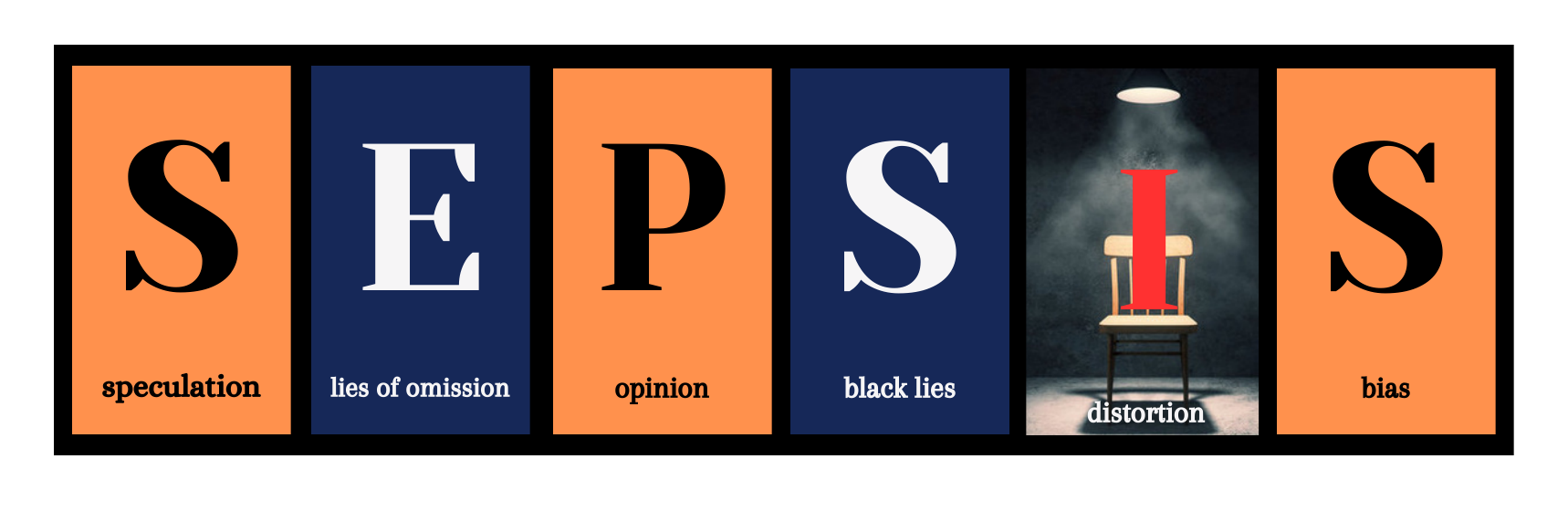}
\vspace{-5mm}
\end{center}
SEPSIS: I Can Catch Your Lies -- A New Paradigm for Deception Detection
}

\author{\textbf{Anku Rani}$^{1}$\thanks{\,\,\, Work was done when the author was at the University of South Carolina} \quad \textbf{Dwip Dalal}$^{2}$ \quad \textbf{Shreya Gautam}$^{3}$ \quad \textbf{Pankaj Gupta}$^{4}$ \\
\textbf{Vinija Jain}\dag\textsuperscript{5,6} \quad
\textbf{Aman Chadha}\dag\textsuperscript{5,6} \quad \textbf{Amit Sheth}$^{7}$ \quad \textbf{Amitava Das}$^{7}$ \quad \\
$^{1}$ Massachusetts Institute of Technology \quad     
$^{2}$IIT Gandhinagar, India \quad \\
$^{3}$Politecnico di Milano, Italy \quad
$^{4}$DTU, India \quad 
$^{5}$Stanford University, USA \\ 
$^{6}$Amazon AI, USA \quad
$^{7}$University of South Carolina, USA \\
\tt  
ankurani@mit.edu
}

\begin{document}
\maketitle
\renewcommand{\thefootnote}{\fnsymbol{footnote}}
\footnotetext[2]{Work does not relate to position at Amazon.}
\renewcommand*{\thefootnote}{\arabic{footnote}}
\setcounter{footnote}{0}
\begin{abstract}

Deception is the intentional practice of twisting information. It is a nuanced societal practice deeply intertwined with human societal evolution, characterized by a multitude of facets. This research explores the problem of deception through the lens of psychology, employing a framework that categorizes deception into three forms: \emph{lies of omission}, \emph{lies of commission}, and \emph{lies of influence.} The primary focus of this study is specifically on investigating only \emph{lies of omission.} We propose a novel framework for deception detection leveraging NLP techniques. We curated an annotated dataset of $876,784$ samples by amalgamating a popular large-scale fake news dataset and scraped news headlines from the Twitter handle of "\emph{Times of India}", a well-known Indian news media house. Each sample has been labeled with four layers, namely: (i) the \ul{type of omission} \textit{(speculation, bias, distortion, sounds factual,} and \textit{opinion)}, (ii) \ul{colors of lies} \textit{(black, white, grey,} and \textit{red)}, and (iii) the \ul{intention of such lies} \textit{(to influence, gain social prestige,} etc\textit{)} (iv) \ul{topic of lies} \textit{(political, educational, religious, racial,} and \textit{ethnicity)}. We present a novel multi-task learning [MTL] pipeline that leverages the dataless merging of fine-tuned language models to address the deception detection task mentioned earlier. Our proposed model achieved an impressive F1 score of 0.87, demonstrating strong performance across all layers, including the \textit{type}, \textit{color}, \textit{intent}, and \textit{topic} aspects of deceptive content.
Finally, our research aims to explore the relationship between \emph{lies of omission} and \emph{propaganda} techniques. To accomplish this, we conducted an in-depth analysis, uncovering compelling findings. For instance, our analysis revealed a significant correlation between \emph{loaded language} and \emph{opinion}, shedding light on their interconnectedness. 
To encourage further research in this field, we are releasing the SEPSIS dataset and code at \url{https://huggingface.co/datasets/ankurani/deception}.

\end{abstract}

\input{1_introduction}

\input{3_data}

\input{4_PromptEngineering}
\input{5_MultitaskLearning}

\input{6_PropagandaTheory}

\input{2_related_works}
\input{7_conclusion}

\bibliography{custom}

\input{9_faq}
\input{10_appendix}

\end{document}

%% file: 1_introduction.tex
\vspace{-6mm}
\section{Defining Deception -- Inspiration from Psychology}

According to \cite{schuiling2004deceive}, deception is a behavior observed in various species and is considered an evolutionary adaptive trait. \cite{depaulo1998everyday} asserts that deception is an integral part of social interactions, with the \ul{majority of humans engaging in deceptive acts at least once or twice a day}. While most instances of deception are relatively minor, there is a frequent association between deception and egregious norm violations, such as theft, murder, and attempts to evade punishment for such crimes. Consequently, researchers have long been interested in identifying behaviors that can differentiate between truthful and deceitful communications.

Numerous studies have delved into describing the behavioral indicators of deceit. However, no single behavior or combination of behaviors has been found to possess the definitive ability to accurately determine deceptive communication. 
The empirical evidence supporting the significance of specific individual behaviors in deception often presents conflicting findings \cite{depaulo1985deceiving, kraut1980humans, vrij2000detecting}. One possible explanation for these contradictions in the literature regarding deception cues is the insufficient differentiation made by researchers between distinct subtypes of deception.

\vspace{-1mm}
In the realm of psychology research, a consensus has yet to be reached regarding the classification of various types of deception. Nevertheless, we discovered that the framework outlined in Hample's work \cite{hample1982empirical}, visually described in figure ~\ref{fig:sepsis}, provides a viable foundation for constructing NLP models. 
Hample, et al, 1982 categorize deception into three distinct forms: \emph{lies of omission}, \emph{lies of commission}, and \emph{lies of influence}. For the purpose of our study, we focus solely on investigating \emph{lies of omission}. It is worth noting that the NLP community has extensively explored the fact verification problem, which is primarily associated with \emph{lies of commission}. Conversely, \emph{lies of omission} have received comparatively less attention. In this paper, we present a comprehensive study on lies of omission, which, to the best of our knowledge, is the first of its kind. 

\vspace{-2mm}

\begin{tcolorbox}[colback=blue!5!white,colframe=blue!75!black,title=\footnotesize{\textbf{\textsc{\ul{Our Contributions}}}: \textls[-10]{SEPSIS dataset, MTL framework, unveiling the relationship between deception and propaganda}.}]
\begin{itemize}
[leftmargin=1mm]
\setlength\itemsep{0em}
\begin{spacing}{0.90}
\vspace{-2mm}
\item[\ding{224}] {\footnotesize \fontfamily{phv}\fontsize{8}{9}\selectfont{ 

We present a pioneering study on the phenomenon of lies of omission.
}}
\vspace{-1mm}
\item[\ding{224}] {\footnotesize 
{\fontfamily{phv}\fontsize{8}{9}\selectfont{

We introduce the SEPSIS corpus and associated resources. The SEPSIS corpus (876,784 data points) incorporates four layers of annotation.}
}}

\vspace{-1mm}
\item[\ding{224}] {\footnotesize 
{\fontfamily{phv}\fontsize{8}{9}\selectfont
{We introduce an MTL pipeline for SEPSIS classification. The MTL pipeline leverages the dataless merging of fine-tuned Language Models (LMs) and further incorporates a tailored loss function specific to each layer, addressing different sub-problems.}
}}

\vspace{-1mm}
\item[\ding{224}] {\footnotesize 
{\fontfamily{phv}\fontsize{8}{9}\selectfont
{Finally, the paper reveals a significant correlation between deception and propaganda techniques.}
}}

\vspace{-6mm}
\end{spacing}
\end{itemize}
\end{tcolorbox}
\vspace{-3mm}

%% file: 3_data.tex
\begin{figure}[!tbh]
\centering
\includegraphics[width=\columnwidth, height=3.75cm]{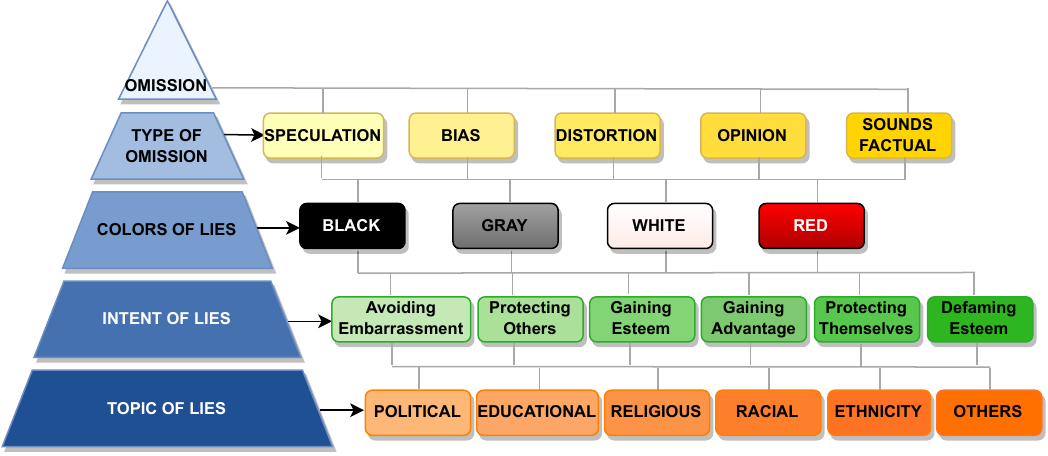}
\vspace{-5mm}
\caption{The figure represents the categorization of the SEPSIS corpus across all layers. The 1$^{st}$ layer represents \textit{type of omission} and its respective categories, 2$^{nd}$ layer represents colors of lies, 3$^{rd}$ layer represents the intent of lies, and 4$^{th}$ layer represents the topic of lies. }

\label{fig:sepsis}
\vspace{-4.5mm}
\end{figure}

\section{Introducing SEPSIS: A novel corpus on \ul{lies of omission}}
\label{sec:introduction}
\vspace{-1mm}
We are delighted to introduce the \textbf{SEPSIS} corpus (\textbf{S}p\textbf{E}culation o\textbf{P}inion bia\textbf{S} d\textbf{I}\textbf{S}tortion), explicitly curated for \emph{lies of omission}. This novel resource will significantly enhance the study and analysis of deceptive communication by focusing on the deliberate exclusion of information. Figure \ref{fig:sepsis} offers a concise visual depiction that effectively summarizes the categorization we present in the SEPSIS. 

In the subsequent paragraphs, we present a collection of scientific inquiries along with their corresponding answers, which serve as the driving force behind our research. Furthermore, we delve into the influence of these questions on the development of our annotation schema, which lays the groundwork for our research framework.
\vspace{-1mm}

\vspace{2mm}
\noindent
\textls[-5]{\textbf{\ul{Is there a specific dialogue act that individuals employ for lies of omission?}} Within the classical switchboard corpus \cite{godfrey1992switchboard}, 
there exist 42 well-defined dialogue acts. Following extensive deliberation and analysis, we have reached the conclusion that individuals often utilize dialogue acts such as \emph{speculation, opinion, bias, and distortions} when engaging in deceptive behavior. 

These dialogue acts function as figurative communication techniques employed by individuals to mask their deceit through encryption \cite{elaad2003effects}, particularly when they desire to disclose certain information selectively.}

\vspace{-3mm}
\begin{itemize}
[leftmargin=1mm]
\setlength\itemsep{0em}
\fontsize{9}{10}\selectfont{\item \textbf{Speculation} entails conjecturing without ample evidence.

\vspace{-1mm}
\item\textbf{Opinion} is a subjective viewpoint formed without relying on factually accepted knowledge.

\vspace{-1mm}
\item\textbf{Bias} refers to unfair prejudice towards a particular individual or group. 
\vspace{-1mm}
\item\textbf{Distortion} is the act of twisting something away from its genuine, inherent, or initial condition.

\vspace{-1mm}
\item Lastly, we define \textbf{sounds factual} as a statement that seems factual but may not be true.}

\end{itemize}

\vspace{-6mm}
\begin{tcolorbox}[enhanced,attach boxed title to top right={yshift=-3mm,yshifttext=-1mm},
  colback=blue!5!white,colframe=blue!75!black,colbacktitle=red!80!black,
  title=$1^{st}$ level: type of omission,fonttitle=\bfseries,
  boxed title style={size=small,colframe=red!50!black},left=0pt, right=0pt]

  \begin{spacing}{0.8}
  \textbf{\ul{\footnotesize Speculation:}} {\fontfamily{lmss}\fontsize{8}{10}\selectfont{Biden warned the US does not have 'resources to win WW3' as tensions rise in the Middle East.}}
  \end{spacing}

  \vspace{-2.5mm}
  \DrawLine
  
  \begin{spacing}{0.8}
  \textbf{\ul{\footnotesize Opinion:} }{\fontfamily{lmss}\fontsize{8}{10}\selectfont
 Poll: Trump receives low overall approval rating but praise for strong economy.}\end{spacing}
  \vspace{-2.5mm}
  \DrawLine
  \begin{spacing}{0.8}
  \textbf{\ul{\footnotesize Bias:} }{
  \fontfamily{lmss}\fontsize{8}{10}\selectfont
  Russia lauds India for following own interests on energy issue.}
  \end{spacing} 

  \vspace{-2.5mm}
  \DrawLine
  
  \begin{spacing}{0.8}
  \textbf{\ul{\footnotesize Distortion:} }{\fontfamily{lmss}\fontsize{8}{10}\selectfont
  Republic TV: Jama Masjid in dark due to non-payment of electricity bills over four crores.}
  \end{spacing}

  \vspace{-2.5mm}
  \DrawLine
  
  \begin{spacing}{0.8}
  \textbf{\ul{\footnotesize Sounds\, Factual:} }{\fontfamily{lmss}\fontsize{8}{10}\selectfont
  A US government study confirms most face recognition systems are racist.}\end{spacing} 
  \vspace{-0.75mm}
\end{tcolorbox}

\vspace{-1mm}
\noindent
\textbf{\ul{What has been omitted?}} In the study of lies of omission, it is crucial to determine what information has been deliberately omitted. To address this, we draw inspiration from journalism, where the use of the 5W framework is common. The 5W framework consists of the questions \textit{who, what, when, where, and why} which are considered fundamental in information gathering and problem-solving. These questions are frequently utilized in journalism and police investigations \cite{10.2307/1023893, stofer2009sports, silverman, su2019study, smarts_2017,article_2023}. As an example:

\vspace{-1.5mm}  
\begin{tcolorbox}[left=0pt, right=0pt]
\small
\vspace{-1mm}
\footnotesize{
\{Hillary Clinton\}\textsubscript{\textcolor{blue}{who\textsubscript{1}}} announces \{Global Climate Resilience Fund\}\textsubscript{\textcolor{blue}{what}} for \{women\}\textsubscript{\textcolor{blue}{who\textsubscript{2}}}  
to\{tackle climate change\}\textsubscript{\textcolor{blue}{why}}}
\vspace{-3mm}
\end{tcolorbox} 

\noindent
\textbf{\ul{What is the vulnerability of the uttered lie?}} In the realm of deception research, it is of utmost importance to comprehend and quantify the susceptibility of lies. One approach involves categorizing lies into different colors, namely \emph{black, red, white, and gray} \cite{ratliff2011behavioral,depaulo2004many}. Each color represents a distinct type of lie with varying levels of vulnerability, as detailed below:

\vspace{-2mm}
\begin{itemize}
[leftmargin=1mm]
\setlength\itemsep{0em}
\fontsize{9}{10}\selectfont{\item \textbf{Black lie} is about simple and callous selfishness. Typically uttered when there is no benefit to others, its sole intention is to extricate oneself from trouble.}
\vspace{-1.5mm}
\item \textbf{White lie} prioritizes others' welfare over personal interests, reflecting an altruistic nature.

\vspace{-1mm}
\item\textbf{Gray lies} exhibits dual behavior, partially benefiting others and partially benefiting oneself, depending on the viewpoint. 

\vspace{-1mm}
\item\textbf{Red lies} are spoken from a hatred and revenge perspective against individuals or groups. 

\end{itemize}

\vspace{-7mm}
\begin{tcolorbox}[enhanced,attach boxed title to top right={yshift=-3mm,yshifttext=-1mm},
  colback=blue!5!white,colframe=blue!75!black,colbacktitle=red!80!black,
  title=$2^{nd}$ level: colors of lie,fonttitle=\bfseries,
  boxed title style={size=small,colframe=red!50!black},left=0pt, right=0pt ]

  \begin{spacing}{0.8}\textbf{\ul{\footnotesize Red:} }{\fontfamily{lmss}\fontsize{8}{10}\selectfont
  Donald Trump's congratulatory post for North Korea's WHO membership sparks outrage and controversy.}
  \end{spacing} 

  \vspace{-2.5mm}
  \DrawLine

  \begin{spacing}{0.8}
  \textbf{\ul{\footnotesize Black:} }{\fontfamily{lmss}\fontsize{8}{10}\selectfont
FTX collapse: Former CEO Sam Bankman-Fried urges court to toss charges.}
  \end{spacing}

  \vspace{-2.5mm}
  \DrawLine

  \begin{spacing}{0.8}
  \textbf{\ul{\footnotesize White:} }{\fontfamily{lmss}\fontsize{8}{10}\selectfont 

An apple a day slashes frailty risk by 20 percent, but a Study points otherwise.
}
  \end{spacing} 

  \vspace{-2.5mm}
  \DrawLine
  
  \begin{spacing}{0.8}
  \textbf{\ul{\footnotesize Gray:} }{\fontfamily{lmss}\fontsize{8}{10}\selectfont
Hillary Clinton Announces Global Climate Resilience Fund For Women To Tackle Climate Change.}
  \end{spacing}
  
  \vspace{-1mm}
\end{tcolorbox}

\input{table/Sepsis_corpus}

\vspace{-2mm}
\noindent
\textbf{\ul{What is the intent of the lie?}}
 
Studying the intent of lies helps to comprehend the objective of deceptive language. We have thus categorized lies into different intents as shown below.

\vspace{-3mm}
\begin{tcolorbox}[enhanced,attach boxed title to top right={yshift=-3mm,yshifttext=-1mm},
  colback=blue!5!white,colframe=blue!75!black,colbacktitle=red!80!black,
  title=$3^{rd}$ level: intent of lie,fonttitle=\bfseries,
  boxed title style={size=small,colframe=red!50!black},left=0pt, right=0pt ]
  
\begin{spacing}{0.8}
\textbf{\ul{\footnotesize Gaining\,Advantage:} }{\fontfamily{lmss}\fontsize{8}{10}\selectfont
  Elizabeth Holmes ordered dinners for Theranos staff but made sure they weren't delivered until after 8 p.m. so they worked late: book.
  }
\end{spacing}  

  \vspace{-2.5mm}
  \DrawLine

  \begin{spacing}{0.8}
  \textbf{\ul{\footnotesize Protecting\,Themselves:} }{\fontfamily{lmss}\fontsize{8}{10}\selectfont
ChatGPT creator Sam Altman testifies to US Congress on AI risks.}
  \end{spacing} 

  \vspace{-2.5mm}
  \DrawLine

  \begin{spacing}{0.8}
  \textbf{\ul{\footnotesize Avoiding\,Embarrassment:} }{\fontfamily{lmss}\fontsize{8}{10}\selectfont 
Trump’s Suggestion That Disinfectants Could Be Used to Treat Coronavirus Prompts Aggressive Pushback, was Sarcastic?}
  \end{spacing} 

  \vspace{-2.5mm}
  \DrawLine

  \begin{spacing}{0.8}
  \textbf{\ul{\footnotesize Gaining\,Esteem:} }{\fontfamily{lmss}\fontsize{8}{10}\selectfont
Sasan Goodarzi, the CEO of software giant Intuit, which has avoided mass layoffs, says tech firms axed jobs because they misread the pandemic.}
  \end{spacing} 

  \vspace{-2.5mm}
  \DrawLine

  \begin{spacing}{0.8}
 \textbf{\ul{\footnotesize Protecting\,Others:} }{\fontfamily{lmss}\fontsize{8}{10}\selectfont 
Nobel Laureate Malala Urges U.S. To Bolster Support For Afghan Girls, Women!}
  \end{spacing}
  
  \vspace{-2.5mm}
  \DrawLine

\begin{spacing}{0.8}
 \textbf{\ul{\footnotesize Defaming\,Esteem:} }{\fontfamily{lmss}\fontsize{8}{10}\selectfont 
Taiwan war would be ‘devastating,’ warns US Defense Secretary Lloyd Austin as he criticizes China at Shangri-La security summit.}
  \end{spacing}
  \vspace{-1mm}
\end{tcolorbox}

\vspace{-3mm}

\vspace{-1.5mm}
\begin{itemize}
[leftmargin=1mm]
\setlength\itemsep{0em}
\fontsize{9}{10}\selectfont{\item\textbf{Intent of Gaining Advantage} can be used as an act of intentionally providing false information or misleading others to gain an unfair advantage over them. 

\vspace{-1mm}
\item\textbf{Intent of Protecting Themselves} can be used as a means of self-preservation or self-defense when an individual feels threatened or vulnerable.

\vspace{-1 mm}
\item\textbf{Intent of Avoiding Embarrassment} can be employed to evade situations that may lead to embarrassment, humiliation, or social discomfort.
\vspace{-1mm}

\item\textbf{Intent of Gaining Esteem} can be utilized to enhance one's reputation, social status, or personal image.


\item\textbf{Intent of Protecting Others} can be used as a means of preservation for others when a group or community feels threatened or vulnerable.
\item\textbf{Intent of Defaming Esteem} intends to damage reputation by spreading false information or rumors.}

\end{itemize}

\vspace{-2mm}

\noindent
\textbf{\ul{What is the topic of lie?}}
To study deception further and to understand its topical influence, this research categorizes different topics of lies such as political, educational, etc.

\vspace{-3mm}
\begin{itemize}
[leftmargin=1mm]
\setlength\itemsep{0em}

\fontsize{9}{10}\selectfont{\item\textbf{Political} deception occurs by the deliberate use of statements by political entities to manipulate public opinion.

\vspace{-1mm}
\item\fontsize{9}{10}\selectfont{\textbf{Educational} deception occurs by the deliberate use of statements by academic entities to manipulate opinion, directed especially towards the younger population.
\vspace{-1mm}
\item\textbf{Racial} deception occurs when individuals intentionally misrepresent their racial identity or engage in deception driven by racial motives.
\vspace{-1mm}
\item\textbf{Religious} deception involves the act of deceiving others by misrepresenting one's religious beliefs.}

\vspace{-1mm}
\item\textbf{Ethnic} deception refers to the act of intentionally manipulating one's ethnic identity by targeting specific ethnic groups.}
\end{itemize}

\vspace{-6mm}
\begin{tcolorbox}[enhanced,attach boxed title to top right={yshift=-3mm,yshifttext=-1mm},
  colback=blue!5!white,colframe=blue!75!black,colbacktitle=red!80!black,
  title=$4^{th}$ level: topic of lie,fonttitle=\bfseries,
  boxed title style={size=small,colframe=red!50!black},left=0pt, right=0pt ]
  
\begin{spacing}{0.8}
  \textbf{\ul{\footnotesize Political:} }{\fontfamily{lmss}\fontsize{8}{10}\selectfont
 No elections safe from AI, deep fake photos, videos of politicians to become common, warns former Google boss.}\end{spacing}

  \vspace{-2.5mm}
  \DrawLine

  \begin{spacing}{0.8}
  \textbf{\ul{\footnotesize Educational:} }{\fontfamily{lmss}\fontsize{8}{10}\selectfont 
Hundreds gather at Florida school board meeting over Disney movie controversy: 'Your policies are not protecting us from anything.}\end{spacing} 

  \vspace{-2.5mm}
  \DrawLine

  \begin{spacing}{0.8}
  \textbf{\ul{\footnotesize Religious:} }{\fontfamily{lmss}\fontsize{8}{10}\selectfont
Pope: Christianity, Islam share common commitment to good life.}\end{spacing}

  \vspace{-2.5mm}
  \DrawLine

  \begin{spacing}{0.8}
  \textbf{\ul{\footnotesize Racial:} }{\fontfamily{lmss}\fontsize{8}{10}\selectfont
  Why shouldn’t a mixed-race actress play Egyptian queen Cleopatra?}\end{spacing}

  \vspace{-2.5mm}
  \DrawLine

  \begin{spacing}{0.8}
 \textbf{\ul{\footnotesize Ethnicity:} }{\fontfamily{lmss}\fontsize{8}{10}\selectfont 
Egyptians complain over Netflix depiction of Cleopatra as black.}\end{spacing}
  
  \vspace{-1mm}
  
\end{tcolorbox}
\vspace{-2mm}


\section{SEPSIS: Data Sources, Annotation, and Agreement}
\vspace{-1mm}  

At the outset, we engaged in the manual annotation of 5,100 sentences through four co-authors, employing four layers of deception. Subsequently, we applied data augmentation techniques as detailed in Section ~\ref{sec:data_augmentation}, culminating in a total of 8,76,784 data points.

\subsection{Data Sources}
\vspace{-0.5mm}  

\input{table/kappa_score_3d}
\textls[0]{In terms of data sources, we have identified two distinct categories of interest. The first category focuses on the presence of omissions in factual data, specifically news data. The second category examines the involvement of omissions in fake news data. To address these categories, we have selected data sources from two prominent outlets: (a) Times of India \cite{timesofindia} Twitter handle, the renowned news agency in India, and (b) Information Security and Object Technology (ISOT) fake news dataset \cite{ISOTFakeNewsDataset}. More information on these sources can be found in the appendix \ref{sec:data sources}. A detailed analysis of the SEPSIS corpus and the results can be found in Appendix \ref{sec: data analysis}.}

\vspace{-1mm}
\subsection{Data Annotation}
\vspace{-1mm}  
\textls[0]{We chose to leverage our four co-authors for annotation purposes, which provides a knowledgeable and reliable solution for annotating sensitive deception datasets, ensuring high-quality expert judgment throughout the process. To maintain annotation consistency, we implemented rigorous checks and measures throughout the entire annotation process. The dataset was annotated at the sentence level using a multi-class annotation approach, allowing each individual feature to be assigned multiple categories during the annotation process. For instance, a statement could be tagged as both speculative and sounding factual, recognizing the possibility for it to either be a verifiable fact or contain speculative elements that satisfy both possibilities. A comprehensive account of the overall annotation process is provided in Appendix \ref{sec: data cleaning}. Notably, during the initial layer of annotation, if a particular text appeared to be factual, we refrained from annotating the specific type, intent, and influence of the lie since it was treated as a fact.}

\subsection{Inter Annotator Agreement and Quality}
\label{sec:iaa_score}  

To ensure quality control in the co-author annotations, we performed cross-validation annotations on 1000 data points. This validation dataset was utilized to assess the consistency of annotations provided by individual co-authors. Based on this assessment, we established annotation guidelines and conducted calibration sessions among the co-author team. For the annotation task, each co-author contributed their expertise across all four layers of the annotation process.
We obtained four annotations per sentence and subsequently consolidated the data using an improved voting technique, as suggested in \cite{hovy-etal-2013-learning}, which has been empirically shown to outperform majority voting. To assess the level of agreement in the annotated corpus, we also calculated the Cohen Kappa score \cite{cohen1960coefficient}. Since there are multiple categories for a given sentence, we report class-wise agreement scores. The overall agreement score is presented in Table \ref{tab: Kappa score}. An overview of data points is presented in Table \ref{tab:SEPSIS_corpus}. To understand how features across these four layers are dependent on each other, we present six heatmaps in Appendix \ref{sec: data analysis}.

\begin{table}[!ht]
\centering
\vspace{-1mm}
\resizebox{0.95\columnwidth}{!}{%
\begin{tabular}{cccccl}
                                                                                                                                                                                                                                                                                                                                                                                                      \\ 
\toprule
\textbf{Data Source} & \multicolumn{1}{c}{\textbf{\begin{tabular}[c]{@{}c@{}}Sentences\end{tabular}}} & \multicolumn{1}{c}{\textbf{\begin{tabular}[c]{@{}c@{}}+  Paraphrasing\end{tabular}}} & \multicolumn{1}{c}{\textbf{\begin{tabular}[c]{@{}c@{}}+  Mask Infilling\end{tabular}} } \\ \toprule
\textbf{Tweets}       &       $2495$                                                                                       &                    $12475$                                                                                                          &      $389105$                                                                                                                            &                                                                                                                                                   \\
\textbf{Fake News}   &      $2605$                                                                                        &     $13025$                                                                                                                         &    $487829$                                                                                                                              &                                                                                                                                                   \\ \midrule
\textbf{Total}       &       $5100$                                                                                       &      $25500$                                                                                                                        &     $876784$                                                                                                                             &                                                                                                                                                   \\ \bottomrule
\end{tabular}
}

\vspace{-1mm}
\caption {Number of original sentences and augmented sentences using \textit{paraphrasing} and \textit{mask infilling}.}
\label{tab:SEPSIS_corpus}
\vspace{-5mm}
\end{table}

%% file: table/Sepsis_corpus.tex
%% file: table/kappa_score_3d.tex
\begin{table*}[ht]
\centering
\resizebox{\textwidth}{!}{%
\begin{tabular}{lccccccccccccccccccccccccc}
\toprule
                                        & \multicolumn{5}{c}{\textbf{Lies of omission}}                                                           & \multicolumn{4}{c}{\centering \textbf{Color of lies}}             & \multicolumn{6}{c}{\textbf{Intent of Lies}}        
                                        
                                        \\
                                        \toprule

                                        & \multicolumn{1}{p{1.2cm}}{\centering \textbf{Specula-}
                                        \textbf{tion}
                                        }                                        


                                        & \textbf{Bias}
                                        & 
                                        \multicolumn{1}{p{1cm}}{\centering \textbf{Distor-}
                                        \textbf{tion}
                                        }   
                                        
                                        & \multicolumn{1}{p{1.35cm}}{\centering \textbf{Opinion}} 
                                        & \multicolumn{1}{p{1.35cm}}{\centering \textbf{Sounds}
                                        \\ \textbf{Factual}}
                                        
                                        & 
                                        
                                        \textbf{Black} & \textbf{White} & \textbf{Grey} & \textbf{Red} 
                                        
                                        & 
                                        
                                        \multicolumn{1}{p{1.5cm}}{\centering \textbf{Gaining} \\ \textbf{Advantage}} & 
                                        \multicolumn{1}{p{1.6cm}}{\centering \textbf{Protecting} \\ \textbf{Themselves}} & 
                                        \multicolumn{1}{p{2.2cm}}{\centering \textbf{Avoiding} \textbf{Embarrassment}} & \multicolumn{1}{p{1.3cm}}{\centering \textbf{Gaining} \\ \textbf{Esteem}} & \multicolumn{1}{p{1.4cm}}{\centering \textbf{Protecting} \\ \textbf{Others}} 
                                        & \multicolumn{1}{p{1.4cm}}{\centering \textbf{Defaming} \\ \textbf{Esteem}}

                                        \\  \hline 



\multicolumn{1}{l}{\textbf{Tweet}} &
\cellcolor{yellow!50}0.678 &
\cellcolor{yellow!50}0.632 &
\cellcolor{yellow!50}0.619 &
\cellcolor{yellow!50}0.62 &
\cellcolor{blue!50}{\textcolor{white}{0.759}} &
\cellcolor{green!50}0.831 &
\cellcolor{green!50}0.807 &
\cellcolor{blue!50}{\textcolor{white}{0.771}} &
\cellcolor{green!50}0.846 &
\cellcolor{blue!50}\textcolor{white}{0.790} &
\cellcolor{blue!50}{\textcolor{white}{0.752}} &
\cellcolor{yellow!50}0.692 &
\cellcolor{blue!50}{\textcolor{white}{0.744}} &
\cellcolor{yellow!50}0.637 &
\cellcolor{yellow!50}0.609 \\

\multicolumn{1}{l}{\textbf{Fake News}} &
\cellcolor{blue!50}{\textcolor{white}{0.719}} &
\cellcolor{yellow!50}0.661 &
\cellcolor{yellow!50}{0.683} &
\cellcolor{yellow!50}0.603 &
\cellcolor{blue!50}{\textcolor{white}{0.727}} &
\cellcolor{green!50}0.878 &
\cellcolor{green!50}0.845 &
\cellcolor{green!50}0.811 &
\cellcolor{green!50}0.892 &
\cellcolor{blue!50}{\textcolor{white}{0.759}} &
\cellcolor{green!50}0.81 &
\cellcolor{blue!50}{\textcolor{white}{0.738}} &
\cellcolor{yellow!50}0.677 &
\cellcolor{blue!50}{\textcolor{white}{0.709}} &
\cellcolor{yellow!50}{0.681} \\


\bottomrule   
\end{tabular}%
} 
\vspace{-1mm}
\caption{Kappa score representation for layer 1: \textit{type of omission} 
layer 2: \textit{colors of lies}, 
and layer 3: \textit{Intent of lies}.
Kappa score for the layer 4 topic of lies
can be found in Appendix \ref{sec: Data Annotation}.}
\vspace{-3.5mm}
\label{tab: Kappa score}
\end{table*}

%% file: 4_PromptEngineering.tex
\section{Data Augmentation}
\vspace{-1mm}
\label{sec:data_augmentation}
It is widely acknowledged that neural network-based techniques have a high demand for data. To address this data requirement, data augmentation has almost become a standard practice in the AI community \cite{van2001art,shorten2021text,liu2020survey}. 
We have utilized three methods for data augmentation here: (i) paraphrasing, (ii) 5W masking followed by infilling \cite{gao-etal-2022-mask}.

\subsection{Paraphrasing Deceptive Datapoints}
\vspace{-1mm}
The motivation for paraphrasing deceptive data stems from the diverse manifestations of textual deceptive content in real-world scenarios, often influenced by variations in writing styles among different news publishing outlets. It is vital to incorporate these variations in order to establish a robust benchmark that facilitates comprehensive evaluation and analysis (cf. Figure \ref{fig: paraphrase} in Appendix \ref{sec:paraphrase-evaluation} for examples).
 
Undoubtedly, manual generation of possible paraphrases is ideal; however, this process is time-consuming and labor-intensive. On the other hand, automatic paraphrasing has garnered significant attention recently \cite{niu2020unsupervised, nicula2021automated, witteveen2019paraphrasing, nighojkar2021improving}. We used GPT-3.5 \cite{brown2020language} (specifically the \textit{text-davinci-003} variant) \cite{brown2020language} model as it generates linguistically diverse, grammatically correct, and a maximum number of considerable paraphrases, i.e., 5 in this case. This is the best-performing model for data augmentation using paraphrasing \cite{rani2023factify5wqa}. Additionally, we conducted experiments with Pegasus \cite{zhang2020pegasus} and T5 (T5-Large) \cite{raffel2020exploring} models, but GPT-3.5 (\texttt{text-davinci-003} variant) \cite{brown2020language} outperformed them, as indicated in Appendix \ref{sec:paraphrase-evaluation}. We gathered a total of 25,500 unique paraphrased deceptive data points through this method. 

At this stage, several important questions arise: (i) \emph{What is the accuracy of the paraphrases generated?} (ii) \emph{How do they differ from or distort the original content?} To address these questions, we have conducted extensive experiments and obtained empirical answers. However, due to space limitations, please refer to Appendix \ref{sec:paraphrase-evaluation} for details of our experiments and conclusions. We have evaluated the paraphrase modules based on three key dimensions: \textit{(i) \ul{Coverage}: number of considerable paraphrase generations, (ii) \ul{Correctness}: correctness of these generations, and (iii) \ul{Diversity}: linguistic diversity in these generations}.

\subsection{Synthetic Data Augmentation using 5W Specific Mask Infilling}
\vspace{-1mm}
As mentioned previously in section ~\ref{sec:introduction}, our hypothesis revolves around the possible omission of the 5W (who, what, when, where, and why) for deceits. With this in mind, we developed a pipeline to detect the presence of the 5W and subsequently replace them with deceptive/null information generated from a language model. In the subsequent subsections, we will present our methodology for designing 5W semantic role labeling and mask filling techniques to address 5W omission.

\vspace{-1mm}
\begin{figure}[!tbh]
\vspace{-1mm}
\centering
\includegraphics[width=1\columnwidth, trim={0cm 0cm 0cm 0cm}]{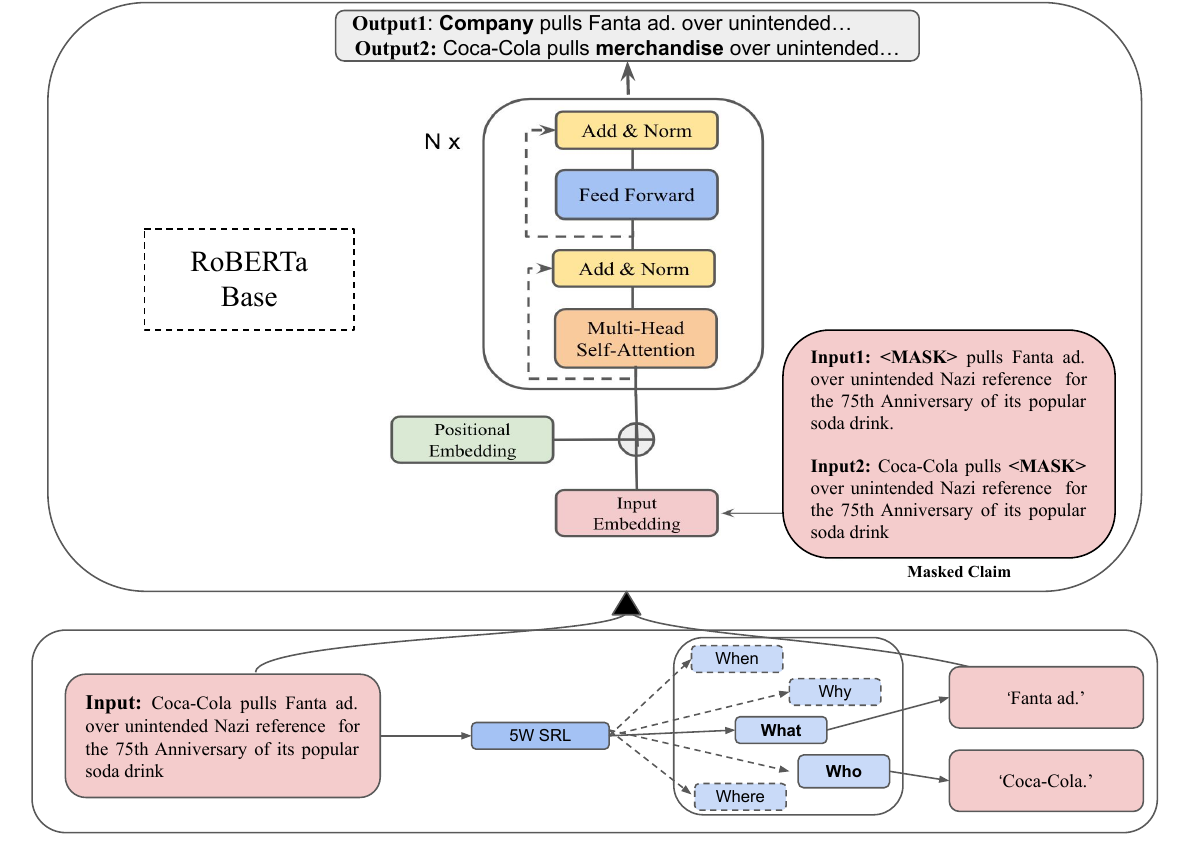}
\vspace{-6mm}
\caption{Architecture representation for the process of leveraging mask infilling using RoBERTa \cite{liu2019roberta} for creating the deception dataset.}
\label{fig: architecture_MaskInfilling}
\vspace{-2.5mm}
\end{figure}

\noindent
\textbf{5W Semantic Role Labeling:}
Identification of the functional semantic roles played by various words or phrases in a given sentence is known as semantic role labeling (SRL). SRL is a well-explored area within the NLP community. There are quite a few off-the-shelf tools available: (i) Stanford SRL \cite{manning2014stanford}, (ii) AllenNLP \cite{allennlpsrl}, etc. A typical SRL system initially identifies the verbs in a given sentence and subsequently associates all the related words/phrases with the verb through relational projection, assigning them appropriate roles. Thematic roles are generally marked by standard roles defined by the Proposition Bank (generally referred to as PropBank) \cite{palmer2005proposition}, such as: \textit{Arg0, Arg1, Arg2}, and so on. We propose a mapping mechanism to map these PropBank arguments to 5W semantic roles (look at the conversion table \ref{tab:5w-map-SRL}, in appendix).

\noindent
\textbf{5W Slot Filling:} Building upon our hypothesis, it is plausible for individuals to deliberately omit any of the given W to transform a statement into a lie of omission. Therefore, once we detect the presence of the Ws, our objective is to generate variations of the original statement by selectively omitting specific Ws. For this purpose, we train a masked LLM as depicted in the Figure \ref{fig: architecture_MaskInfilling}. For the 5W slot-filling task we have experimented with five models: (i) MPNet \cite{song2020mpnet}
, (ii) ELECTRA \cite{clark2020electra},
(iii) RoBERTa \cite{liu2019roberta}, (iv) ALBERT \cite{lan2019albert}, and (v) BERT \cite{devlin2018bert}.

RoBERTa \cite{liu2019roberta}, a language model that leverages large-scale pre-training and removes the next sentence prediction objective, significantly enhancing language understanding. With its transformer architecture and fine-tuning, it predicts the original masked tokens in an \textit{input sequence X} by maximizing the likelihood of the true masked tokens given the predicted \textit{probabilities P}. Considering the scenario where all the Ws are present in a sentence, it is feasible to generate five variations. At this juncture, a crucial question arises: is there a high likelihood that the generated sentences deviate substantially from the original deceptive input? To substantiate we have calculated BLEU \cite{papineni2002bleu} score and MoverScore \cite{zhao-etal-2019-moverscore} between the original input and all the perturbed generations, reported in Table~\ref{tab:Evaluation_MaskInfilling}.


\input{table/MaskInfillingEvaluation}

%% file: table/MaskInfillingEvaluation.tex
\begin{table}[h]
\centering
\vspace{-1mm}
\resizebox{0.95\columnwidth}{!}{
\begin{tabular}{lcc}
\toprule
\multicolumn{1}{l}{\textbf{Model}} & \textbf{BLEU Score} & \textbf{MoverScore} \\ \toprule
RoBERTa-base                       & \bf{0.7457}  & \bf{0.7216}          \\ 
MPNet-base               & 0.7329       & 0.7194     \\ 
ELECTRA-large-generator     & 0.7225   &0.7129           \\ 
BERT-base-uncase                   & 0.7222   &0.712           \\ 
ALBERT-large-v2                    & 0.7116    &0.703          \\ 
\bottomrule
\end{tabular}}
\vspace{-1mm}
\caption{BLEU Score for various models for mask infilling. RoBERTa performed the best.}
\label{tab:Evaluation_MaskInfilling}
\vspace{-5mm}
\end{table}
\vspace{-2mm}

%% file: 5_MultitaskLearning.tex
\section{Designing the SEPSIS Classifier}
\label{sec:sepsis_classifier}
\vspace{-1mm}


\begin{figure*}[!htp]
\centering
\vspace{-3mm}
\includegraphics[width=0.86\textwidth, trim={0cm 0cm 0cm 0cm}]{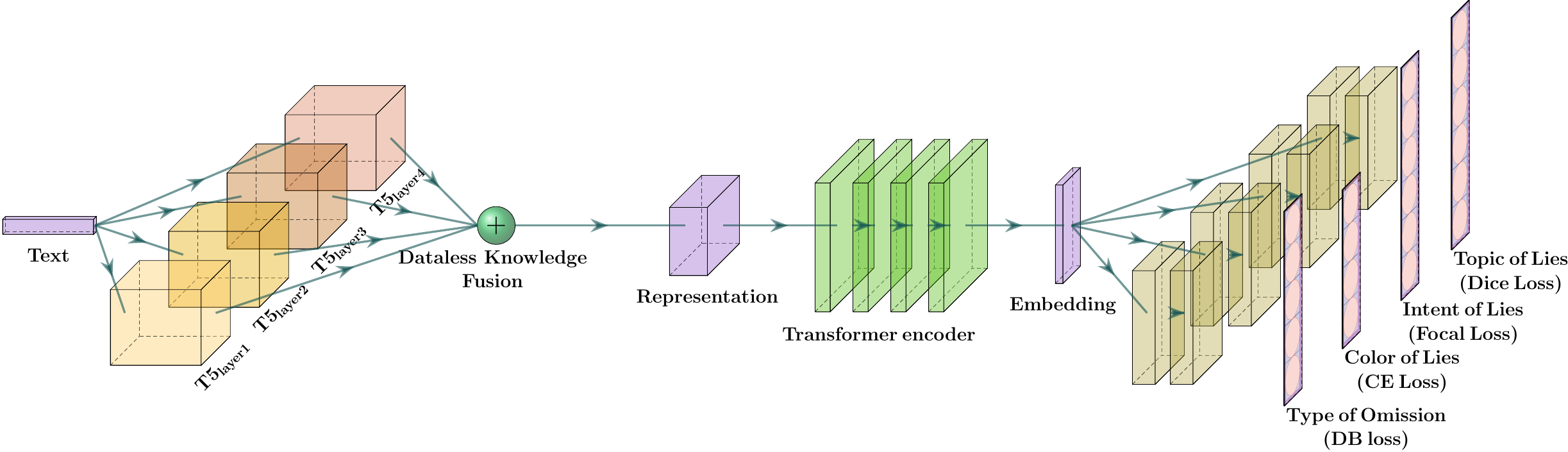}
\vspace{-2mm}
\caption{Multi-task learning architecture delineating the process of an input text going through labeling along four dimensions: (i) types of omission, (ii) colors of lie, (iii) intention of lie, and (iv) topic of lie. Here, DB Loss stands for Distribution-balanced Loss and CE loss stands for Cross Entropy loss (cf. Appendix \ref{sec:MTL_loss_function}).}
\label{fig:MTL}
\vspace{-3.5mm}
\end{figure*}

\begin{figure}[H]
\centering
\vspace{-2mm}
\includegraphics[width=0.49\textwidth, trim={0cm 0cm 0cm 0cm}]{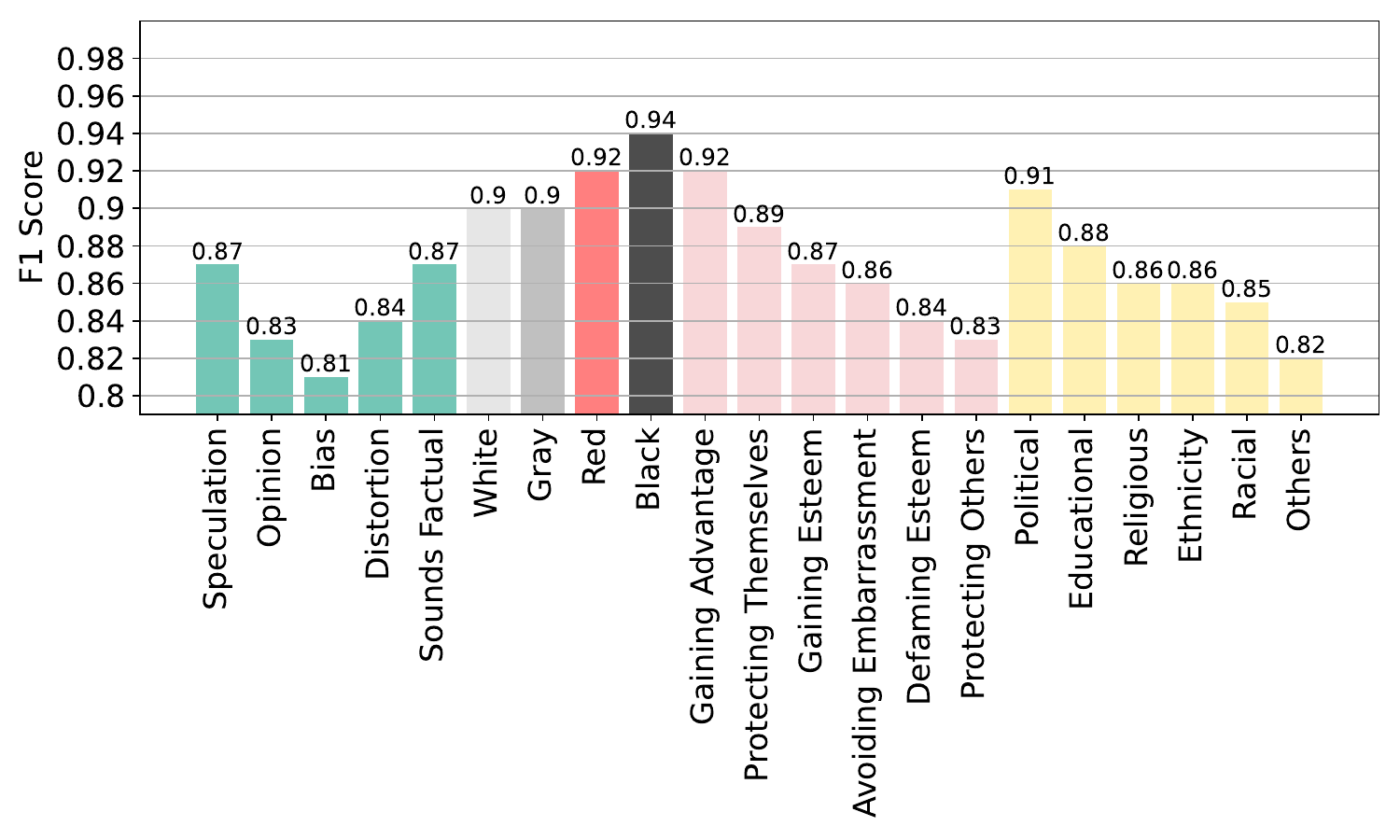}
\vspace{-7mm}
\caption{SEPSIS's F1 score for all classes of deception varies from 0.81 to 0.94. We have reported accuracy, precision, and recall as well (cf. Appendix \ref{sec:experimental setup}, tab \ref{tab:overall_exp}).}
\label{fig:MTL_result}
\vspace{-4mm}
\end{figure}

\textls[-10]{SEPSIS, by its design, is a multitask-multilabel problem requiring the application of Multitask Learning (MTL) techniques. In general MTL framework utilizes a shared representation for all the tasks. It has been observed by several researchers \cite{parisotto2015actor, rusu2015policy, yu2020gradient, fifty2021efficiently} that shared representation has its own limitations and further effects on learning task-specific loss functions. In our approach, we introduced two specific innovations, detailed in subsequent sections. Using the MTL model (Fig. \ref{fig:MTL}), we achieved a score of 0.81 F1 score on the human-annotated dataset (5000 samples) and 0.87 F1 score on the SEPSIS dataset (0.8M data points). Fig.~\ref{fig:MTL_result} shows the F1 score across deception classes on the SEPSIS dataset (cf. Appendix \ref{sec:MTL}).}

\vspace{-1.5mm}
\vspace{-2mm}

\noindent
\vspace{-5mm}
\subsection{Merging Finetuned LMs Brings Power!}
\vspace{-1mm}
Drawing inspiration from \cite{jin2022dataless}, we incorporated techniques for merging multiple fine-tuned LMs, a process referred to as \emph{dataless merging}. During our experimentation with various LMs, we found that T5 performed exceptionally well for our specific case, and was also the best LM for dataless merging as emphasized in \cite{jin2022dataless}. For the four layers of deception, we fine-tuned four T5 models using the data outlined in Table ~\ref{tab:SEPSIS_corpus}. These models are denoted as T5\textsubscript{layer1}, T5\textsubscript{layer2}, T5\textsubscript{layer3}, and T5\textsubscript{layer4}. By leveraging the methodology proposed in \cite{jin2022dataless}, we merged these fine-tuned T5 models to achieve a better-shared representation tailored to our specific objectives.  Figure ~\ref{fig:MTL} visually depicts the merging process via an architecture diagram. The code for reproducing experiments can be found at \url{https://bit.ly/3FglMtB}.
\noindent
\vspace{-3mm}
\subsection{Tailored Loss Function}
\vspace{-2mm}

\textls[-10]{During our exploration for suitable sub-task loss functions, we experimented with several available options, including (i) cross-entropy loss, (ii) focal loss \cite{lin2017focal}, (iii) dice loss \cite{li2019dice}, and (iv) distribution-balanced loss (DB) \cite{huang-etal-2021-balancing}. After a thorough evaluation, we observed that distribution-balanced loss yielded the best performance for layer 1, cross-entropy loss was most effective for layer 2, focal loss performed well for layer 3, and dice loss was the optimal choice for layer 4. For a comprehensive overview of the results and an in-depth discussion of different loss functions, please refer to the Appendix \ref{sec:MTL_loss_function}.}

%% file: 6_PropagandaTheory.tex
\vspace{-3mm}
\section{Dissecting Propaganda through the Lens of Deception}
\vspace{-1.5mm}
As mentioned earlier, numerous studies have explored the behavioral indicators of lying, but there is hardly any consensus on categorization. However, the focus of this paper specifically revolves around investigating \emph{lies of omission} and their connection to related research within the scientific community. Notably, there are works that have extensively examined the analysis of \emph{propaganda} through language \cite{da-san-martino-etal-2019-fine,martino2020survey}.

\begin{figure}[h]
\vspace{-3mm}  
\centering
  \centering
\includegraphics[width=0.75\columnwidth]{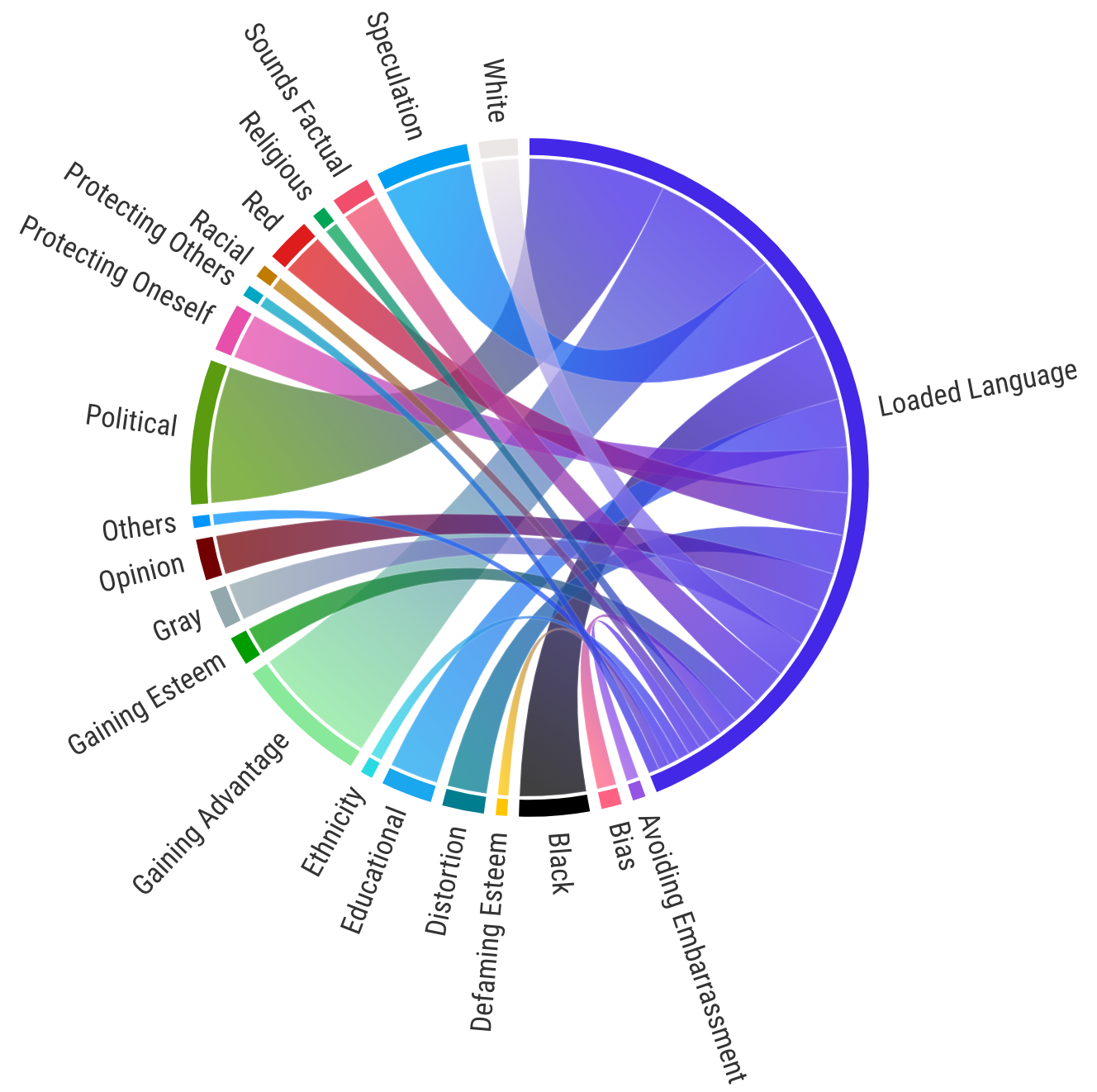}
\vspace{-1.5mm}
  \caption{The Circos presents the co-occurrence of all the layers of deception with a propaganda technique named \emph{loaded language}.}
  \label{fig:loaded_language}
\vspace{-4mm}  
\end{figure}

Our scientific curiosity led us to further investigate the specific types of \emph{lies of omission} employed in strategizing particular propaganda, such as \textit{exaggeration} and/or \textit{red herring}. To conduct this study, we utilized the propaganda datasets introduced by \cite{da-san-martino-etal-2019-fine} and applied the SEPSIS classifier, as discussed in section ~\ref{sec:sepsis_classifier} on the data. Through the analysis of these experiments, we made intriguing discoveries, including: (i) \emph{the prevalence of political topic in loaded language compared to other propaganda types}, (ii) \emph{the close association between the intention of gaining advantage and Name Calling}, and (iii) \emph{the complexity underlying causal simplification as a form of speculation.} A Circos \cite{Flourish} example is presented in Fig. \ref{fig:loaded_language} for a propaganda technique named
\textit{loaded language} (cf. Appendix \ref{sec: Propaganda} for Circos diagrams corresponding to propaganda techniques). Therefore, we firmly believe that our research on SEPSIS not only stands on its own but also acts as a bridge, facilitating a deeper understanding of deception.

%% file: 2_related_works.tex
\section{Related Works}
\vspace{-1mm}

Deception detection has been explored on a wide range of applications, such as online dating services \cite{toma2010reading} \cite{guadagno2012dating}, social networks \cite{ho2013guess}, consumer reviews \cite{li2014towards} \cite{ott2011finding}, and court transcripts \cite{fornaciari2013automatic} \cite{perez2015deception}. Significant research findings have demonstrated a correlation between gender and deceit \cite{perez2015experiments}, as well as a connection between deception and cultural factors \cite{perez2014cross}. The majority of conducted experiments are predicated on a binary classification approach for analyzing input text, specifically distinguishing between deceptive and non-deceptive instances as explored by \cite{mbaziira2016text} and \cite{mihalcea2009lie}. To the best of our knowledge, there is currently no computational study that comprehensively defines and categorizes deception by drawing insights from psychology. In our paper, we introduce SEPSIS, which presents a novel definition and dataset aimed at tackling the issue of \emph{lies of omission} in language. We firmly believe that SEPSIS holds the potential for establishing a connection between deception and fake news, and we intend to explore this further.

%% file: 7_conclusion.tex
\vspace{-1mm}
\section{Conclusion and Future Avenues}
\vspace{-1mm}


We have introduced SEPSIS, a novel multi-layered corpus focused on lies of omission. Furthermore, our MTL framework leverages recent advances in language model fine-tuning and dataless merging to optimize deception detection, achieving a 0.87 F1 score. Finally, we have uncovered compelling relationships between propaganda techniques and lies of omission through empirical analysis. The public release of our dataset and models will catalyze future research on this complex societal phenomenon.
\section{Discussion and Limitations}
\vspace{-1mm}
In this section, we self-criticize a few aspects that could be improved and also detail how we (tentatively) plan to improve upon those specific aspects-
\subsection{Categorization of deception}
We have considered the four layers and categories based on our understanding of the psychological framework and going manually through multiple samples to understand the type, intent, topic, and colors of lie. However, this list may not be exhaustive. This is the reason for us to have put an \textit{others} category in the topic of lies. Categories could increase when categorizing deception in real life.

\subsection{Data Augmentation}
We used paraphrasing and mask infilling for building the sepsis corpus. However, we understand that a few generations might not be deceptive and could have generated non-deceptive texts. However, we have done extensive manual testing, and believe such cases are nominal. 

\subsection{SEPSIS Classifier}

One of the limitations of the SEPSIS Classifier is the computational heaviness associated with fine-tuning the T5 model for each specific layer. This process requires considerable computational resources and time. As the T5 models need to be finetuned for each task head, so total computational time increase significantly with an increase in the number of task head. It is important to consider these computational limitations when implementing multi-task learning architectures, as they can impact the feasibility and scalability of the approach, particularly in scenarios with limited computational resources or a large number of output tasks.

\section{Ethical Considerations}


Through this framework, we propose models to classify deception. We also developed a large augmented deceptive dataset. However, we must address the potential misuse of the dataset and models by entities who may exploit the framework to generate deceptive texts such as creating fake news by manipulating the content. The deliberate dissemination of deceptive news, spreading propaganda techniques to shape public opinion, is also a significant concern. We vehemently discourage such misuse and strongly advise against it.

%% file: 9_faq.tex
\newpage
\onecolumn
\section*{Frequently Asked Questions (FAQs)}\label{sec:FAQs}

\begin{enumerate}

    \item 
    [\ding{93}] {\fontfamily{lmss} \selectfont \textbf{What were the specific instructions provided to the annotators and the criteria used for selecting them in the crowd annotation process of 5000 sentences through AMT?}}
    
    \vspace{-3mm}
    \begin{description}
    \item[\ding{224}] The annotation pipeline outlines a step-by-step approach to deception detection based on different layers, as shown in Figure 1. To ensure reliable annotations, the dataset source was kept undisclosed from the annotators. 
    Notably, for sentences categorized as "Sounds Factual," no additional annotations were made apart from missing W's.
    \end{description}


    \item[\ding{93}] {\fontfamily{lmss} \selectfont \textbf{How were the loss functions determined, specifically for each task head?}}

    \vspace{-3mm}
    \begin{description}
    \item[\ding{224}] The selection of loss functions for each task head was based on the characteristics of the class distribution for that specific task. If the class distribution was imbalanced, loss functions designed to handle such scenarios were chosen. Detailed explanations and experimental results supporting the choice of each loss function can be found in the appendix section \ref{sec:MTL}.
    \end{description}

    \item[\ding{93}] {\fontfamily{lmss} \selectfont \textbf{Why RoBERTa was finally chosen as our baseline model for the Mask Infilling task?}}

    \vspace{-3mm}
    \begin{description}
    \item[\ding{224}] Our experimentation in comparison to other state-of-the-art language models like RoBERTa-base, MPNet-base, ELECTRA-large-generator, BERT-base-uncase, and ALBERT-large-v2 revealed a higher Bilingual Evaluation Understudy (BLEU) score using RoBERTa. The selection of RoBERTa as the preferred model for the mask infilling task, based on its highest BLEU score, implies that RoBERTa's generated outputs exhibited a greater resemblance to the desired reference outputs. This characteristic of RoBERTa's performance is particularly advantageous for generating deceptive sentences that closely resemble reference sentences. By leveraging RoBERTa's capabilities, the task of producing deceptive sentences can be effectively achieved with a higher degree of fidelity to the reference sentences.
    \end{description}
    
    \item[\ding{93}] {\fontfamily{lmss} \selectfont \textbf{Why was the T5 base model chosen for model merging, and how was its performance evaluated?}}

    \vspace{-3mm}
    \begin{description}
    \item[\ding{224}] The selection of the T5 base model for model merging involved extensive experimentation and evaluation of various language models (LLMs), such as RoBERTa, T5, and DeBERTa. Our evaluation aimed to identify the LLM that would deliver the best performance for our specific case. Initially, we assessed the individual performance of each LLM by utilizing them in the architecture to generate word embeddings, without employing model merging or fine-tuning. However, there was no significant improvement in scores observed for RoBERTa and DeBERTa when compared to using the LLM as-is (without merging) or with model merging. In contrast, the T5 model demonstrated an additional 4-5\% improvement after applying Dataless Knowledge Fusion.
    \end{description}

    \item[\ding{93}] {\fontfamily{lmss} \selectfont \textbf{What are the details of the train-test validation split and other hyperparameters used for replicating the experiments?}} 

    \vspace{-3mm}
    \begin{description}
    \item[\ding{224}] The dataset was divided into an 80-20 train-test split, where 80\% of the data was used for training and 20\% for testing. To assess the model's performance, we employed 5-fold cross-validation.The train-test split was meticulously crafted to ensure that each sentence and its augmented versions are exclusively present in either the train set or the test set, but never in both. This careful arrangement guarantees the absence of any sentence overlap (i.e. sentence "S" present in train split and paraphrased version of sentence "S" present in test spilt), maintaining the integrity of the data and enhancing the overall quality of the split. The train-test split of the dataset would be made available along with all the hyperparameters of the code on GitHub for replication of the results.
    \end{description}

    \item[\ding{93}] {\fontfamily{lmss} \selectfont \textbf{What motivated the use of data augmentations and multi-task learning, and what improvement was achieved?}} 

\vspace{-3mm}
\begin{description}
\item[\ding{224}] In our initial experiment, without employing multi-task learning and data augmentation, we achieved an average accuracy score of 0.758 (averaged across all classes). Recognizing correlations between the classes, we introduced multi-task learning to capitalize on these relationships. To further enhance the model's robustness, we applied data augmentation. The improvements in average accuracy are detailed in the appendix table \ref{tab:overall_exp}. The code for reproducing experiments can be found at  \url{https://anonymous.4open.science/r/deception_MTL-60DB/}.
\end{description}

\end{enumerate}

%% file: 10_appendix.tex
\newpage
\onecolumn
\appendix
\renewcommand{\thesubsection}{\Alph{section}.\arabic{subsection}}
\renewcommand{\thesection}{\Alph{section}}
\setcounter{section}{0}

\section*{Appendix}\label{sec:appendix}
This section provides supplementary material in the form of additional examples, implementation details, etc. to bolster the reader's understanding of the concepts presented in this work.

\section{Lies of omission -- across cultures}\label{sec:app-A}
Instances of lies of omission can be discovered in ancient literature from diverse cultures across the globe. In order to stimulate further discussion and provide motivation, we will present (in the appendix - due to obvious space limitation) two specific examples—one from the Western tradition and another from the Eastern tradition. These examples serve to highlight the prevalence and significance of lies of omission in literature and emphasize the need for deeper exploration of this phenomenon.

\noindent
\textbf{The merchant of Venice}: In Shakespeare's play, Antonio, an antisemitic merchant, borrows money from the Jewish moneylender Shylock in order to assist his friend in pursuing a relationship with Portia. Antonio can't repay the loan, and without mercy, Shylock demands a pound of his flesh as collateral. At this critical moment, Portia, who is now married to Antonio's friend, disguises herself as a lawyer and intervenes to save Antonio. Though the agreement allows Shylock to claim a pound of flesh, he must ensure that not a single drop of blood is shed, as causing harm to a Christian is strictly forbidden by law.

\noindent
\textbf{Mahabharata} - \emph{Ashwathama hatho, naro va kunjaro va}: This story is derived from an ancient Indian epic \emph{"The Mahabharta"}. In this excerpt, \emph{Ashwathama} is an elephant. \emph{Ashwathama} was also the name of the son of Guru Dronacharya. Yudhishtir, one of the Pandavas and \emph{Dharmraj} (which means he would never lie), faces the daunting task of confronting his unbeatable mentor, Guru Dronacharya, from whom he and his brothers had learned the art of warfare. Reluctant to engage in direct combat against his beloved teacher, Yudhishtir follows the advice of Lord Krishna and employs a strategy of omission. He announces the death of Ashwathama, but discreetly adds the words "naro va kunjaro va," indicating that it is actually a question whether the deceased Ashwathama is a human or an elephant. While Yudhishtir technically did not prevaricate, the news of his son's supposed demise deeply affects Guru Dronacharya, causing him to lose his will to fight and making it easier for Yudhishtir to overcome him. The story highlights Yudhishtir's adherence to his principles of truthfulness while employing a clever tactic of omission to gain an advantage in the battle.

\section{Dataset Curation}
This contains additional information on data sources, data cleaning, annotation, and Inter annotator agreement
\subsection{Data Sources}\label{sec:data sources}
Information Security and Object Technology (ISOT) fake news dataset \cite{ISOTFakeNewsDataset}: This dataset contains two types of articles fake and real news. This dataset was collected from real-world sources; the truthful articles were obtained by crawling articles from Reuters.com (News website). As for the fake news articles, they were collected from different sources. The fake news articles were collected from unreliable websites that were flagged by Politifact (a fact-checking organization in the USA) and Wikipedia. For this research, the fake news dataset is leveraged. The data source has a file named “Fake.csv” which contains more than 12,600 articles from different fake news outlet resources. Each article contains the following information: article title, text, type, and the date the article was published on. We chose 2500 data points randomly from this set for this research.

\subsection{Data Cleaning and Annotation Quality check}\label{sec: data cleaning}
Data cleaning involves two iterations, data set preparation, and a human-level review of the manual annotations. The process involved the removal of URLs and unnecessary internet taxonomy with the aim of increasing data quality. To further increase the quality of data for human understanding, we reviewed the annotations manually by following the below-mentioned steps:
\vspace{-3mm}
\begin{itemize}
\setlength\itemsep{0em}
    \item Accounting for multiple annotations against a single field by the same annotator by getting rid of one of the two annotations along the lines of the definitions formulated at the start of the process.
   
    \item Filling in for fields annotated by the first entity and missed by the second entity by accounting for the gaps by building along the lines of definitions established earlier. 
    Correcting typographical errors implicating a similar meaning.
    
    \item Overriding annotations for a couple of data items where the reviewer found them overwhelmingly wrong.
\end{itemize}
\vspace{-3mm}

\subsection{Inter Annotator Agreement
}\label{sec: Data Annotation}
In the ~\cref{sec:iaa_score} we have reported inter-annotator scores for all the 3 layers in ~\cref{tab: Kappa score}. In addition, here we are reporting inter-annotator agreement for the topic of lie in the ~\cref{tab:iaa_topic_of_lie}.

\begin{table}[!tbh]
\centering
\resizebox{0.7\textwidth}{!}{
\begin{tabular}{lcccccc}
\toprule
          & Political & Educational & Religious & Ethnicity & Racial & Others \\
\midrule
Twitter   &\cellcolor{green!50} 0.82      &\cellcolor{blue!50} 0.78        & \cellcolor{green!50}0.81      &\cellcolor{blue!50} 0.73      & \cellcolor{blue!50}0.76   & \cellcolor{blue!50}0.72   \\
Fake News & \cellcolor{green!50}0.87      &\cellcolor{green!50} 0.84        &\cellcolor{green!50} 0.85      &\cellcolor{blue!50} 0.77      & \cellcolor{green!50}0.82   &\cellcolor{blue!50} 0.79  
\\
\bottomrule
\end{tabular}
}
\label{tab:iaa_topic_of_lie}
\caption{Inter Annotator Agreement score for Topic of Lies.}
\end{table}

\vspace{-2mm}
\subsection{Data Analysis of SEPSIS Corpus and Insights}\label{sec: data analysis}

This section contains a thorough analysis of the entire corpus.

\begin{figure}[H]
  \centering
  \begin{subfigure}[b]{0.3\textwidth}
    \includegraphics[width=\linewidth, trim={1cm 1cm 1cm 1cm}]{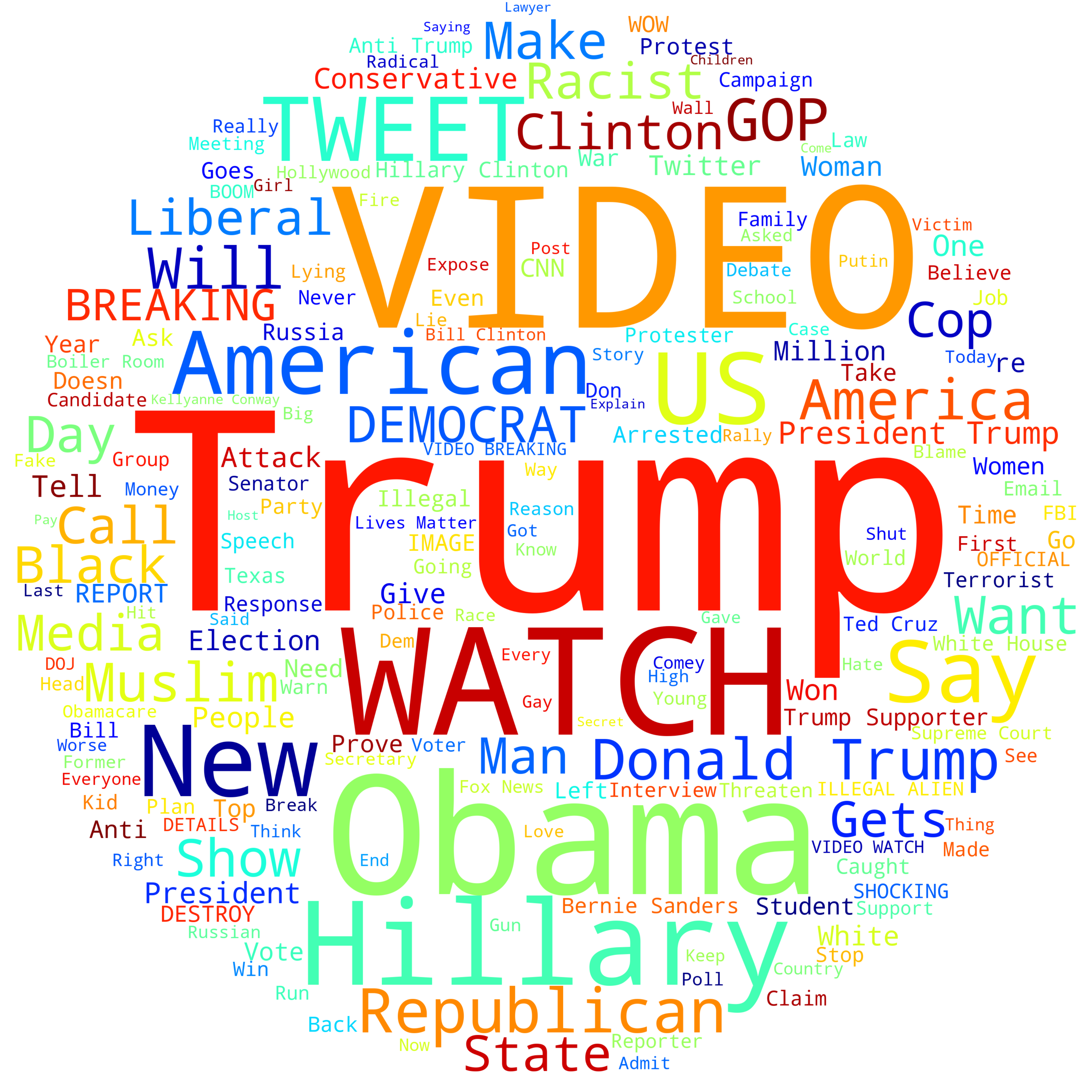}
    \caption{Word cloud of data collected from ISOT fake news.}
    \label{fig:fakenews-5k}
  \end{subfigure}
    \hspace{2cm}
    \begin{subfigure}[b]{0.3\textwidth}\includegraphics[width=\linewidth, trim={1cm 1cm 1cm 1cm}]{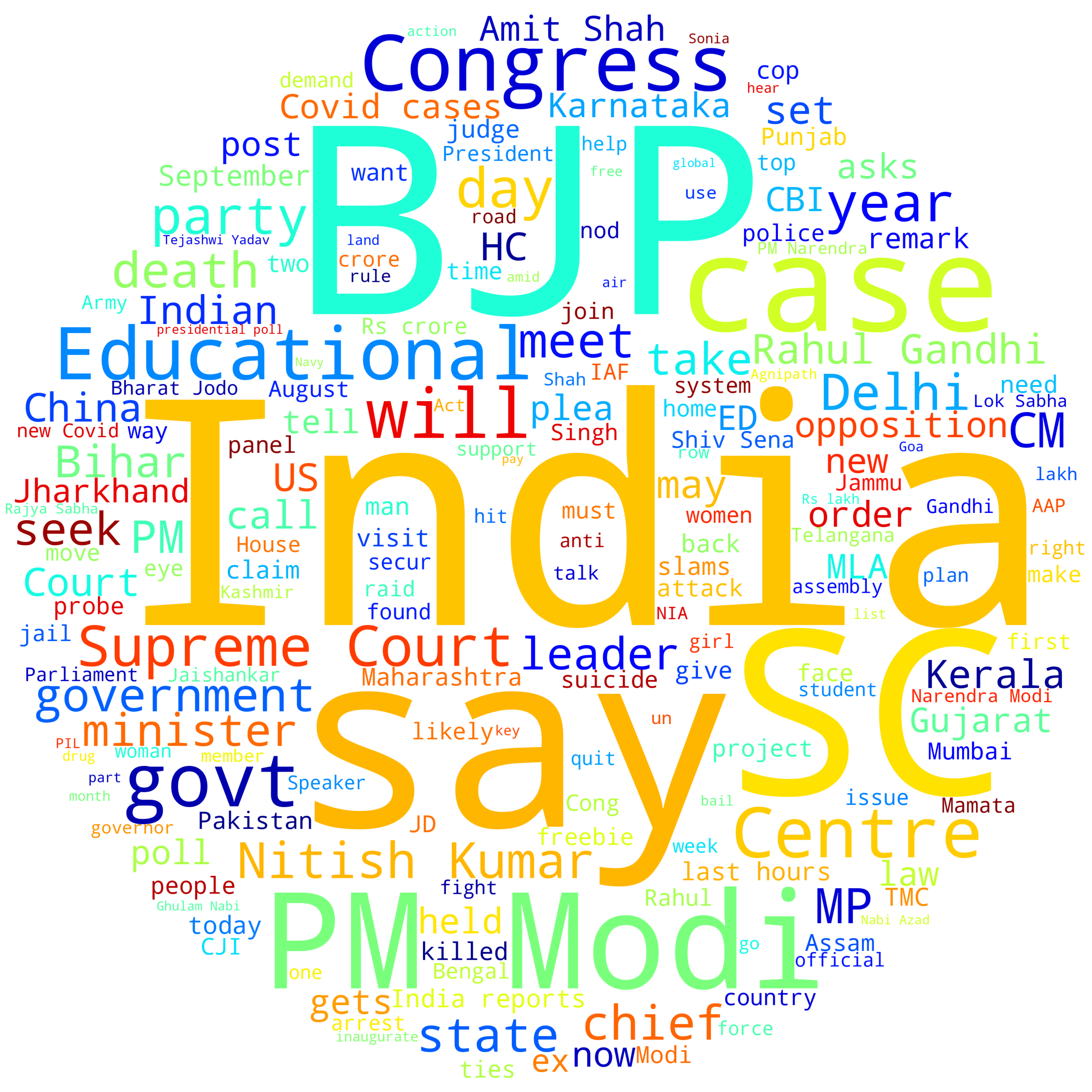}
    \caption{Word cloud of data collected from Times of India.}
    \label{fig:tweets-5k}
  \end{subfigure}
  \label{fig:combined_figure}
\end{figure}

\noindent
\textbf{Word representation of the sepsis corpus}: We have utilized two different data sources to understand the frequency of words, we present the word clouds in fig \ref{fig:fakenews-5k} and fig \ref{fig:tweets-5k}. An interesting insight is figure \ref{fig:fakenews-5k} represents US news and figure \ref{fig:tweets-5k} represents the Indian media house.

\noindent
\textbf{Statistics on categories across entire corpus:}
We further present the percentage of each feature across the entire dataset as represented in table \ref{tab: SEPSIS_breakup}.



\input{table/appendix_layer_values}


\noindent
\textbf{Percentage presence of 5Ws across all datapoints}:
Since we utilize 5W-based mask infilling, we also present \% of 5Ws across the entire dataset. and the statistics around it can be found in the table \ref{tab:percentage_presence} below.

\input{table/5W_presence}

\noindent
\textbf{Co-occurence percentage}: The four layers are connected to the input sentence. To study the co-occurrence across all categories and layers, we present them in heatmaps as described in fig \ref{fig:heatmap_layers}.

When analyzing lies of omission and colors of lies, we observe a strong correlation between speculation and black lies. Additionally, a significant majority of speculative texts can be categorized as political in nature. This association becomes even more apparent when we delve into the Intent of Lie on Lies of Omission. It is evident that the primary objective behind the creation of speculative texts is to gain an advantage. Black lies, in particular, are frequently employed for this purpose. It is noteworthy that political texts predominantly consist of black lies, serving as a means to gain an advantage.

\input{heatmaps/fourlayerconnection}

\newpage
\section{Data Augmentation}
For data augmentation, we have used two techniques (i) Paraphrasing and (ii) 5W Mask Infilling.
We provide additional information on these techniques in the following subsection.
\subsection{Paraphrasing Deceptive Datapoints}\label{sec:paraphrase-evaluation}

The underlying drive for paraphrasing textual assertions stems from the need to address variations that exist in real-life written content. The same textual claim might take on several different shapes since different news publishing companies use a variety of writing techniques. It is essential to create a solid standard for a thorough examination by taking these variations into account ( example in Figure \ref{fig: paraphrase}).

\begin{figure}[!tbh]
\centering
\resizebox{\columnwidth}{!}{%
\fbox
{%
    \parbox{\columnwidth}{%
    \fontsize{9}{10}\selectfont{
    \textcolor{blue}{Sasan Goodarzi, the CEO of software giant Intuit, which has avoided mass layoffs, says tech firms axed jobs because they misread the pandemic.}}
    \\
    \textbf{Prphr 1:} Sasan Goodarzi, the CEO of Intuit, a software giant that refrained from massive layoffs, explains that tech companies terminated employees due to their misinterpretation of the pandemic.
     \\
    \textbf{Prphr 2:} Intuit's CEO, Sasan Goodarzi, highlights that unlike other tech firms, the software giant avoided extensive job cuts as they correctly understood the impact of the pandemic.
     \\
    \textbf{Prphr 3:} The pandemic was misinterpreted by tech companies, leading them to lay off employees, according to Sasan Goodarzi, CEO of Intuit, a software giant that took a different approach and did not resort to mass layoffs.
     \\
    \textbf{Prphr 4:} Sasan Goodarzi, the CEO of Intuit, a software giant, asserts that tech companies made a mistake by laying off staff members because they failed to comprehend the true nature of the pandemic.
     \\
    \textbf{Prphr 5:} In contrast to tech firms that made the wrong call and downsized their workforce, Intuit, led by CEO Sasan Goodarzi, correctly assessed the pandemic and refrained from mass layoffs.
    }
    }%
}

\caption{Deceptive paraphrased data obtained using \texttt{text-davinci-003} \cite{brown2020language}.}
\label{fig: paraphrase}
\end{figure}

\begin{wrapfigure}{R}{8cm}
\centering
\resizebox{0.5\columnwidth}{!}{
\includegraphics[width=0.85\columnwidth, height=8cm, trim={0cm 0.5cm 0cm 0cm}]{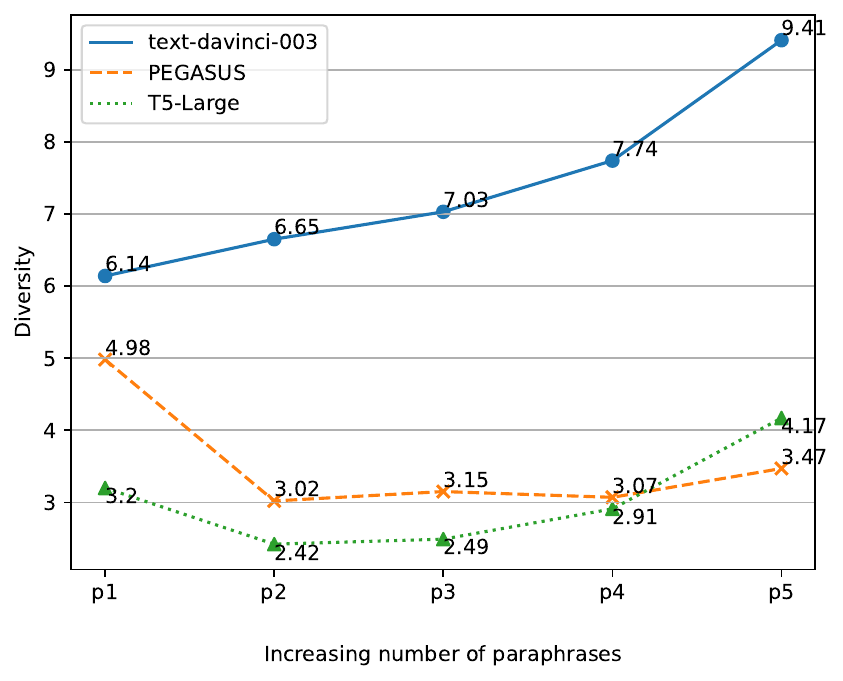}}
\caption{A higher diversity score depicts an increase in the number of generated paraphrases and linguistic variations in those generated paraphrases.}
\label{fig: parr}
\end{wrapfigure}

 To generate multiple paraphrases for a given claim, we employ state-of-the-art (SoTA) models. When selecting the appropriate paraphrase model from a list of available options, our main consideration is to ensure that the generated paraphrases exhibit both linguistic correctness and rich diversity. The process we follow to achieve this can be outlined as follows: Let's assume we have a claim denoted as $c$. Using a paraphrasing model, we generate $n$ paraphrases, resulting in a set of paraphrases $p_1^c$, $p_2^c$, ..., $p_n^c$. Subsequently, we conduct pairwise comparisons between these paraphrases and the original claim $c$, giving us comparisons such as $c-p_1^c$, $c-p_2^c$, ..., $c-p_n^c$. At this stage, we identify the examples that exhibit entailment, selecting only those for further consideration. To determine entailment, we utilize RoBERTa Large \cite{liu2019roberta}, a state-of-the-art model trained on the SNLI task \cite{bowman2015large}.

However, it is important to consider various secondary factors when evaluating paraphrase models. For instance, one model may generate a limited number of paraphrase variations compared to others, but those variations might be more accurate and consistent. Therefore, we took into account three key dimensions in our evaluation: \textit{(i) the number of meaningful paraphrase generations, (ii) the correctness of those generations, and (iii) the linguistic diversity exhibited by the generated paraphrases}. In our experiments, we explored the capabilities of three available models: (a) Pegasus \cite{zhang2020pegasus}, (b) T5 (T5-Large) \cite{raffel2020exploring}, and (c) GPT-3 (specifically, the \texttt{text-davinci-003} variant) \cite{brown2020language}. Based on empirical observations and analysis, we found that GPT-3 consistently outperformed the other models. To ensure transparency regarding our experimental process, we provide a detailed description of the aforementioned evaluation dimensions as follows.

\input{table/Appendix_paraphrasing}

\textbf{Coverage - Generating a substantial number of paraphrases:} Our objective is to generate up to five paraphrases for each given claim. After generating the paraphrases, we employ the concept of minimum edit distance (MED) \cite{wagner1974string} to assess the similarity between the paraphrase candidates and the original claim (with word-level units instead of individual characters). If the MED exceeds a threshold of ±2 for a particular paraphrase candidate (e.g., $c-p_1^c$), we consider it as a viable paraphrase and retain it for further evaluation. However, if the MED is within the threshold, we discard that particular paraphrase. By employing this setup, we evaluated all three models to determine which one generates the highest number of meaningful paraphrases.

\textbf{Correctness - Ensuring correctness in the generated paraphrases:} Following the initial filtration step, we conducted pairwise entailment assessments using the RoBERTa Large model \cite{liu2019roberta}, which is a state-of-the-art model trained on the SNLI dataset \cite{bowman2015large}. We retained only those paraphrase candidates that were identified as entailed by the RoBERTa Large model.

\textbf{Diversity - Ensuring linguistic diversity in the generated paraphrases:} Our focus was to select a model that could produce paraphrases with greater linguistic diversity. To assess the dissimilarities between the generated paraphrase claims, we compared pairs such as $c-p_n^c$, $p_1^c-p_n^c$, $p_2^c-p_n^c$, ..., $p_{n-1}^c-p_n^c$ for each paraphrase. We repeated this process for all other paraphrases and calculated the average dissimilarity score. Since there is no specific metric to measure dissimilarity, we utilized the inverse of the BLEU score \cite{papineni2002bleu}. This allowed us to gauge the linguistic diversity exhibited by a given model. Based on these experiments, we observed that the \texttt{text-davinci-003} variant performed the best in terms of linguistic diversity. The results of the experiment are presented in the table below. Moreover, we prioritized the selection of a model that maximized linguistic variations, and \texttt{text-davinci-003} excelled in this regard as well. The diversity vs. chosen models plot is illustrated in Figure ~\ref{fig: parr}.

\subsection{Data Augmentation using 5W Mask Infilling}
This mapping describes how Propbank roles are mapped to 5Ws(Who, What, When, Where, Why). We have used this mapping for mask infilling.
\input{table/map_5w_srl}

\section{Multi-Task Learning}\label{sec:MTL}

In this section, we delve into the specific architectural choices, experimental setup, and the formulation of the loss function employed for multi-task learning frameworks: The SEPSIS Classifier. By exploring the intricacies of this approach, we aim to shed light on the systematic integration of multiple tasks into a unified learning framework, ultimately enabling the model to effectively leverage synergistic information across layers of Deception. 

\subsection{Architectural Discussion}

 Multi-task learning (MTL) has emerged as a powerful paradigm for training deep neural networks to perform multiple related tasks simultaneously. In this paper, we propose a multi-task learning-based architecture for predicting four different tasks of the Deception dataset. The main advantage of using multi-task learning is the ability to leverage shared information across tasks, leading to improved model generalization and increased efficiency in training and inference. By jointly training multiple tasks, the model learns useful representations that are transferable to other related tasks, leading to better overall performance \cite{caruana1997multitask}. 

 \subsubsection{Dataless Knowledge Fusion}

 In many cases, LLMs are trained using domain-specific datasets, which can limit their performance when applied to out-of-domain cases. To address this challenge, we employ a fine-tuning approach on the T5-base model for each specific task, resulting in a total of four finetuned T5-based models (one model corresponding to one task). To leverage these models in our Multitask learning architecture, we employ Dataless Knowledge Fusion \cite{jin2022dataless} on these four finetuned T5-models into a single, more generalized model that exhibits improved performance in multitask learning (from here referred \textit{merged-fine-tuned-T5}).

 \subsubsection{Methodology}

 Our methodology takes a sentence as input and converts it into a latent embedding. The process of creating this rich embedding involves a two-stage approach. Firstly, we leverage the model-merging technique \cite{jin2022dataless}, which merges fine-tuned models sharing the same architecture and pre-trained weights, resulting in enhanced performance and improved generalization capabilities, particularly when dealing with out-of-domain data \cite{jin2022dataless}. Once the word embeddings are obtained from this merged model, the second stage involves converting them into a latent representation using the transformer encoder module. This representation is then propagated through four task-specific multilabel heads to obtain the output labels for each of the layers of Deception.



\subsection{Loss Functions}\label{sec:MTL_loss_function}
This section contains an in-depth discussion of different loss functions that we used for different tasks of MTL architecture.

\subsubsection{Cross-Entropy Loss}
Cross entropy loss, also known as log loss or logistic loss, is a commonly used loss function in machine learning, particularly in classification tasks. It measures the dissimilarity between the predicted probabilities of classes and the true labels of the data. The log loss function penalizes incorrect predictions more strongly, meaning that as the predicted probability deviates further from the true label, the loss increases. The loss approaches zero when the predicted probability aligns with the true label.

For the SEPSIS classifier, i.e., multi-label classification task with n classes, the cross-entropy loss is calculated as the average of the individual binary cross-entropy losses for each class. 
\begin{equation}
L_{B C E}= \begin{cases}-\log \left(p_i^k\right) & \text { if } y_i^k=1 \\ -\log \left(1-p_i^k\right) & \text { otherwise }\end{cases}
\end{equation}

where,
\begin{itemize}
\item $y^k=$ $\left[y_1^k, \ldots, y_C^k\right] \in\{0,1\}^C(C$ is the number of classes),
\item{$p_i^k$ is the predicted probability distribution across the classes}
\end{itemize}

\subsubsection{Focal Loss}

Focal loss is a modification of the cross entropy loss that addresses the issue of class imbalance in multi-class classification tasks \cite{lin2017focal}. In the standard multi-class cross-entropy loss, all classes are treated equally, which can be problematic when dealing with imbalanced datasets where certain classes have a much smaller representation. Focal loss aims to down-weight the contribution of well-classified examples and focuses more on difficult and misclassified examples. The focal loss for multi-label classification is defined as follows:
\begin{equation}
L_{F L}= \begin{cases}-\left(1-p_i^k\right)^\gamma \log \left(p_i^k\right) & \text { if } y_i^k=1 \\ -\left(p_i^k\right)^\gamma \log \left(1-p_i^k\right) & \text { otherwise }\end{cases}
\end{equation}

where:
\begin{itemize}
\item{$p_i^k$ is the predicted probability distribution across the classes}
\item{\(\gamma\) is the focusing parameter that controls the degree of down weighting. It is usually set to a value greater than 0. We used \(\gamma\) = 2 in our experiment.}
\end{itemize}
The focal loss formula introduces the term \((1 - p_{i})^{\gamma}\) which acts as a modulating factor. This factor down weights well-classified examples 
\(p_i^k\) close to 1 and assigns them a lower contribution to the loss. The focusing parameter gamma controls how much the loss is down-weighted. Higher values of gamma place more emphasis on difficult examples.
By incorporating the focal loss into the training objective, the model can effectively handle class imbalance and focus more on challenging examples.

\subsubsection{Dice Loss}
The Dice loss is a similarity-based loss function commonly used in image segmentation tasks and data-imbalanced multi-class classification problems. It measures the overlap or similarity between predicted and true labels. For multi-label classification, the Dice loss can be defined as follows:
\begin{equation}
L_{DL} = 1 - \frac{2 \sum_{i=1}^C y_i^k \cdot p_i^k+\epsilon}{\sum_{i=1}^C y_i^k+\sum_{i=1}^C p_i^k+\epsilon}
\end{equation}

\begin{itemize}
\item{C is the number of classes}
\item{\(y_i^k\) represents the true label for class C, which can be either 0 or 1 for each label.}
\item{\(p_i^k\) represents the predicted probability or output for class c}

\end{itemize}
The formula calculates the Dice coefficient for each example by summing the products of the true labels \(y_i^k\) and predicted probabilities \(p_i^k\) for each class C. The numerator represents the intersection between the predicted and true labels, while the denominator represents the sum of the predicted and true labels, which corresponds to the union of the two sets. By subtracting the Dice coefficient from 1, we obtain the Dice loss.

By using the Dice loss, the model is encouraged to focus on correctly identifying and predicting the minority classes, as the loss is computed based on the intersection and sum of true and predicted labels for each class. This property is especially valuable in data-imbalanced settings, as it helps to alleviate the bias towards majority classes and improve the model's ability to capture and predict the minority classes accurately.

\subsubsection{Distribution-balanced Loss}

The distribution-balanced (DB) loss function is a promising solution for addressing class imbalance and label dependency in multilabel text classification tasks. Unlike traditional approaches such as resampling and re-weighting, which often lead to oversampling common labels, the DB loss function tackles these challenges directly. By inherently considering the class distribution and label linkage, it offers a more effective alternative for achieving balanced training.

According to \cite{huang-etal-2021-balancing}, the application of the DB loss function has demonstrated superior performance compared to commonly used loss functions in multi-label scenarios. This novel approach addresses the problem of class imbalance, where certain labels are significantly underrepresented, and considers the relationship and dependencies between different labels. By striking a balance between these factors, the DB loss function ensures that the training process is fair and unbiased, resulting in improved accuracy and robustness in multilabel text classification tasks. 

For multi-label classification, the Distribution-balanced loss can be defined as follows:
\begin{equation}
L_{D B}= \begin{cases}-\hat{r}_{D B}\left(1-q_i^k\right)^\gamma \log \left(q_i^k\right) & \text { if } y_i^k=1 \\ -\hat{r}_{D B} \frac{1}{\lambda}\left(q_i^k\right)^\gamma \log \left(1-q_i^k\right) & \text { otherwise }\end{cases}
\end{equation}

where:
\begin{itemize}
\item{C is the number of classes}
\item $\hat{r}_{D B}=\alpha+\sigma\left(\beta \times\left(r_{D B}-\mu\right)\right)$ $\rightarrow$ $r_{D B}= \frac{\frac{1}{C} \frac{1}{n_i}} {\frac{1}{C} \sum_{y_i^k=1} \frac{1}{n_i}}$
\item{\(y_{i}\) represents the true label }
\item {$\lambda$ scale factor}
\end{itemize}

The distribution-balanced loss combines rebalanced weighting and negative-tolerant regularization (NTR) to address key challenges in multi-label scenarios. It effectively reduces redundant information arising from label co-occurrence, which is crucial in such tasks. Additionally, the loss explicitly assigns lower weights to negative instances that are considered "easy-to-classify," thereby improving the model's ability to handle these instances effectively. \cite{wu2020distribution}



\subsubsection{Rationale for choosing loss function for the particular task.}

The selection of specific loss functions for each task is driven by various factors and considerations. 

\begin{enumerate}

    \item \textbf{Distribution-balanced loss function for Types of Omission:} Due to the strong multi-label nature and skewed distribution of the Types of Omission layer, the Distribution-balanced loss function is utilized \cite{huang-etal-2021-balancing}. This loss function is specifically designed to handle extreme multi-label scenarios and skewed class distributions, providing a more balanced and effective training process for the model.

    \item \textbf{Cross Entropy loss for Color of Lie}:
The Color of Lie layer is relatively class-wise balanced. In such cases, the Cross-Entropy loss is a commonly used and standard loss function. It is well-suited for balanced class distributions and helps the model effectively learn and classify the color of lies.

    \item \textbf{Focal loss for Intent of Lie:}  The Intent of Lie layer is a class-imbalanced scenario. In such situations, the Focal loss has shown to perform well. Focal loss down-weights easy examples and focuses more on hard, misclassified examples, which helps in addressing class imbalance and improving the model's performance on classification of minority classes.

    \item \textbf{Dice loss for Topic of Lie:} The Topic of Lie layer is also a class-imbalanced scenario. The Dice loss has demonstrated effectiveness in handling class imbalance. Hence we used the Dice loss for this layer so that, the model can better capture and predict the minority topics.

The rationale behind selecting focal loss for the Intent of lie and Dice loss for the topic of lie is based on experimentation. Initially, we tried the opposite combination, which resulted in an F1 score of 0.85 for the Intent of lie and a score of 0.85 for the topic of lie. However, in the current configuration, we achieved improved performance with an F1 score of 0.87 for the Intent of lie and a score of 0.86 for the topic of lie. Therefore, after careful evaluation, we opted for focal loss and Dice loss for their respective categories to maximize overall performance.

\end{enumerate}

\subsection{Experimental results} \label{sec:experimental setup}
For overall experiments, we had 4 setups broadly.
\begin{itemize}

\item{T5 with LSTM encoder combined with no model merging}
\item{T5 with LSTM encoder combined with model merging}
\item{T5 with transformer encoder combined with no model merging}
\item{T5 with transformer encoder combined with model merging}

\end{itemize}
We used accuracy, precision, recall, and F1 score for evaluating the performance of our model. T5 with transformer encoder combined with model merging performed the best and results on these metrics for all experiments are presented in table \ref{tab:overall_exp}.

\input{table/Thelargetablefullversion}
\newpage
\section{Propaganda Techniques}\label{sec: Propaganda}
Propaganda techniques are strategies used to manipulate and influence people's opinions, emotions, and behavior in order to promote a particular agenda or ideology \cite{da-san-martino-etal-2019-fine, martino2020survey}. These techniques are often employed in mass media, advertising, politics, and public relations. While they can vary in their specific methods, we present definitions of 18 propaganda techniques that we have used in this study in the left box in the subsequent section. In the box on the right side, we present insights from propaganda techniques through deception.

\input{table/Appendix_PT}

\input{table/Appendix_PT2}


\input{table/appendix_circos}

%% file: table/appendix_layer_values.tex
\begin{table}[H]
\centering
\resizebox{0.5\columnwidth}{!}{
\begin{tabular}{clcc}
\toprule
\textbf{Layers of Deception}                                                                     & \textbf{\begin{tabular}[c]{@{}l@{}}Categories within the layer\end{tabular}} & \textbf{\begin{tabular}[c]{@{}c@{}}Number of datapoints\end{tabular}} & \textbf{Percentage} \\ \toprule
\multirow{5}{*}{\textbf{\begin{tabular}[c]{@{}c@{}}Layer 1:\\ \\ Type of Omission\end{tabular}}} & Speculation                                                                     & 311754                                                                        & 35.56\%                   \\
                                                                                                 & Bias                                                                            & 72268                                                                        & 8.24\%                   \\
                                                                                                 & Distortion                                                                      & 150249                                                                        & 17.14\%                   \\
                                                                                                 & Opinion                                                                         & 154590                                                                        & 17.63\%                   \\
                                                                                                 & Sounds Factual                                                                  & 187923                                                                        & 21.43\%                   \\ \hline
\multirow{4}{*}{\textbf{\begin{tabular}[c]{@{}c@{}}Layer 2:\\ Colors of Lies\end{tabular}}}      & Black                                                                           & 322634                                                                        & 45.31\%                   \\
                                                                                                 & White                                                                           & 90019                                                                        & 12.64\%                   \\
                                                                                                 & Gray                                                                            & 182161                                                                        & 25.58\%                   \\
                                                                                                 & Red                                                                             & 117245                                                                        & 16.47\%                   \\ \hline
\multirow{6}{*}{\textbf{\begin{tabular}[c]{@{}c@{}}Layer 3:\\ Intent of Lies\end{tabular}}}      & Gaining Advantage                                                               & 332661                                                                        & 47.73\%                  \\
                                                                                                 & Protecting Themselves                                                           & 202395                                                                        & 29.04\%                  \\
                                                                                                 & Gaining Esteem                                                                  & 124197                                                                        & 17.96\%                  \\
                                                                                                 & Avoiding Embarrasment                                                           & 24505                                                                        & 3.52\%                  \\
                                                                                                 & Defaming Esteem                                                                 & 6938                                                                        & 1.00\%                  \\
                                                                                                 & Protecting Others                                                               & 5236                                                                        & 0.75\%                  \\ \hline
\multirow{6}{*}{\textbf{\begin{tabular}[c]{@{}c@{}}Layer 4:\\ Topic of Lies\end{tabular}}}       & Political                                                                       & 546780                                                                        & 72.36\%                  \\
                                                                                                 & Educational                                                                     & 109596                                                                        & 14.50\%                  \\
                                                                                                 & Ethnicity                                                                       & 29343                                                                        & 3.88\%                  \\
                                                                                                 & Religious                                                                       & 27575                                                                        & 3.64\%                  \\
                                                                                                 & Racial                                                                          & 27354                                                                        & 3.61\%                  \\
                                                                                                 & Others                                                                          & 15250                                                                        & 2.01\%                  \\ \bottomrule
\end{tabular}
}
\caption{Breakup of SEPSIS datapoints over layers of deception and categories within each layer.}
\label{tab: SEPSIS_breakup}
\end{table}

%% file: table/5W_presence.tex
\begin{table}[H]
\centering
\resizebox{0.8\columnwidth}{!}{%
\begin{tabular}{@{}lccccc@{}}
\toprule
{}           &  \textbf{Who}  & \textbf{What} & \textbf{Why} & \textbf{When} & \textbf{Where}\\ \midrule
\% presence of 5W for tweets from Times of India          &   34.84\% & 53.06\% & 1.02\% & 6.31\% & 4.77\%      \\
\% presence of 5W from ISOT fake news dataset             &   36.40\% & 52.73\% & 1.41\% & 6.30\% & 3.16\%   \\
\bottomrule
\end{tabular}%
}
\caption{\% of 5Ws across the entire dataset.}
\label{tab:percentage_presence}
\end{table}

%% file: heatmaps/fourlayerconnection.tex
\begin{figure*}[!h]
    \begin{subfigure}[b]{0.5\textwidth}
    \centering
        \includegraphics[width=0.8\textwidth]{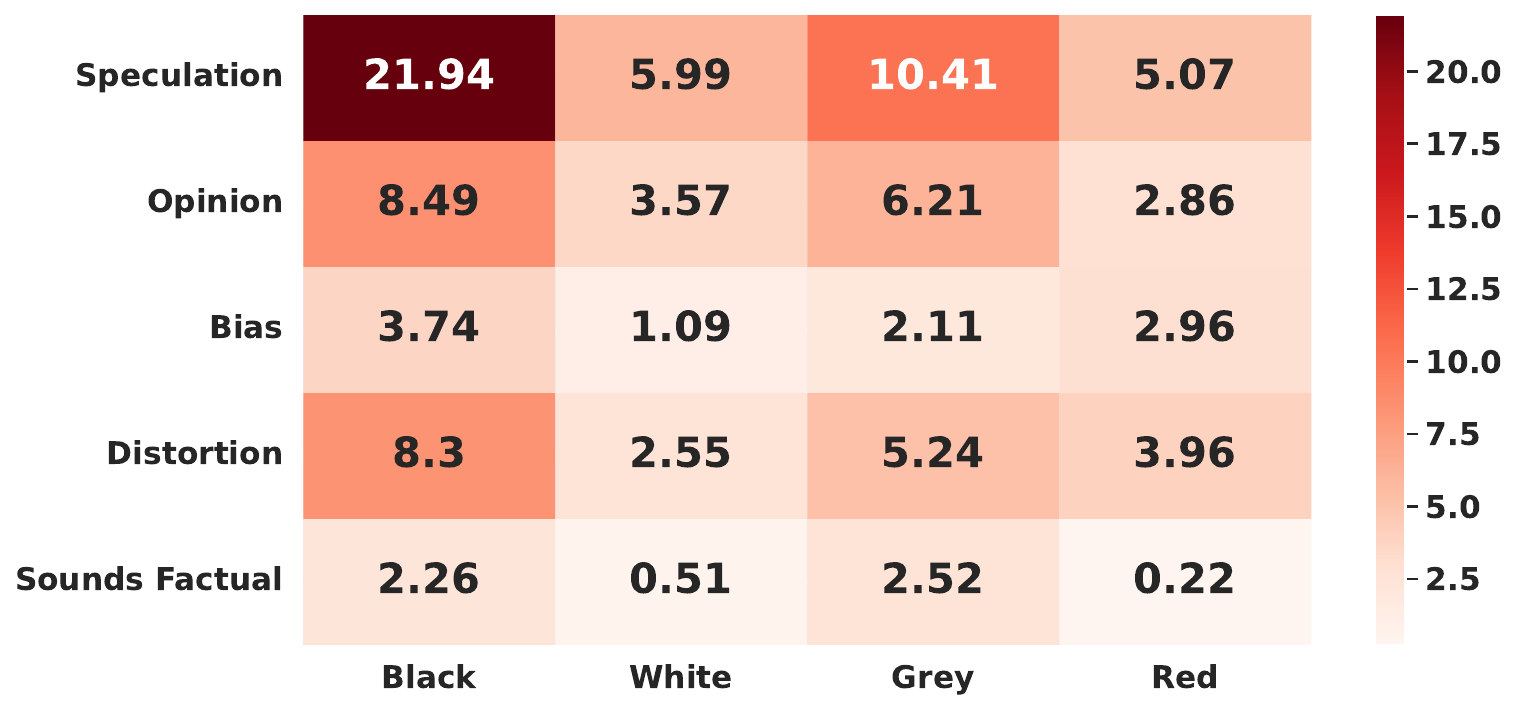}
        \caption{Lies of Omission-Colors of Lie.}
    \end{subfigure}
    \begin{subfigure}[b]{0.5\textwidth}
    \centering
        \includegraphics[width=0.8\textwidth]{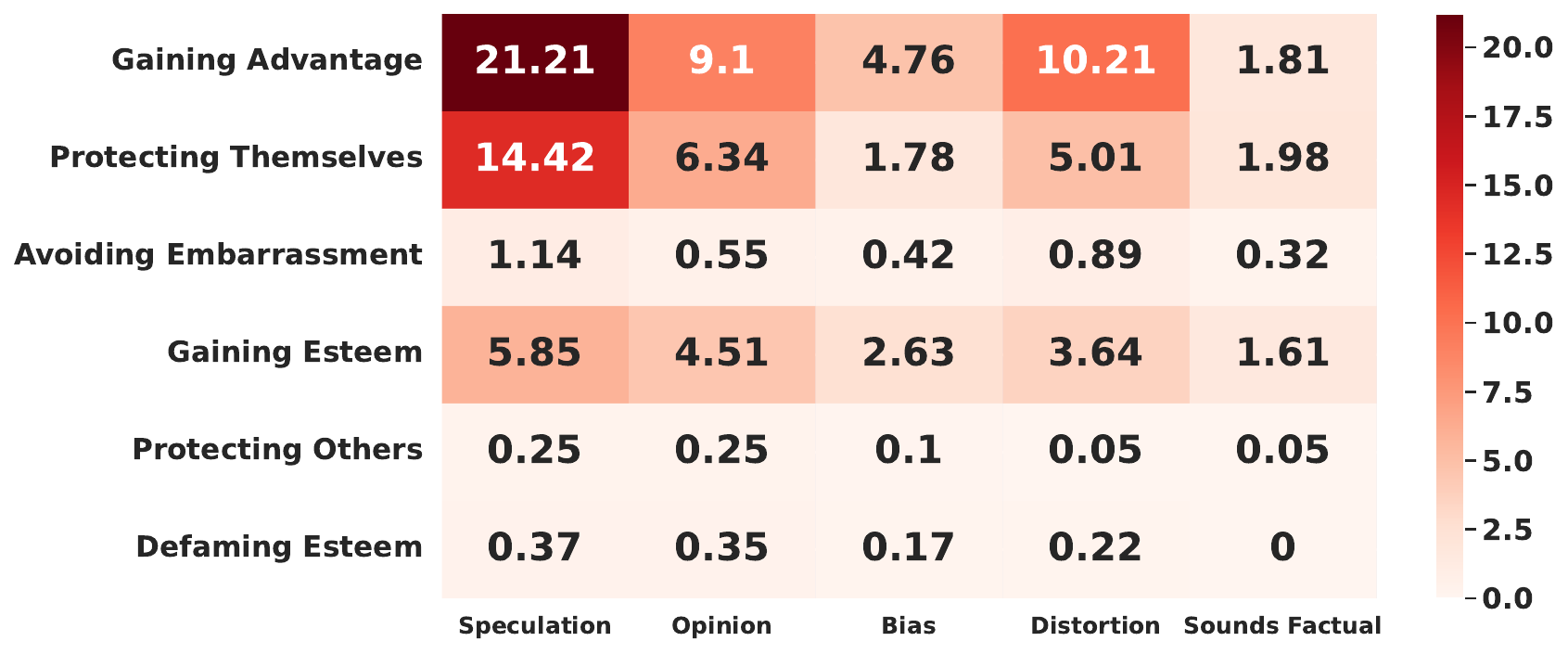}
        \caption{Lies of Omission-Intent of Lie.}
    \end{subfigure}
    
    \begin{subfigure}[b]{0.5\textwidth}
    \centering
        \includegraphics[width=0.8\textwidth]{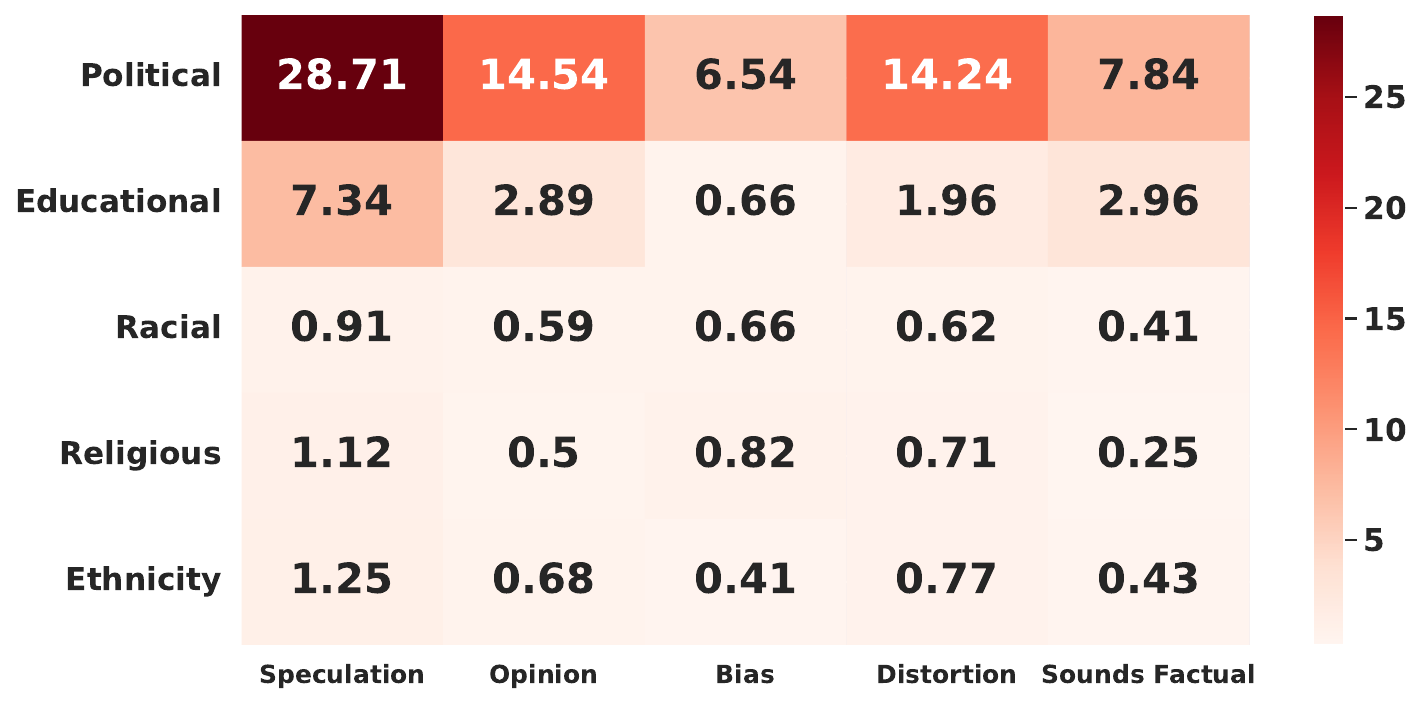}
        \caption{Type of omission-Topic of lie.}
    \end{subfigure}            
    \begin{subfigure}[b]{0.5\textwidth}
    \centering
        \includegraphics[width=0.8\textwidth]{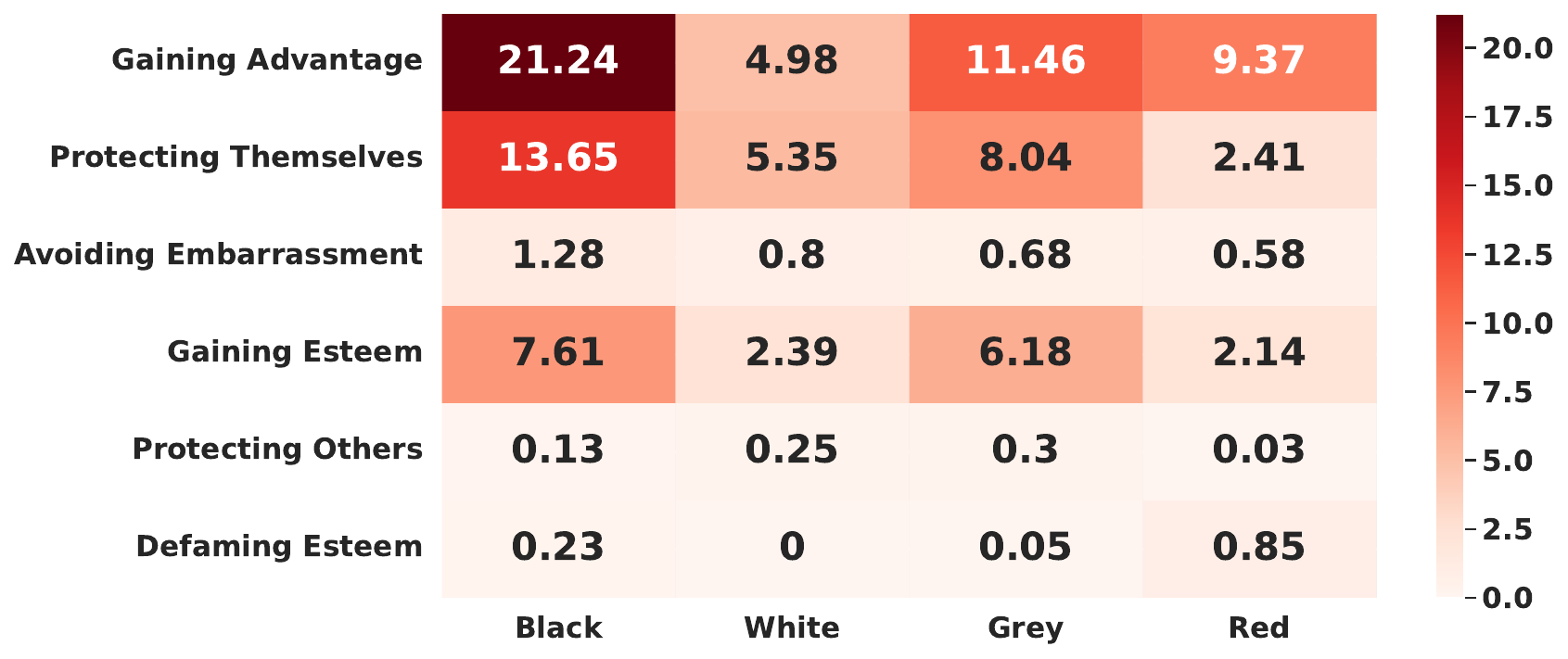}
        \caption{Colors of Lie-Intent of Lie.}
    \end{subfigure}
    
    \begin{subfigure}[b]{0.5\textwidth}
    \centering
        \includegraphics[width=0.8\textwidth]{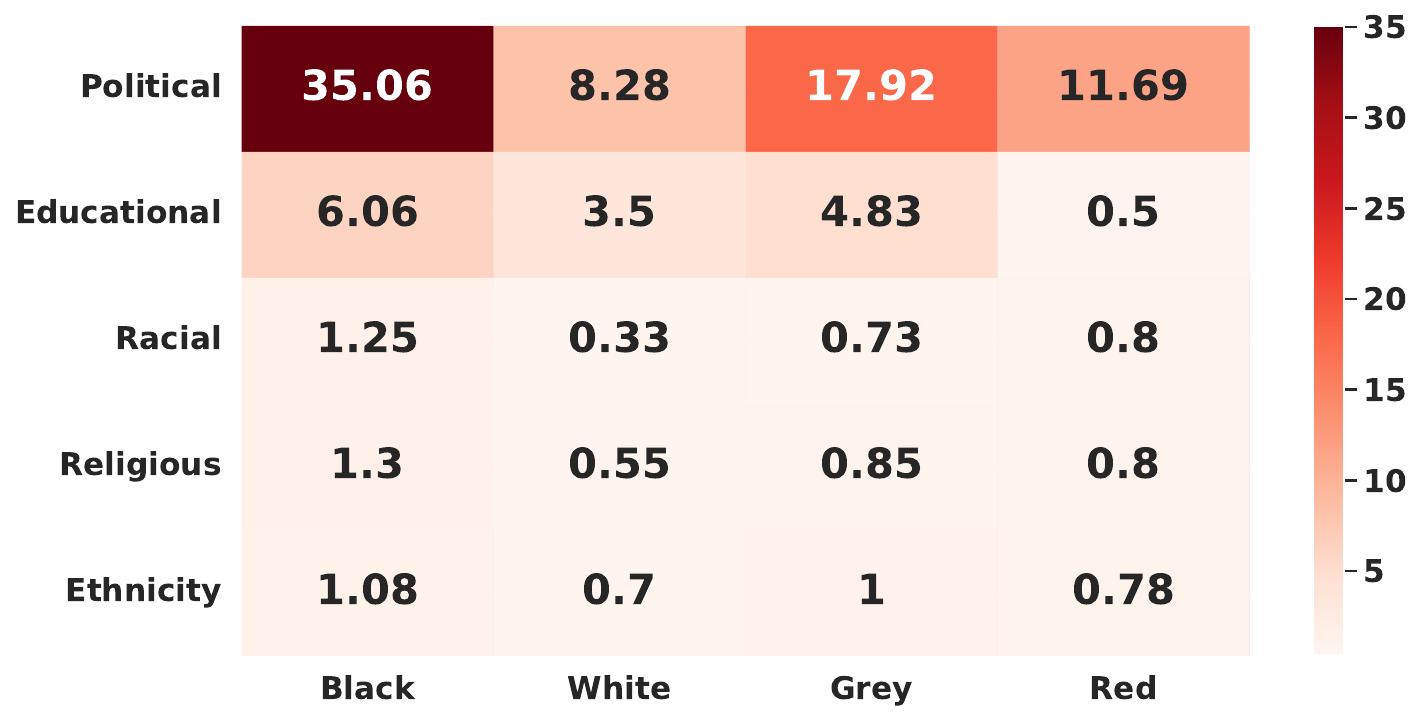}

        \caption{Colors of Lie-Influence of Lie.}
    \end{subfigure}
    \begin{subfigure}[b]{0.5\textwidth}
    \centering
        \includegraphics[width=0.8\textwidth]{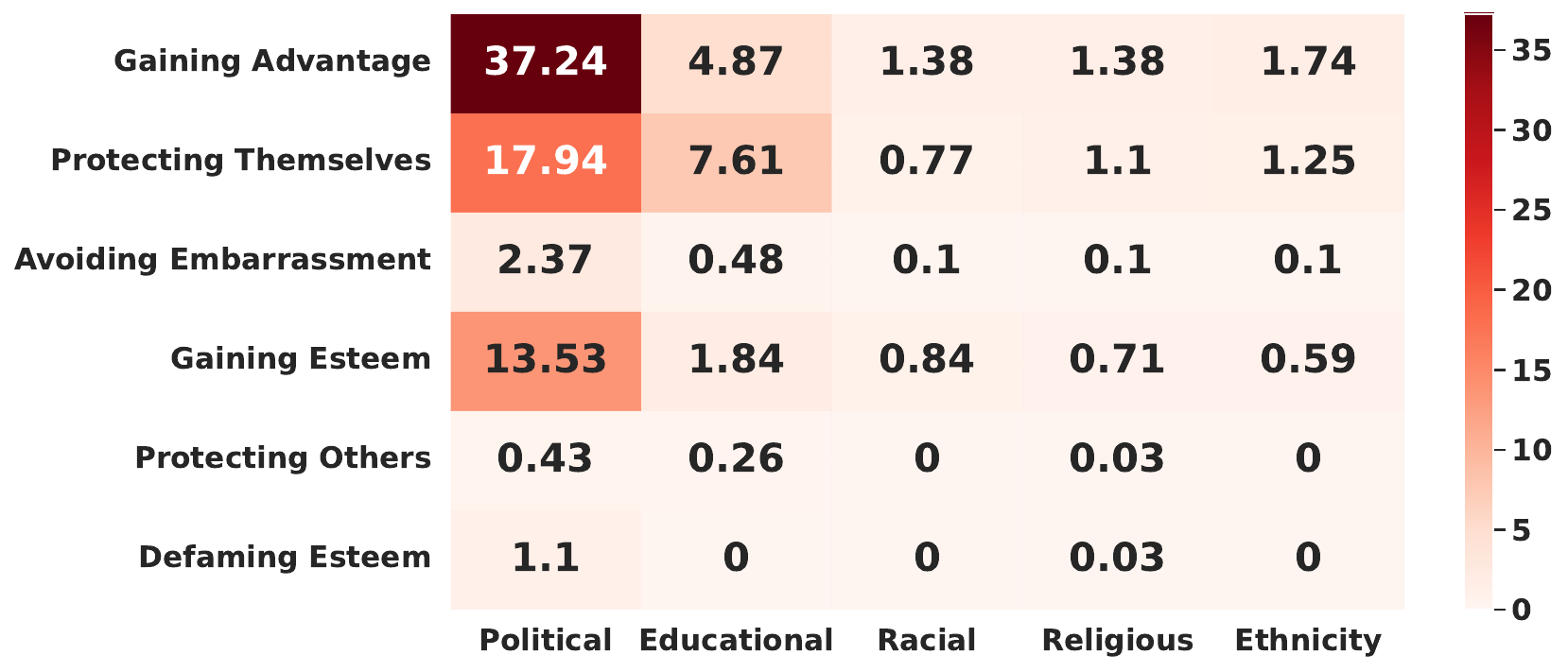}
        \caption{Intent of Lie-Influence of Lie.}
    \end{subfigure}
    
    \caption{The heatmaps provide a concise overview of the interconnections and overlaps among various layers of Lies. Numbers represents \% overlap. }
    \label{fig:heatmap_layers}
\end{figure*}

%% file: table/Appendix_paraphrasing.tex
\begin{table}[H]
\centering
\resizebox{0.5\columnwidth}{!}{%
\begin{tabular}{@{}lcccc@{}}
\toprule
Model           &  Coverage  & Correctness & Diversity \\ \midrule
Pegasus          &   31.98   &    93.23\%       &     3.53      \\
T5               &  30.09       &      84.56\%       &    3.04       \\
GPT-3 &   35.19   &   89.67\%      &     7.39      \\
\bottomrule
\end{tabular}
}
\caption{Experimental results of automatic paraphrasing models based on three factors: \textit{(i) coverage, (ii) correctness and (iii) diversity}; GPT-3 (\texttt{text-davinci-003}) can be seen as the most performant.}
\label{tab:my-table}
\end{table}

%% file: table/map_5w_srl.tex
\begin{table}[H]
\centering
\resizebox{0.5\columnwidth}{!}{
\begin{tabular}{ccccccc}
\toprule 
\textbf{PropBank Role }& \textbf{Who} & \textbf{What} & \textbf{When} & \textbf{Where} & \textbf{Why} & \textbf{How} \\
\midrule
\textbf{ARG0} & \textbf{84.48} & 0.00 & 3.33 & 0.00 & 0.00 & 0.00 \\
\textbf{ARG1} & 10.34 & \textbf{53.85} & 0.00 & 0.00 & 0.00 & 0.00 \\
\textbf{ARG2} & 0.00 & 9.89 & 0.00 & 0.00 & 0.00 & 0.00 \\
\textbf{ARG3} & 0.00 & 0.00 & 0.00 & 22.86 & 0.00 & 0.00 \\
\textbf{ARG4} & 0.00 & 3.29 & 0.00 & 34.29 & 0.00 & 0.00 \\
\textbf{ARGM-TMP} & 0.00 & 1.09 & \textbf{60.00} & 0.00 & 0.00 & 0.00 \\
\textbf{ARGM-LOC} & 0.00 & 1.09 & 10.00 & \textbf{25.71} & 0.00 & 0.00 \\
\textbf{ARGM-CAU} & 0.00 & 0.00 & 0.00 & 0.00 & \textbf{100.00} & 0.00 \\
\textbf{ARGM-ADV} & 0.00 & 4.39 & 20.00 & 0.00 & 0.00 & 0.06 \\
\textbf{ARGM-MNR} & 0.00 & 3.85 & 0.00 & 8.57 & 0.00 & \textbf{90.91} \\
\textbf{ARGM-MOD} & 0.00 & 4.39 & 0.00 & 0.00 & 0.00 & 0.00 \\
\textbf{ARGM-DIR} & 0.00 & 0.01 & 0.00 & 5.71 & 0.00 & 3.03 \\
\textbf{ARGM-DIS} & 0.00 & 1.65 & 0.00 & 0.00 & 0.00 & 0.00 \\
\textbf{ARGM-NEG} & 0.00 & 1.09 & 0.00 & 0.00 & 0.00 & 0.00 \\
\bottomrule
\end{tabular}
}
\caption{A mapping table from PropBank \cite{palmer2005proposition} {(\textit{Arg0, Arg1, ...})} to 5W {(\textit{Who, What, When, Where, and Why})}.}
\label{tab:5w-map-SRL}
\end{table}

%% file: table/Thelargetablefullversion.tex
\begin{table}[!tbh]
\resizebox{\columnwidth}{!}{
\begin{tabular}{l|l|l|llllllll||llllllll}
	\toprule
\multirow{2}{*}{}      & \multirow{2}{*}{\textbf{SEPSIS}}         & \multirow{2}{*}{\textbf{Labels}} & \multicolumn{8}{c||}{\textbf{Without Model Merging}}                                                                                                                                                                                                                                     & \multicolumn{8}{c|}{\textbf{With Model Merging}}                                                                                                                                                                                                                                        \\ \cline{4-19} 
                       &                                 &                         & \multicolumn{2}{c|}{\textbf{Accuracy} \%}                                         & \multicolumn{2}{l|}{\textbf{Precision}}                                       & \multicolumn{2}{l|}{\textbf{Recall}}                                            & \multicolumn{2}{l||}{\textbf{F1-Score}}                     & \multicolumn{2}{l|}{\textbf{Accuracy} \%}                                         & \multicolumn{2}{l}{\textbf{Precision}}                                         & \multicolumn{2}{l|}{\textbf{Recall}}                                            & \multicolumn{2}{l|}{\textbf{F1-Score}}                     \\
                       
                       \toprule
                       \multirow{21}{*}{\textbf{\parbox[c]{2.5cm}{\centering T5 with \\ LSTM \\ encoder}}}

 & \multirow{5}{*}{Type of Omission}     & Speculation             & \multicolumn{1}{l|}{82.58} & \multicolumn{1}{l|}{\multirow{5}{*}{80.25}} & \multicolumn{1}{l|}{0.78} & \multicolumn{1}{l|}{\multirow{5}{*}{0.77}} & \multicolumn{1}{l|}{0.83} & \multicolumn{1}{l|}{\multirow{5}{*}{0.80}} & \multicolumn{1}{l|}{0.8}  & \multirow{5}{*}{0.78} & \multicolumn{1}{l|}{86.15} & \multicolumn{1}{l|}{\multirow{5}{*}{82.89}} & \multicolumn{1}{l|}{0.84} & \multicolumn{1}{l|}{\multirow{5}{*}{0.81}} & \multicolumn{1}{l|}{0.85} & \multicolumn{1}{l|}{\multirow{5}{*}{0.83}} & \multicolumn{1}{l|}{0.84} & \multirow{5}{*}{0.82} \\ \cline{3-4} \cline{6-6} \cline{8-8} \cline{10-10} \cline{12-12} \cline{14-14} \cline{16-16} \cline{18-18}
                       &                                 & Opinion                 & \multicolumn{1}{l|}{80.76} & \multicolumn{1}{l|}{}                       & \multicolumn{1}{l|}{0.80} & \multicolumn{1}{l|}{}                      & \multicolumn{1}{l|}{0.79} & \multicolumn{1}{l|}{}                      & \multicolumn{1}{l|}{0.79} &                       & \multicolumn{1}{l|}{82.54} & \multicolumn{1}{l|}{}                       & \multicolumn{1}{l|}{0.82} & \multicolumn{1}{l|}{}                      & \multicolumn{1}{l|}{0.81} & \multicolumn{1}{l|}{}                      & \multicolumn{1}{l|}{0.81} &                       \\ \cline{3-4} \cline{6-6} \cline{8-8} \cline{10-10} \cline{12-12} \cline{14-14} \cline{16-16} \cline{18-18}
                       &                                 & Bais                    & \multicolumn{1}{l|}{74.92} & \multicolumn{1}{l|}{}                       & \multicolumn{1}{l|}{0.73} & \multicolumn{1}{l|}{}                      & \multicolumn{1}{l|}{0.76} & \multicolumn{1}{l|}{}                      & \multicolumn{1}{l|}{0.74} &                       & \multicolumn{1}{l|}{77.39} & \multicolumn{1}{l|}{}                       & \multicolumn{1}{l|}{0.75} & \multicolumn{1}{l|}{}                      & \multicolumn{1}{l|}{0.80} & \multicolumn{1}{l|}{}                      & \multicolumn{1}{l|}{0.77} &                       \\ \cline{3-4} \cline{6-6} \cline{8-8} \cline{10-10} \cline{12-12} \cline{14-14} \cline{16-16} \cline{18-18}
                       &                                 & Distortion              & \multicolumn{1}{l|}{79.51} & \multicolumn{1}{l|}{}                       & \multicolumn{1}{l|}{0.75} & \multicolumn{1}{l|}{}                      & \multicolumn{1}{l|}{0.78} & \multicolumn{1}{l|}{}                      & \multicolumn{1}{l|}{0.76} &                       & \multicolumn{1}{l|}{81.87} & \multicolumn{1}{l|}{}                       & \multicolumn{1}{l|}{0.8}  & \multicolumn{1}{l|}{}                      & \multicolumn{1}{l|}{0.82} & \multicolumn{1}{l|}{}                      & \multicolumn{1}{l|}{0.81} &                       \\ \cline{3-4} \cline{6-6} \cline{8-8} \cline{10-10} \cline{12-12} \cline{14-14} \cline{16-16} \cline{18-18}
                       &                                 & Sound Factual           & \multicolumn{1}{l|}{83.50} & \multicolumn{1}{l|}{}                       & \multicolumn{1}{l|}{0.79} & \multicolumn{1}{l|}{}                      & \multicolumn{1}{l|}{0.83} & \multicolumn{1}{l|}{}                      & \multicolumn{1}{l|}{0.81} &                       & \multicolumn{1}{l|}{86.48} & \multicolumn{1}{l|}{}                       & \multicolumn{1}{l|}{0.83} & \multicolumn{1}{l|}{}                      & \multicolumn{1}{l|}{0.86} & \multicolumn{1}{l|}{}                      & \multicolumn{1}{l|}{0.84} &                       \\ \cline{2-19} 
                       & \multirow{4}{*}{Color of Lie}   & White                   & \multicolumn{1}{l|}{85.68} & \multicolumn{1}{l|}{\multirow{4}{*}{86.37}} & \multicolumn{1}{l|}{0.83} & \multicolumn{1}{l|}{\multirow{4}{*}{0.84}} & \multicolumn{1}{l|}{0.86} & \multicolumn{1}{l|}{\multirow{4}{*}{0.84}} & \multicolumn{1}{l|}{0.84} & \multirow{4}{*}{0.84} & \multicolumn{1}{l|}{88.95} & \multicolumn{1}{l|}{\multirow{4}{*}{88.84}} & \multicolumn{1}{l|}{0.86} & \multicolumn{1}{l|}{\multirow{4}{*}{0.87}} & \multicolumn{1}{l|}{0.88} & \multicolumn{1}{l|}{\multirow{4}{*}{0.88}} & \multicolumn{1}{l|}{0.87} & \multirow{4}{*}{0.87} \\ \cline{3-4} \cline{6-6} \cline{8-8} \cline{10-10} \cline{12-12} \cline{14-14} \cline{16-16} \cline{18-18}
                       &                                 & Grey                    & \multicolumn{1}{l|}{84.50} & \multicolumn{1}{l|}{}                       & \multicolumn{1}{l|}{0.87} & \multicolumn{1}{l|}{}                      & \multicolumn{1}{l|}{0.83} & \multicolumn{1}{l|}{}                      & \multicolumn{1}{l|}{0.85} &                       & \multicolumn{1}{l|}{86.38} & \multicolumn{1}{l|}{}                       & \multicolumn{1}{l|}{0.89} & \multicolumn{1}{l|}{}                      & \multicolumn{1}{l|}{0.85} & \multicolumn{1}{l|}{}                      & \multicolumn{1}{l|}{0.87} &                       \\ \cline{3-4} \cline{6-6} \cline{8-8} \cline{10-10} \cline{12-12} \cline{14-14} \cline{16-16} \cline{18-18}
                       &                                 & Red                     & \multicolumn{1}{l|}{86.87} & \multicolumn{1}{l|}{}                       & \multicolumn{1}{l|}{0.84} & \multicolumn{1}{l|}{}                      & \multicolumn{1}{l|}{0.83} & \multicolumn{1}{l|}{}                      & \multicolumn{1}{l|}{0.83} &                       & \multicolumn{1}{l|}{88.20} & \multicolumn{1}{l|}{}                       & \multicolumn{1}{l|}{0.87} & \multicolumn{1}{l|}{}                      & \multicolumn{1}{l|}{0.89} & \multicolumn{1}{l|}{}                      & \multicolumn{1}{l|}{0.88} &                       \\ \cline{3-4} \cline{6-6} \cline{8-8} \cline{10-10} \cline{12-12} \cline{14-14} \cline{16-16} \cline{18-18}
                       &                                 & Black                   & \multicolumn{1}{l|}{88.43} & \multicolumn{1}{l|}{}                       & \multicolumn{1}{l|}{0.82} & \multicolumn{1}{l|}{}                      & \multicolumn{1}{l|}{0.85} & \multicolumn{1}{l|}{}                      & \multicolumn{1}{l|}{0.83} &                       & \multicolumn{1}{l|}{91.83} & \multicolumn{1}{l|}{}                       & \multicolumn{1}{l|}{0.87} & \multicolumn{1}{l|}{}                      & \multicolumn{1}{l|}{0.90} & \multicolumn{1}{l|}{}                      & \multicolumn{1}{l|}{0.88} &                       \\ \cline{2-19} 
                       & \multirow{6}{*}{Intent of lie}  & Gaining Advantage       & \multicolumn{1}{l|}{87.62} & \multicolumn{1}{l|}{\multirow{6}{*}{83.69}} & \multicolumn{1}{l|}{0.85} & \multicolumn{1}{l|}{\multirow{6}{*}{0.84}} & \multicolumn{1}{l|}{0.83} & \multicolumn{1}{l|}{\multirow{6}{*}{0.79}} & \multicolumn{1}{l|}{0.84} & \multirow{6}{*}{0.81} & \multicolumn{1}{l|}{91.08} & \multicolumn{1}{l|}{\multirow{6}{*}{86.12}} & \multicolumn{1}{l|}{0.87} & \multicolumn{1}{l|}{\multirow{6}{*}{0.84}} & \multicolumn{1}{l|}{0.89} & \multicolumn{1}{l|}{\multirow{6}{*}{0.85}} & \multicolumn{1}{l|}{0.88} & \multirow{6}{*}{0.84} \\ \cline{3-4} \cline{6-6} \cline{8-8} \cline{10-10} \cline{12-12} \cline{14-14} \cline{16-16} \cline{18-18}
                       &                                 & Protecting Themselves   & \multicolumn{1}{l|}{84.87} & \multicolumn{1}{l|}{}                       & \multicolumn{1}{l|}{0.86} & \multicolumn{1}{l|}{}                      & \multicolumn{1}{l|}{0.81} & \multicolumn{1}{l|}{}                      & \multicolumn{1}{l|}{0.83} &                       & \multicolumn{1}{l|}{88.23} & \multicolumn{1}{l|}{}                       & \multicolumn{1}{l|}{0.84} & \multicolumn{1}{l|}{}                      & \multicolumn{1}{l|}{0.88} & \multicolumn{1}{l|}{}                      & \multicolumn{1}{l|}{0.86} &                       \\ \cline{3-4} \cline{6-6} \cline{8-8} \cline{10-10} \cline{12-12} \cline{14-14} \cline{16-16} \cline{18-18}
                       &                                 & Gaining Esteem          & \multicolumn{1}{l|}{82.97} & \multicolumn{1}{l|}{}                       & \multicolumn{1}{l|}{0.82} & \multicolumn{1}{l|}{}                      & \multicolumn{1}{l|}{0.77} & \multicolumn{1}{l|}{}                      & \multicolumn{1}{l|}{0.79} &                       & \multicolumn{1}{l|}{84.49} & \multicolumn{1}{l|}{}                       & \multicolumn{1}{l|}{0.85} & \multicolumn{1}{l|}{}                      & \multicolumn{1}{l|}{0.83} & \multicolumn{1}{l|}{}                      & \multicolumn{1}{l|}{0.84} &                       \\ \cline{3-4} \cline{6-6} \cline{8-8} \cline{10-10} \cline{12-12} \cline{14-14} \cline{16-16} \cline{18-18}
                       &                                 & Avoiding Embarrassment  & \multicolumn{1}{l|}{80.91} & \multicolumn{1}{l|}{}                       & \multicolumn{1}{l|}{0.84} & \multicolumn{1}{l|}{}                      & \multicolumn{1}{l|}{0.79} & \multicolumn{1}{l|}{}                      & \multicolumn{1}{l|}{0.81} &                       & \multicolumn{1}{l|}{82.97} & \multicolumn{1}{l|}{}                       & \multicolumn{1}{l|}{0.83} & \multicolumn{1}{l|}{}                      & \multicolumn{1}{l|}{0.80} & \multicolumn{1}{l|}{}                      & \multicolumn{1}{l|}{0.81} &                       \\ \cline{3-4} \cline{6-6} \cline{8-8} \cline{10-10} \cline{12-12} \cline{14-14} \cline{16-16} \cline{18-18}
                       &                                 & Defaming Esteem         & \multicolumn{1}{l|}{82.06} & \multicolumn{1}{l|}{}                       & \multicolumn{1}{l|}{0.83} & \multicolumn{1}{l|}{}                      & \multicolumn{1}{l|}{0.75} & \multicolumn{1}{l|}{}                      & \multicolumn{1}{l|}{0.79} &                       & \multicolumn{1}{l|}{83.87} & \multicolumn{1}{l|}{}                       & \multicolumn{1}{l|}{0.81} & \multicolumn{1}{l|}{}                      & \multicolumn{1}{l|}{0.84} & \multicolumn{1}{l|}{}                      & \multicolumn{1}{l|}{0.82} &                       \\ \cline{3-4} \cline{6-6} \cline{8-8} \cline{10-10} \cline{12-12} \cline{14-14} \cline{16-16} \cline{18-18}
                       &                                 & Protecting others       & \multicolumn{1}{l|}{80.11} & \multicolumn{1}{l|}{}                       & \multicolumn{1}{l|}{0.75} & \multicolumn{1}{l|}{}                      & \multicolumn{1}{l|}{0.79} & \multicolumn{1}{l|}{}                      & \multicolumn{1}{l|}{0.77} &                       & \multicolumn{1}{l|}{82.11} & \multicolumn{1}{l|}{}                       & \multicolumn{1}{l|}{0.79} & \multicolumn{1}{l|}{}                      & \multicolumn{1}{l|}{0.81} & \multicolumn{1}{l|}{}                      & \multicolumn{1}{l|}{0.8}  &                       \\ \cline{2-19} 
                       & \multirow{6}{*}{Topic of Lies} & Political               & \multicolumn{1}{l|}{88.70} & \multicolumn{1}{l|}{\multirow{6}{*}{83.60}} & \multicolumn{1}{l|}{0.82} & \multicolumn{1}{l|}{\multirow{6}{*}{0.81}} & \multicolumn{1}{l|}{0.86} & \multicolumn{1}{l|}{\multirow{6}{*}{0.82}} & \multicolumn{1}{l|}{0.84} & \multirow{6}{*}{0.81} & \multicolumn{1}{l|}{91.88} & \multicolumn{1}{l|}{\multirow{6}{*}{86.13}} & \multicolumn{1}{l|}{0.86} & \multicolumn{1}{l|}{\multirow{6}{*}{0.83}} & \multicolumn{1}{l|}{0.88} & \multicolumn{1}{l|}{\multirow{6}{*}{0.84}} & \multicolumn{1}{l|}{0.87} & \multirow{6}{*}{0.83} \\ \cline{3-4} \cline{6-6} \cline{8-8} \cline{10-10} \cline{12-12} \cline{14-14} \cline{16-16} \cline{18-18}
                       &                                 & Educational             & \multicolumn{1}{l|}{83.98} & \multicolumn{1}{l|}{}                       & \multicolumn{1}{l|}{0.84} & \multicolumn{1}{l|}{}                      & \multicolumn{1}{l|}{0.81} & \multicolumn{1}{l|}{}                      & \multicolumn{1}{l|}{0.82} &                       & \multicolumn{1}{l|}{86.79} & \multicolumn{1}{l|}{}                       & \multicolumn{1}{l|}{0.85} & \multicolumn{1}{l|}{}                      & \multicolumn{1}{l|}{0.86} & \multicolumn{1}{l|}{}                      & \multicolumn{1}{l|}{0.85} &                       \\ \cline{3-4} \cline{6-6} \cline{8-8} \cline{10-10} \cline{12-12} \cline{14-14} \cline{16-16} \cline{18-18}
                       &                                 & Regilious               & \multicolumn{1}{l|}{84.18} & \multicolumn{1}{l|}{}                       & \multicolumn{1}{l|}{0.81} & \multicolumn{1}{l|}{}                      & \multicolumn{1}{l|}{0.85} & \multicolumn{1}{l|}{}                      & \multicolumn{1}{l|}{0.83} &                       & \multicolumn{1}{l|}{84.98} & \multicolumn{1}{l|}{}                       & \multicolumn{1}{l|}{0.85} & \multicolumn{1}{l|}{}                      & \multicolumn{1}{l|}{0.83} & \multicolumn{1}{l|}{}                      & \multicolumn{1}{l|}{0.84} &                       \\ \cline{3-4} \cline{6-6} \cline{8-8} \cline{10-10} \cline{12-12} \cline{14-14} \cline{16-16} \cline{18-18}
                       &                                 & Ethnicity               & \multicolumn{1}{l|}{79.29} & \multicolumn{1}{l|}{}                       & \multicolumn{1}{l|}{0.83} & \multicolumn{1}{l|}{}                      & \multicolumn{1}{l|}{0.75} & \multicolumn{1}{l|}{}                      & \multicolumn{1}{l|}{0.79} &                       & \multicolumn{1}{l|}{83.84} & \multicolumn{1}{l|}{}                       & \multicolumn{1}{l|}{0.81} & \multicolumn{1}{l|}{}                      & \multicolumn{1}{l|}{0.82} & \multicolumn{1}{l|}{}                      & \multicolumn{1}{l|}{0.81} &                       \\ \cline{3-4} \cline{6-6} \cline{8-8} \cline{10-10} \cline{12-12} \cline{14-14} \cline{16-16} \cline{18-18}
                       &                                 & Racial                  & \multicolumn{1}{l|}{81.85} & \multicolumn{1}{l|}{}                       & \multicolumn{1}{l|}{0.77} & \multicolumn{1}{l|}{}                      & \multicolumn{1}{l|}{0.82} & \multicolumn{1}{l|}{}                      & \multicolumn{1}{l|}{0.79} &                       & \multicolumn{1}{l|}{83.16} & \multicolumn{1}{l|}{}                       & \multicolumn{1}{l|}{0.80} & \multicolumn{1}{l|}{}                      & \multicolumn{1}{l|}{0.79} & \multicolumn{1}{l|}{}                      & \multicolumn{1}{l|}{0.79} &                       \\ \cline{3-4} \cline{6-6} \cline{8-8} \cline{10-10} \cline{12-12} \cline{14-14} \cline{16-16} \cline{18-18}
                       &                                 & Other                   & \multicolumn{1}{l|}{76.95} & \multicolumn{1}{l|}{}                       & \multicolumn{1}{l|}{0.72} & \multicolumn{1}{l|}{}                      & \multicolumn{1}{l|}{0.77} & \multicolumn{1}{l|}{}                      & \multicolumn{1}{l|}{0.74} &                       & \multicolumn{1}{l|}{81.90} & \multicolumn{1}{l|}{}                       & \multicolumn{1}{l|}{0.76} & \multicolumn{1}{l|}{}                      & \multicolumn{1}{l|}{0.79} & \multicolumn{1}{l|}{}                      & \multicolumn{1}{l|}{0.77} &                       \\ 
\midrule
&                                  & Speculation              & \multicolumn{1}{l|}{85.67} & \multicolumn{1}{l|}{}                        & \multicolumn{1}{l|}{0.83} & \multicolumn{1}{l|}{}                       & \multicolumn{1}{l|}{0.81} & \multicolumn{1}{l|}{}                       & \multicolumn{1}{l|}{0.82} &                        & \multicolumn{1}{l|}{\cellcolor[HTML]{FFFE65}\textbf{89.91}} & \multicolumn{1}{l|}{\cellcolor[HTML]{FFFE65}}                                 & \multicolumn{1}{l|}{\cellcolor[HTML]{FFFE65}\textbf{0.86}} & \multicolumn{1}{l|}{\cellcolor[HTML]{FFFE65}}                                & \multicolumn{1}{l|}{\cellcolor[HTML]{FFFE65}\textbf{0.88}} & \multicolumn{1}{l|}{\cellcolor[HTML]{FFFE65}}                                & \multicolumn{1}{l|}{\cellcolor[HTML]{FFFE65}\textbf{0.87}} & \cellcolor[HTML]{FFFE65}                                \\ \cline{3-4} \cline{6-6} \cline{8-8} \cline{10-10} \cline{12-12} \cline{14-14} \cline{16-16} \cline{18-18}
                                                                                               &                                  & Opinion                  & \multicolumn{1}{l|}{83.40} & \multicolumn{1}{l|}{}                        & \multicolumn{1}{l|}{0.80} & \multicolumn{1}{l|}{}                       & \multicolumn{1}{l|}{0.82} & \multicolumn{1}{l|}{}                       & \multicolumn{1}{l|}{0.81} &                        & \multicolumn{1}{l|}{\cellcolor[HTML]{FFFE65}\textbf{87.09}} & \multicolumn{1}{l|}{\cellcolor[HTML]{FFFE65}}                                 & \multicolumn{1}{l|}{\cellcolor[HTML]{FFFE65}\textbf{0.84}} & \multicolumn{1}{l|}{\cellcolor[HTML]{FFFE65}}                                & \multicolumn{1}{l|}{\cellcolor[HTML]{FFFE65}\textbf{0.83}} & \multicolumn{1}{l|}{\cellcolor[HTML]{FFFE65}}                                & \multicolumn{1}{l|}{\cellcolor[HTML]{FFFE65}\textbf{0.83}} & \cellcolor[HTML]{FFFE65}                                \\ \cline{3-4} \cline{6-6} \cline{8-8} \cline{10-10} \cline{12-12} \cline{14-14} \cline{16-16} \cline{18-18}
                                                                                               &                                  & Bais                     & \multicolumn{1}{l|}{76.30} & \multicolumn{1}{l|}{}                        & \multicolumn{1}{l|}{0.77} & \multicolumn{1}{l|}{}                       & \multicolumn{1}{l|}{0.75} & \multicolumn{1}{l|}{}                       & \multicolumn{1}{l|}{0.76} &                        & \multicolumn{1}{l|}{\cellcolor[HTML]{FFFE65}\textbf{80.49}} & \multicolumn{1}{l|}{\cellcolor[HTML]{FFFE65}}                                 & \multicolumn{1}{l|}{\cellcolor[HTML]{FFFE65}\textbf{0.79}} & \multicolumn{1}{l|}{\cellcolor[HTML]{FFFE65}}                                & \multicolumn{1}{l|}{\cellcolor[HTML]{FFFE65}\textbf{0.83}} & \multicolumn{1}{l|}{\cellcolor[HTML]{FFFE65}}                                & \multicolumn{1}{l|}{\cellcolor[HTML]{FFFE65}\textbf{0.81}} & \cellcolor[HTML]{FFFE65}                                \\ \cline{3-4} \cline{6-6} \cline{8-8} \cline{10-10} \cline{12-12} \cline{14-14} \cline{16-16} \cline{18-18}
                                                                                               &                                  & Distortion               & \multicolumn{1}{l|}{80.44} & \multicolumn{1}{l|}{}                        & \multicolumn{1}{l|}{0.81} & \multicolumn{1}{l|}{}                       & \multicolumn{1}{l|}{0.79} & \multicolumn{1}{l|}{}                       & \multicolumn{1}{l|}{0.8}  &                        & \multicolumn{1}{l|}{\cellcolor[HTML]{FFFE65}\textbf{85.77}} & \multicolumn{1}{l|}{\cellcolor[HTML]{FFFE65}}                                 & \multicolumn{1}{l|}{\cellcolor[HTML]{FFFE65}\textbf{0.83}} & \multicolumn{1}{l|}{\cellcolor[HTML]{FFFE65}}                                & \multicolumn{1}{l|}{\cellcolor[HTML]{FFFE65}\textbf{0.85}} & \multicolumn{1}{l|}{\cellcolor[HTML]{FFFE65}}                                & \multicolumn{1}{l|}{\cellcolor[HTML]{FFFE65}\textbf{0.84}} & \cellcolor[HTML]{FFFE65}                                \\ \cline{3-4} \cline{6-6} \cline{8-8} \cline{10-10} \cline{12-12} \cline{14-14} \cline{16-16} \cline{18-18}
                                                                                               & \multirow{-5}{*}{Type of Omission}     & Sound Factual            & \multicolumn{1}{l|}{85.32} & \multicolumn{1}{l|}{\multirow{-5}{*}{82.22}} & \multicolumn{1}{l|}{0.84} & \multicolumn{1}{l|}{\multirow{-5}{*}{0.81}} & \multicolumn{1}{l|}{0.80} & \multicolumn{1}{l|}{\multirow{-5}{*}{0.79}} & \multicolumn{1}{l|}{0.82} & \multirow{-5}{*}{0.80} & \multicolumn{1}{l|}{\cellcolor[HTML]{FFFE65}\textbf{88.23}} & \multicolumn{1}{l|}{\multirow{-5}{*}{\cellcolor[HTML]{FFFE65}\textbf{86.30}}} & \multicolumn{1}{l|}{\cellcolor[HTML]{FFFE65}\textbf{0.86}} & \multicolumn{1}{l|}{\multirow{-5}{*}{\cellcolor[HTML]{FFFE65}\textbf{0.84}}} & \multicolumn{1}{l|}{\cellcolor[HTML]{FFFE65}\textbf{0.89}} & \multicolumn{1}{l|}{\multirow{-5}{*}{\cellcolor[HTML]{FFFE65}\textbf{0.86}}} & \multicolumn{1}{l|}{\cellcolor[HTML]{FFFE65}\textbf{0.87}} & \multirow{-5}{*}{\cellcolor[HTML]{FFFE65}\textbf{0.84}} \\ \cline{2-19} 
                                                                                               &                                  & White                    & \multicolumn{1}{l|}{87.36} & \multicolumn{1}{l|}{}                        & \multicolumn{1}{l|}{0.88} & \multicolumn{1}{l|}{}                       & \multicolumn{1}{l|}{0.86} & \multicolumn{1}{l|}{}                       & \multicolumn{1}{l|}{0.87} &                        & \multicolumn{1}{l|}{\cellcolor[HTML]{FFFE65}\textbf{91.23}} & \multicolumn{1}{l|}{\cellcolor[HTML]{FFFE65}}                                 & \multicolumn{1}{l|}{\cellcolor[HTML]{FFFE65}\textbf{0.90}} & \multicolumn{1}{l|}{\cellcolor[HTML]{FFFE65}}                                & \multicolumn{1}{l|}{\cellcolor[HTML]{FFFE65}\textbf{0.89}} & \multicolumn{1}{l|}{\cellcolor[HTML]{FFFE65}}                                & \multicolumn{1}{l|}{\cellcolor[HTML]{FFFE65}\textbf{0.90}} & \cellcolor[HTML]{FFFE65}                                \\ \cline{3-4} \cline{6-6} \cline{8-8} \cline{10-10} \cline{12-12} \cline{14-14} \cline{16-16} \cline{18-18}
                                                                                               &                                  & Grey                     & \multicolumn{1}{l|}{89.05} & \multicolumn{1}{l|}{}                        & \multicolumn{1}{l|}{0.88} & \multicolumn{1}{l|}{}                       & \multicolumn{1}{l|}{0.84} & \multicolumn{1}{l|}{}                       & \multicolumn{1}{l|}{0.86} &                        & \multicolumn{1}{l|}{\cellcolor[HTML]{FFFE65}\textbf{94.53}} & \multicolumn{1}{l|}{\cellcolor[HTML]{FFFE65}}                                 & \multicolumn{1}{l|}{\cellcolor[HTML]{FFFE65}\textbf{0.92}} & \multicolumn{1}{l|}{\cellcolor[HTML]{FFFE65}}                                & \multicolumn{1}{l|}{\cellcolor[HTML]{FFFE65}\textbf{0.88}} & \multicolumn{1}{l|}{\cellcolor[HTML]{FFFE65}}                                & \multicolumn{1}{l|}{\cellcolor[HTML]{FFFE65}\textbf{0.90}} & \cellcolor[HTML]{FFFE65}                                \\ \cline{3-4} \cline{6-6} \cline{8-8} \cline{10-10} \cline{12-12} \cline{14-14} \cline{16-16} \cline{18-18}
                                                                                               &                                  & Red                      & \multicolumn{1}{l|}{88.41} & \multicolumn{1}{l|}{}                        & \multicolumn{1}{l|}{0.86} & \multicolumn{1}{l|}{}                       & \multicolumn{1}{l|}{0.85} & \multicolumn{1}{l|}{}                       & \multicolumn{1}{l|}{0.85} &                        & \multicolumn{1}{l|}{\cellcolor[HTML]{FFFE65}\textbf{93.45}} & \multicolumn{1}{l|}{\cellcolor[HTML]{FFFE65}}                                 & \multicolumn{1}{l|}{\cellcolor[HTML]{FFFE65}\textbf{0.91}} & \multicolumn{1}{l|}{\cellcolor[HTML]{FFFE65}}                                & \multicolumn{1}{l|}{\cellcolor[HTML]{FFFE65}\textbf{0.92}} & \multicolumn{1}{l|}{\cellcolor[HTML]{FFFE65}}                                & \multicolumn{1}{l|}{\cellcolor[HTML]{FFFE65}\textbf{0.92}} & \cellcolor[HTML]{FFFE65}                                \\ \cline{3-4} \cline{6-6} \cline{8-8} \cline{10-10} \cline{12-12} \cline{14-14} \cline{16-16} \cline{18-18}
                                                                                               & \multirow{-4}{*}{Color of Lie}   & Black                    & \multicolumn{1}{l|}{91.62} & \multicolumn{1}{l|}{\multirow{-4}{*}{89.11}} & \multicolumn{1}{l|}{0.89} & \multicolumn{1}{l|}{\multirow{-4}{*}{0.88}} & \multicolumn{1}{l|}{0.85} & \multicolumn{1}{l|}{\multirow{-4}{*}{0.85}} & \multicolumn{1}{l|}{0.87} & \multirow{-4}{*}{0.86} & \multicolumn{1}{l|}{\cellcolor[HTML]{FFFE65}\textbf{96.17}} & \multicolumn{1}{l|}{\multirow{-4}{*}{\cellcolor[HTML]{FFFE65}\textbf{93.84}}} & \multicolumn{1}{l|}{\cellcolor[HTML]{FFFE65}\textbf{0.94}} & \multicolumn{1}{l|}{\multirow{-4}{*}{\cellcolor[HTML]{FFFE65}\textbf{0.92}}} & \multicolumn{1}{l|}{\cellcolor[HTML]{FFFE65}\textbf{0.93}} & \multicolumn{1}{l|}{\multirow{-4}{*}{\cellcolor[HTML]{FFFE65}\textbf{0.91}}} & \multicolumn{1}{l|}{\cellcolor[HTML]{FFFE65}\textbf{0.94}} & \multirow{-4}{*}{\cellcolor[HTML]{FFFE65}\textbf{0.92}} \\ \cline{2-19} 
                                                                                               &                                  & Gaining Advantage        & \multicolumn{1}{l|}{89.35} & \multicolumn{1}{l|}{}                        & \multicolumn{1}{l|}{0.88} & \multicolumn{1}{l|}{}                       & \multicolumn{1}{l|}{0.86} & \multicolumn{1}{l|}{}                       & \multicolumn{1}{l|}{0.87} &                        & \multicolumn{1}{l|}{\cellcolor[HTML]{FFFE65}\textbf{92.54}} & \multicolumn{1}{l|}{\cellcolor[HTML]{FFFE65}}                                 & \multicolumn{1}{l|}{\cellcolor[HTML]{FFFE65}\textbf{0.91}} & \multicolumn{1}{l|}{\cellcolor[HTML]{FFFE65}}                                & \multicolumn{1}{l|}{\cellcolor[HTML]{FFFE65}\textbf{0.93}} & \multicolumn{1}{l|}{\cellcolor[HTML]{FFFE65}}                                & \multicolumn{1}{l|}{\cellcolor[HTML]{FFFE65}\textbf{0.92}} & \cellcolor[HTML]{FFFE65}                                \\ \cline{3-4} \cline{6-6} \cline{8-8} \cline{10-10} \cline{12-12} \cline{14-14} \cline{16-16} \cline{18-18}
                                                                                               &                                  & Protecting Themselves    & \multicolumn{1}{l|}{88.74} & \multicolumn{1}{l|}{}                        & \multicolumn{1}{l|}{0.86} & \multicolumn{1}{l|}{}                       & \multicolumn{1}{l|}{0.85} & \multicolumn{1}{l|}{}                       & \multicolumn{1}{l|}{0.85} &                        & \multicolumn{1}{l|}{\cellcolor[HTML]{FFFE65}\textbf{90.78}} & \multicolumn{1}{l|}{\cellcolor[HTML]{FFFE65}}                                 & \multicolumn{1}{l|}{\cellcolor[HTML]{FFFE65}\textbf{0.89}} & \multicolumn{1}{l|}{\cellcolor[HTML]{FFFE65}}                                & \multicolumn{1}{l|}{\cellcolor[HTML]{FFFE65}\textbf{0.90}} & \multicolumn{1}{l|}{\cellcolor[HTML]{FFFE65}}                                & \multicolumn{1}{l|}{\cellcolor[HTML]{FFFE65}\textbf{0.89}} & \cellcolor[HTML]{FFFE65}                                \\ \cline{3-4} \cline{6-6} \cline{8-8} \cline{10-10} \cline{12-12} \cline{14-14} \cline{16-16} \cline{18-18}
                                                                                               &                                  & Gaining Esteem           & \multicolumn{1}{l|}{85.67} & \multicolumn{1}{l|}{}                        & \multicolumn{1}{l|}{0.85} & \multicolumn{1}{l|}{}                       & \multicolumn{1}{l|}{0.82} & \multicolumn{1}{l|}{}                       & \multicolumn{1}{l|}{0.83} &                        & \multicolumn{1}{l|}{\cellcolor[HTML]{FFFE65}\textbf{88.56}} & \multicolumn{1}{l|}{\cellcolor[HTML]{FFFE65}}                                 & \multicolumn{1}{l|}{\cellcolor[HTML]{FFFE65}\textbf{0.88}} & \multicolumn{1}{l|}{\cellcolor[HTML]{FFFE65}}                                & \multicolumn{1}{l|}{\cellcolor[HTML]{FFFE65}\textbf{0.86}} & \multicolumn{1}{l|}{\cellcolor[HTML]{FFFE65}}                                & \multicolumn{1}{l|}{\cellcolor[HTML]{FFFE65}\textbf{0.87}} & \cellcolor[HTML]{FFFE65}                                \\ \cline{3-4} \cline{6-6} \cline{8-8} \cline{10-10} \cline{12-12} \cline{14-14} \cline{16-16} \cline{18-18}
                                                                                               &                                  & Avoiding Embarrassment   & \multicolumn{1}{l|}{83.25} & \multicolumn{1}{l|}{}                        & \multicolumn{1}{l|}{0.82} & \multicolumn{1}{l|}{}                       & \multicolumn{1}{l|}{0.83} & \multicolumn{1}{l|}{}                       & \multicolumn{1}{l|}{0.82} &                        & \multicolumn{1}{l|}{\cellcolor[HTML]{FFFE65}\textbf{87.19}} & \multicolumn{1}{l|}{\cellcolor[HTML]{FFFE65}}                                 & \multicolumn{1}{l|}{\cellcolor[HTML]{FFFE65}\textbf{0.85}} & \multicolumn{1}{l|}{\cellcolor[HTML]{FFFE65}}                                & \multicolumn{1}{l|}{\cellcolor[HTML]{FFFE65}\textbf{0.88}} & \multicolumn{1}{l|}{\cellcolor[HTML]{FFFE65}}                                & \multicolumn{1}{l|}{\cellcolor[HTML]{FFFE65}\textbf{0.86}} & \cellcolor[HTML]{FFFE65}                                \\ \cline{3-4} \cline{6-6} \cline{8-8} \cline{10-10} \cline{12-12} \cline{14-14} \cline{16-16} \cline{18-18}
                                                                                               &                                  & Defaming Esteem          & \multicolumn{1}{l|}{83.46} & \multicolumn{1}{l|}{}                        & \multicolumn{1}{l|}{0.83} & \multicolumn{1}{l|}{}                       & \multicolumn{1}{l|}{0.82} & \multicolumn{1}{l|}{}                       & \multicolumn{1}{l|}{0.82} &                        & \multicolumn{1}{l|}{\cellcolor[HTML]{FFFE65}\textbf{86.88}} & \multicolumn{1}{l|}{\cellcolor[HTML]{FFFE65}}                                 & \multicolumn{1}{l|}{\cellcolor[HTML]{FFFE65}\textbf{0.85}} & \multicolumn{1}{l|}{\cellcolor[HTML]{FFFE65}}                                & \multicolumn{1}{l|}{\cellcolor[HTML]{FFFE65}\textbf{0.84}} & \multicolumn{1}{l|}{\cellcolor[HTML]{FFFE65}}                                & \multicolumn{1}{l|}{\cellcolor[HTML]{FFFE65}\textbf{0.84}} & \cellcolor[HTML]{FFFE65}                                \\ \cline{3-4} \cline{6-6} \cline{8-8} \cline{10-10} \cline{12-12} \cline{14-14} \cline{16-16} \cline{18-18}
                                                                                               & \multirow{-6}{*}{Intent of lie}  & Protecting others        & \multicolumn{1}{l|}{81.16} & \multicolumn{1}{l|}{\multirow{-6}{*}{86.09}} & \multicolumn{1}{l|}{0.80} & \multicolumn{1}{l|}{\multirow{-6}{*}{0.85}} & \multicolumn{1}{l|}{0.79} & \multicolumn{1}{l|}{\multirow{-6}{*}{0.84}} & \multicolumn{1}{l|}{0.79} & \multirow{-6}{*}{0.84} & \multicolumn{1}{l|}{\cellcolor[HTML]{FFFE65}\textbf{85.04}} & \multicolumn{1}{l|}{\multirow{-6}{*}{\cellcolor[HTML]{FFFE65}\textbf{88.49}}} & \multicolumn{1}{l|}{\cellcolor[HTML]{FFFE65}\textbf{0.83}} & \multicolumn{1}{l|}{\multirow{-6}{*}{\cellcolor[HTML]{FFFE65}\textbf{0.87}}} & \multicolumn{1}{l|}{\cellcolor[HTML]{FFFE65}\textbf{0.84}} & \multicolumn{1}{l|}{\multirow{-6}{*}{\cellcolor[HTML]{FFFE65}\textbf{0.88}}} & \multicolumn{1}{l|}{\cellcolor[HTML]{FFFE65}\textbf{0.83}} & \multirow{-6}{*}{\cellcolor[HTML]{FFFE65}\textbf{0.87}} \\ \cline{2-19} 
                                                                                               &                                  & Political                & \multicolumn{1}{l|}{90.59} & \multicolumn{1}{l|}{}                        & \multicolumn{1}{l|}{0.88} & \multicolumn{1}{l|}{}                       & \multicolumn{1}{l|}{0.86} & \multicolumn{1}{l|}{}                       & \multicolumn{1}{l|}{0.87} &                        & \multicolumn{1}{l|}{\cellcolor[HTML]{FFFE65}\textbf{94.16}} & \multicolumn{1}{l|}{\cellcolor[HTML]{FFFE65}}                                 & \multicolumn{1}{l|}{\cellcolor[HTML]{FFFE65}\textbf{0.93}} & \multicolumn{1}{l|}{\cellcolor[HTML]{FFFE65}}                                & \multicolumn{1}{l|}{\cellcolor[HTML]{FFFE65}\textbf{0.90}} & \multicolumn{1}{l|}{\cellcolor[HTML]{FFFE65}}                                & \multicolumn{1}{l|}{\cellcolor[HTML]{FFFE65}\textbf{0.91}} & \cellcolor[HTML]{FFFE65}                                \\ \cline{3-4} \cline{6-6} \cline{8-8} \cline{10-10} \cline{12-12} \cline{14-14} \cline{16-16} \cline{18-18}
                                                                                               &                                  & Educational              & \multicolumn{1}{l|}{86.77} & \multicolumn{1}{l|}{}                        & \multicolumn{1}{l|}{0.87} & \multicolumn{1}{l|}{}                       & \multicolumn{1}{l|}{0.88} & \multicolumn{1}{l|}{}                       & \multicolumn{1}{l|}{0.87} &                        & \multicolumn{1}{l|}{\cellcolor[HTML]{FFFE65}\textbf{90.66}} & \multicolumn{1}{l|}{\cellcolor[HTML]{FFFE65}}                                 & \multicolumn{1}{l|}{\cellcolor[HTML]{FFFE65}\textbf{0.90}} & \multicolumn{1}{l|}{\cellcolor[HTML]{FFFE65}}                                & \multicolumn{1}{l|}{\cellcolor[HTML]{FFFE65}\textbf{0.87}} & \multicolumn{1}{l|}{\cellcolor[HTML]{FFFE65}}                                & \multicolumn{1}{l|}{\cellcolor[HTML]{FFFE65}\textbf{0.88}} & \cellcolor[HTML]{FFFE65}                                \\ \cline{3-4} \cline{6-6} \cline{8-8} \cline{10-10} \cline{12-12} \cline{14-14} \cline{16-16} \cline{18-18}
                                                                                               &                                  & Regilious                & \multicolumn{1}{l|}{85.46} & \multicolumn{1}{l|}{}                        & \multicolumn{1}{l|}{0.84} & \multicolumn{1}{l|}{}                       & \multicolumn{1}{l|}{0.84} & \multicolumn{1}{l|}{}                       & \multicolumn{1}{l|}{0.84} &                        & \multicolumn{1}{l|}{\cellcolor[HTML]{FFFE65}\textbf{87.83}} & \multicolumn{1}{l|}{\cellcolor[HTML]{FFFE65}}                                 & \multicolumn{1}{l|}{\cellcolor[HTML]{FFFE65}\textbf{0.87}} & \multicolumn{1}{l|}{\cellcolor[HTML]{FFFE65}}                                & \multicolumn{1}{l|}{\cellcolor[HTML]{FFFE65}\textbf{0.85}} & \multicolumn{1}{l|}{\cellcolor[HTML]{FFFE65}}                                & \multicolumn{1}{l|}{\cellcolor[HTML]{FFFE65}\textbf{0.86}} & \cellcolor[HTML]{FFFE65}                                \\ \cline{3-4} \cline{6-6} \cline{8-8} \cline{10-10} \cline{12-12} \cline{14-14} \cline{16-16} \cline{18-18}
                                                                                               &                                  & Ethnicity                & \multicolumn{1}{l|}{84.69} & \multicolumn{1}{l|}{}                        & \multicolumn{1}{l|}{0.84} & \multicolumn{1}{l|}{}                       & \multicolumn{1}{l|}{0.85} & \multicolumn{1}{l|}{}                       & \multicolumn{1}{l|}{0.84} &                        & \multicolumn{1}{l|}{\cellcolor[HTML]{FFFE65}\textbf{88.67}} & \multicolumn{1}{l|}{\cellcolor[HTML]{FFFE65}}                                 & \multicolumn{1}{l|}{\cellcolor[HTML]{FFFE65}\textbf{0.86}} & \multicolumn{1}{l|}{\cellcolor[HTML]{FFFE65}}                                & \multicolumn{1}{l|}{\cellcolor[HTML]{FFFE65}\textbf{0.87}} & \multicolumn{1}{l|}{\cellcolor[HTML]{FFFE65}}                                & \multicolumn{1}{l|}{\cellcolor[HTML]{FFFE65}\textbf{0.86}} & \cellcolor[HTML]{FFFE65}                                \\ \cline{3-4} \cline{6-6} \cline{8-8} \cline{10-10} \cline{12-12} \cline{14-14} \cline{16-16} \cline{18-18}
                                                                                               &                                  & Racial                   & \multicolumn{1}{l|}{81.84} & \multicolumn{1}{l|}{}                        & \multicolumn{1}{l|}{0.83} & \multicolumn{1}{l|}{}                       & \multicolumn{1}{l|}{0.82} & \multicolumn{1}{l|}{}                       & \multicolumn{1}{l|}{0.82} &                        & \multicolumn{1}{l|}{\cellcolor[HTML]{FFFE65}\textbf{85.89}} & \multicolumn{1}{l|}{\cellcolor[HTML]{FFFE65}}                                 & \multicolumn{1}{l|}{\cellcolor[HTML]{FFFE65}\textbf{0.87}} & \multicolumn{1}{l|}{\cellcolor[HTML]{FFFE65}}                                & \multicolumn{1}{l|}{\cellcolor[HTML]{FFFE65}\textbf{0.84}} & \multicolumn{1}{l|}{\cellcolor[HTML]{FFFE65}}                                & \multicolumn{1}{l|}{\cellcolor[HTML]{FFFE65}\textbf{0.85}} & \cellcolor[HTML]{FFFE65}                                \\ \cline{3-4} \cline{6-6} \cline{8-8} \cline{10-10} \cline{12-12} \cline{14-14} \cline{16-16} \cline{18-18}
\multirow{-21}{*}{\textbf{\begin{tabular}[c]{@{}c@{}}T5 with\\ Transformer \\ Encoder\end{tabular}}}
 & \multirow{-6}{*}{Topic of Lies} & Other                    & \multicolumn{1}{l|}{79.18} & \multicolumn{1}{l|}{\multirow{-6}{*}{85.87}} & \multicolumn{1}{l|}{0.78} & \multicolumn{1}{l|}{\multirow{-6}{*}{0.85}} & \multicolumn{1}{l|}{0.78} & \multicolumn{1}{l|}{\multirow{-6}{*}{0.85}} & \multicolumn{1}{l|}{0.78} & \multirow{-6}{*}{0.85} & \multicolumn{1}{l|}{\cellcolor[HTML]{FFFE65}\textbf{82.34}} & \multicolumn{1}{l|}{\multirow{-6}{*}{\cellcolor[HTML]{FFFE65}\textbf{88.26}}} & \multicolumn{1}{l|}{\cellcolor[HTML]{FFFE65}\textbf{0.84}} & \multicolumn{1}{l|}{\multirow{-6}{*}{\cellcolor[HTML]{FFFE65}\textbf{0.87}}} & \multicolumn{1}{l|}{\cellcolor[HTML]{FFFE65}\textbf{0.81}} & \multicolumn{1}{l|}{\multirow{-6}{*}{\cellcolor[HTML]{FFFE65}\textbf{0.86}}} & \multicolumn{1}{l|}{\cellcolor[HTML]{FFFE65}\textbf{0.82}} & \multirow{-6}{*}{\cellcolor[HTML]{FFFE65}\textbf{0.86}} \\ \hline
                    
\end{tabular}
}
\caption{Experiment results: The table showcases the results obtained from different experiments using varying encoder architectures, namely LSTM and Transformer. The term "Without Model Merging" refers to the utilization of the T5-3b model without any fine-tuning. Conversely, the term "With Model Merging" signifies the fine-tuning of four T5 models, each corresponding to a distinct layer, followed by Dataless Knowledge fusion. \cite{jin2022dataless}}\
\label{tab:overall_exp}
\end{table}

%% file: table/Appendix_PT.tex
\begin{tcbraster}[raster columns=2,raster equal height]
\begin{tcolorbox}[nobeforeafter, title=box 1,colback=brown!5!white,colframe=brown!75!black,title=\footnotesize{\textbf{\textsc{Propaganda Technique Definition}}}]
\begin{itemize}
[leftmargin=1mm]
\setlength\itemsep{0em}
\begin{spacing}{0.90}
\vspace{-2.2mm}
\item[\ding{224}] {\footnotesize \fontfamily{phv}\fontsize{8}{9}\selectfont{\textbf{Flag Waving:} Playing on strong national feeling (or to any group, e.g., race, gender, etc) to justify or promote an action or an idea.
}}
\vspace{-2.2mm}
\item[\ding{224}] {\footnotesize 
{\fontfamily{phv}\fontsize{8}{9}\selectfont{\textbf{Slogans:}A brief and striking phrase that may include labeling and stereotyping.
}
}}

\vspace{-2.2mm}
\item[\ding{224}] {\footnotesize 
{\fontfamily{phv}\fontsize{8}{9}\selectfont
{\textbf{Appeal to fear - prejudices:}Seeking to build support for an idea by instilling anxiety and/or panic in the population towards an alternative.}
}}
\vspace{-2.2mm}
\item[\ding{224}] {\footnotesize 
{\fontfamily{phv}\fontsize{8}{9}\selectfont
{\textbf{Exaggeration-Minimization}: Either representing something in an excessive manner: making things larger, better, worse (e.g., the best of the best) or making something seem less important or smaller than it really is (e.g., saying that an insult was actually just a joke).}
}}

\vspace{-2.2mm}
\item[\ding{224}] {\footnotesize 
{\fontfamily{phv}\fontsize{8}{9}\selectfont
{\textbf{Repetition:} Repeating the same message over and over again so that the audience will eventually accept it.}
}}

\vspace{-2.2mm}
\item[\ding{224}] {\footnotesize 
{\fontfamily{phv}\fontsize{8}{9}\selectfont
{\textbf{Name Calling Labelling:} Labeling the object of the propaganda campaign as something that the target audience fears, hates, finds undesirable, or loves or praises. }
}}

\vspace{-2.2mm}
\item[\ding{224}] {\footnotesize 
{\fontfamily{phv}\fontsize{8}{9}\selectfont
{\textbf{Bandwagon:} Attempting to persuade the target audience to join in and take the course of action because “everyone else is taking the same action.”}
}}

\vspace{-2.2mm}
\item[\ding{224}] {\footnotesize 
{\fontfamily{phv}\fontsize{8}{9}\selectfont
{\textbf{Loaded Language:} Using specific words and phrases with strong emotional implications (either positive or negative) to influence an audience. }
}}

\vspace{-2.2mm}
\item[\ding{224}] {\footnotesize 
{\fontfamily{phv}\fontsize{8}{9}\selectfont
{\textbf{Casual Oversimplification:} Assuming a single cause or reason when there are actually multiple causes for an issue.}
}}

\vspace{-2.2mm}
\item[\ding{224}] {\footnotesize 
{\fontfamily{phv}\fontsize{8}{9}\selectfont
{\textbf{Red herring:} Introducing irrelevant material to the issue being discussed so that everyone’s attention is diverted away from the points made.}
}}

\vspace{-2.2mm}
\item[\ding{224}] {\footnotesize 
{\fontfamily{phv}\fontsize{8}{9}\selectfont
{\textbf{Appeal to authority:} Stating that a claim is true simply because a valid authority or expert on the issue said it was true.}
}}

\vspace{-2.2mm}
\item[\ding{224}] {\footnotesize 
{\fontfamily{phv}\fontsize{8}{9}\selectfont
{\textbf{Thought terminating cliches:} Words or phrases that discourage critical thought and meaningful discussion about a given topic.}
}}

\vspace{-2.2mm}
\item[\ding{224}] {\footnotesize 
{\fontfamily{phv}\fontsize{8}{9}\selectfont
{\textbf{Whataboutism:} A technique that attempts to discredit an opponent’s position by charging them with hypocrisy without directly disproving their argument.}
}}

\vspace{-6mm}
\end{spacing}
\end{itemize}


\end{tcolorbox}
\begin{tcolorbox}[ nobeforeafter, title=box 2, colback=teal!5!white,colframe=teal!75!black,title=\footnotesize{\textbf{\textsc{Propaganda through Deception}}}]
\begin{itemize}
[leftmargin=1mm]
\setlength\itemsep{0em}
\begin{spacing}{0.90}
\vspace{-2.2mm}
\item[\ding{224}] {\footnotesize \fontfamily{phv}\fontsize{8}{9}\selectfont{\textbf{Flag Waving:} Flag waving maps to speculation in layer 1, black lies in layer 2, gaining advantage in layer 3, and religious aspects in layer 4.
}}
\vspace{-2.2mm}
\item[\ding{224}] {\footnotesize 
{\fontfamily{phv}\fontsize{8}{9}\selectfont{\textbf{Slogans:} This technique is mostly mapped with speculation in layer1, white lie in layer 2, political in layer 3 and gaining advantage in layer 4.
}
}}

\vspace{-2.2mm}
\item[\ding{224}] {\footnotesize 
{\fontfamily{phv}\fontsize{8}{9}\selectfont
{\textbf{Appeal to fear - prejudices:} This technqiue primarily corresponds to speculation in layer 1, black lie in layer 2, political in layer 3 and gaining advantage in layer 4.}
}}

\vspace{-2.2mm}
\item[\ding{224}] {\footnotesize 
{\fontfamily{phv}\fontsize{8}{9}\selectfont
{\textbf{Exaggeration-Minimization}: In the Layers of Omission, Exaggeration or Minimization is mostly mapped to speculation in layer 1, black lie in layer 2, political in layer 3 and gaining advantage in layer 4.}
}}

\vspace{-2.2mm}
\item[\ding{224}] {\footnotesize 
{\fontfamily{phv}\fontsize{8}{9}\selectfont
{\textbf{Repetition:} Repetition is mostly mapped to Speculation, Black lie, intention of gaining advantage and in political influence.}
}}

\vspace{-2.2mm}
\item[\ding{224}] {\footnotesize 
{\fontfamily{phv}\fontsize{8}{9}\selectfont
{\textbf{Name Calling Labelling:} Name Calling or Labelling is largely mapped to speculation in layer 1, black lie in layer 2, gaining advantage in layer 3 and political in layer 4.}
}}

\vspace{-2.2mm}
\item[\ding{224}] {\footnotesize 
{\fontfamily{phv}\fontsize{8}{9}\selectfont
{\textbf{Bandwagon:} Bandwagon is mostly mapped to speculation in layer 1. It is mapped with both white and gray lie in layer 2. It is mapped with protecting oneself in layer 3 and education in layer 4. }
}}

\vspace{-2.2mm}
\item[\ding{224}] {\footnotesize 
{\fontfamily{phv}\fontsize{8}{9}\selectfont
{\textbf{Loaded Language:} Loaded Language is mapped mostly with speculation in layer 1, black lie in layer 2, gaining advantage in layer 3 and political in layer 4. }
}}

\vspace{-2.2mm}
\item[\ding{224}] {\footnotesize 
{\fontfamily{phv}\fontsize{8}{9}\selectfont
{\textbf{Casual Oversimplification:} Causal Oversimplification is mapped mostly with speculation in layer 1, with black lie and in some cases with red lie in layer 2, gaining advantage in layer 3 and political in layer 4. }
}}

\vspace{-2.2mm}
\item[\ding{224}] {\footnotesize 
{\fontfamily{phv}\fontsize{8}{9}\selectfont
{\textbf{Red herring:} In layer 1, Red Herring corresponds to both speculation and opinion. Layer 2 primarily associates it with black lies, occasionally with white lies. In layer 3, it largely aligns with gaining advantage, while layer 4 relates to political aspects.}
}}

\vspace{-2.2mm}
\item[\ding{224}] {\footnotesize 
{\fontfamily{phv}\fontsize{8}{9}\selectfont
{\textbf{Appeal to authority:} This technique largely maps with opinion and with speculation too. In the 2nd layer, it maps with black and gray lies and with gaining advantage in 3rd layer and political in 4th layer. }
}}

\vspace{-2.2mm}
\item[\ding{224}] {\footnotesize 
{\fontfamily{phv}\fontsize{8}{9}\selectfont
{\textbf{Thought terminating cliches:} This technique mostly maps with speculation in layer 1, gray and black lie in layer 2, gaining advantage in layer 3 and political in layer 4.}
}}

\vspace{-2.2mm}
\item[\ding{224}] {\footnotesize 
{\fontfamily{phv}\fontsize{8}{9}\selectfont
{\textbf{Whataboutism:} Whataboutism mostly maps with speculation in layer 1, black lie in layer 2, gaining advantage in layer 3 and political in layer 4. }
}}

\vspace{-6mm}
\end{spacing}
\end{itemize}

\end{tcolorbox}
\end{tcbraster}

%% file: table/Appendix_PT2.tex
\begin{tcbraster}[raster columns=2,raster equal height]
\begin{tcolorbox}[nobeforeafter, title=box 1,colback=brown!5!white,colframe=brown!75!black,title=\footnotesize{\textbf{\textsc{Propaganda Technique Definition}}}]
\begin{itemize}
[leftmargin=1mm]
\setlength\itemsep{0em}
\begin{spacing}{0.90}

\item[\ding{224}] {\footnotesize 
{\fontfamily{phv}\fontsize{8}{9}\selectfont
{\textbf{Straw Men:}Substituting an opponent’s proposition with a similar one, which is then refuted in place of the original proposition.}
}}

\item[\ding{224}] {\footnotesize 
{\fontfamily{phv}\fontsize{8}{9}\selectfont
{\textbf{Doubt:}Questioning the credibility of someone or something.}
}}

\vspace{-2.2mm}
\item[\ding{224}] {\footnotesize 
{\fontfamily{phv}\fontsize{8}{9}\selectfont
{\textbf{Obfuscation:} Using words that are deliberately not clear, so that the audience may have their own interpretations.}
}}

\vspace{-2.2mm}
\item[\ding{224}] {\footnotesize 
{\fontfamily{phv}\fontsize{8}{9}\selectfont
{\textbf{Reductio ad Hitlerum:} An attempt to invalidate someone else’s argument on the basis that the same idea was promoted.}
}}

\vspace{-2.2mm}
\item[\ding{224}] {\footnotesize 
{\fontfamily{phv}\fontsize{8}{9}\selectfont
{\textbf{Black and White Fallacy:}Using words that depict the fallacy of leaping from the undesirability of one proposition to the truth of an extreme opposite.
}
}}

\vspace{-6mm}
\end{spacing}
\end{itemize}


\end{tcolorbox}
\begin{tcolorbox}[ nobeforeafter, title=box 2, colback=teal!5!white,colframe=teal!75!black,title=\footnotesize{\textbf{\textsc{Propaganda through Deception}}}]
\begin{itemize}
[leftmargin=1mm]
\setlength\itemsep{0em}
\begin{spacing}{0.90}
\vspace{-2.2mm}
\item[\ding{224}] {\footnotesize 
{\fontfamily{phv}\fontsize{8}{9}\selectfont
{\textbf{Straw Men:} Straw Men maps mostly with speculation but sometimes with opinion too. It maps with both black and white lie of layer 2 in most cases and gaining advantage in layer 3 and political in layer 4. }
}}

\vspace{-2.2mm}
\item[\ding{224}] {\footnotesize 
{\fontfamily{phv}\fontsize{8}{9}\selectfont
{\textbf{Doubt:} Doubt maps mostly with speculation in layer 1, black lie in layer 2, gaining advantage in layer 3 and political in layer 4.}
}}

\vspace{-2.2mm}
\item[\ding{224}] {\footnotesize 
{\fontfamily{phv}\fontsize{8}{9}\selectfont
{\textbf{Obfuscation:} This technique maps mostly with speculation in layer 1, red lie in layer 2, gaining advantage in layer 3 and political in layer 4. }
}}

\vspace{-2.2mm}
\item[\ding{224}] {\footnotesize 
{\fontfamily{phv}\fontsize{8}{9}\selectfont
{\textbf{Reductio ad Hitlerum:} This technique maps with speculation and distrotion in layer1, black lies and occasional white lies in layer 2. Layer 3 and layer 4 are primarily associated with gaining advantage and politics, respectively. }
}}

\vspace{-2.2mm}
\item[\ding{224}] {\footnotesize 
{\fontfamily{phv}\fontsize{8}{9}\selectfont
{\textbf{Black and White Fallacy:} This technique predominantly involves speculation and opinion, with elements of black lies in the second layer. In the third layer, it is mostly aligned with gaining advantage but occasionally tied to protecting oneself and political and educational in layer 4.
}
}}

\vspace{-6mm}
\end{spacing}
\end{itemize}

\end{tcolorbox}
\end{tcbraster}

%% file: table/appendix_circos.tex
\begin{figure*}[htbp]
    \begin{subfigure}[b]{0.50\textwidth}
    \centering
        \includegraphics[width=1.1\textwidth]{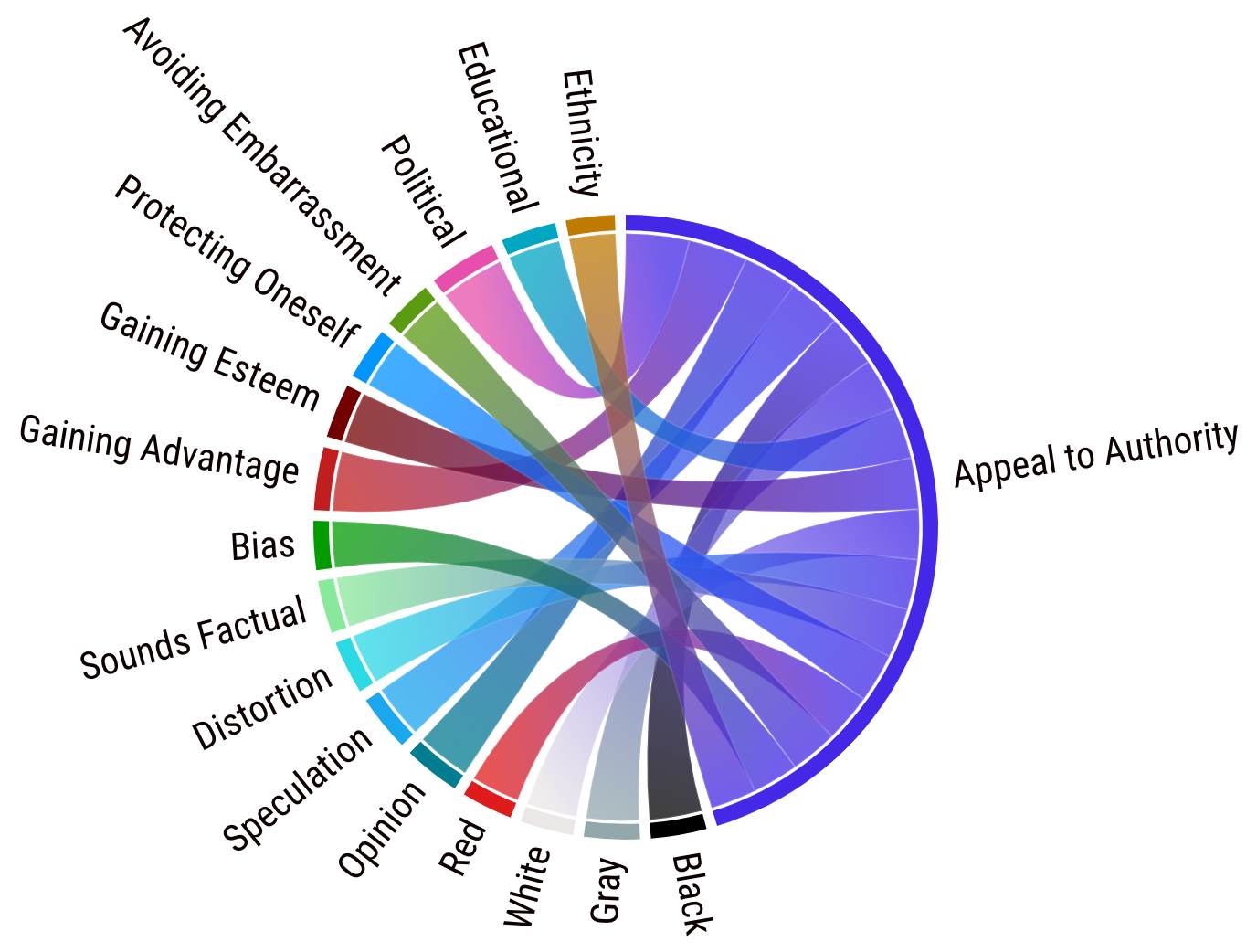}
        \caption{Layers of Deception-Appeal to Authority}
    \end{subfigure}
    \begin{subfigure}[b]{0.50\textwidth}
    \centering
        \includegraphics[width=0.8\textwidth]{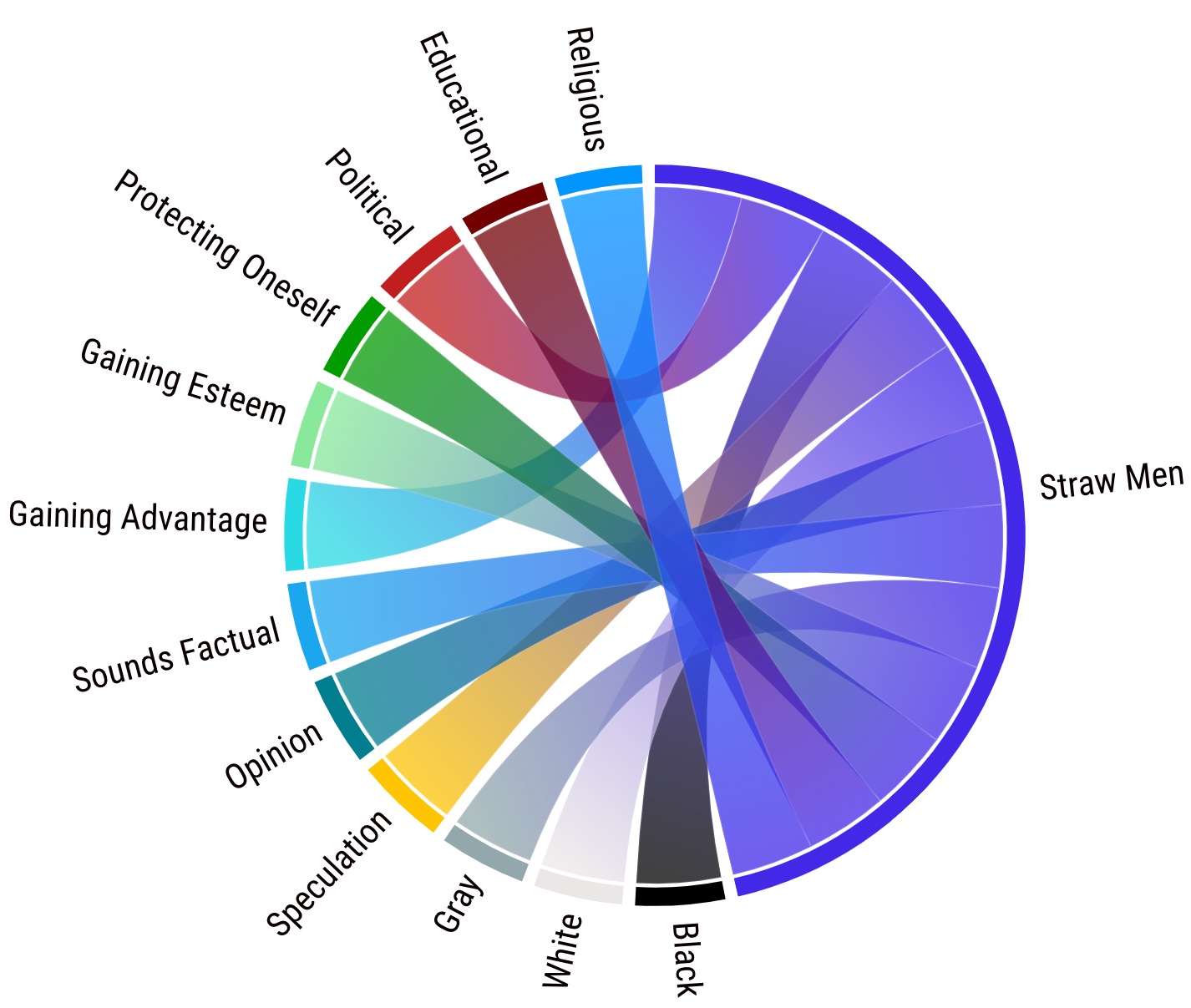}
        \caption{Layers of Deception-Straw Men}
    \end{subfigure}    
    \begin{subfigure}[b]{0.50\textwidth}
    \centering
        \includegraphics[width=0.9\textwidth]{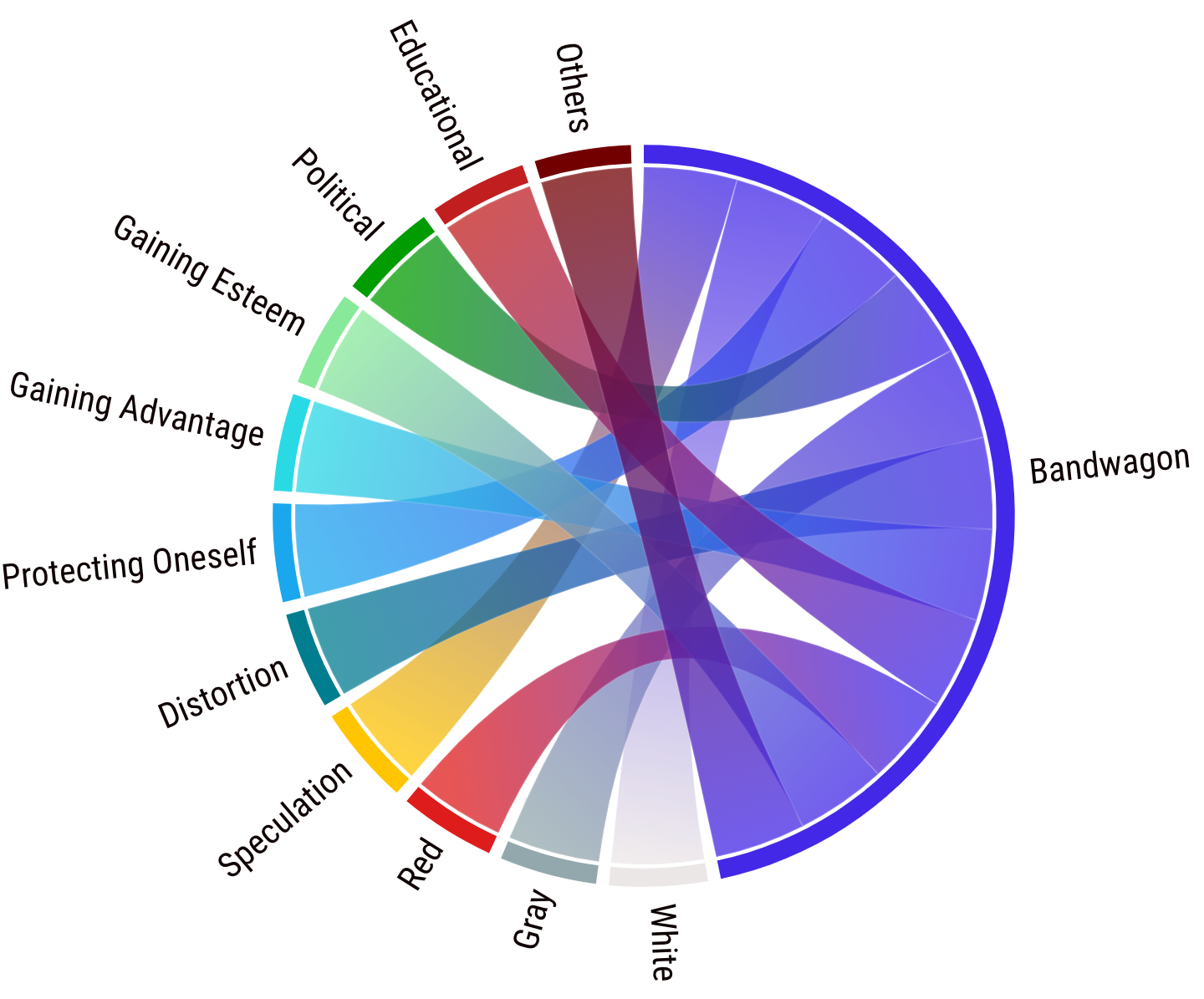}
        \caption{Layers of Deception-Bandwagon}
    \end{subfigure}
    \begin{subfigure}[b]{0.50\textwidth}
    \centering
        \includegraphics[width=0.95\textwidth]{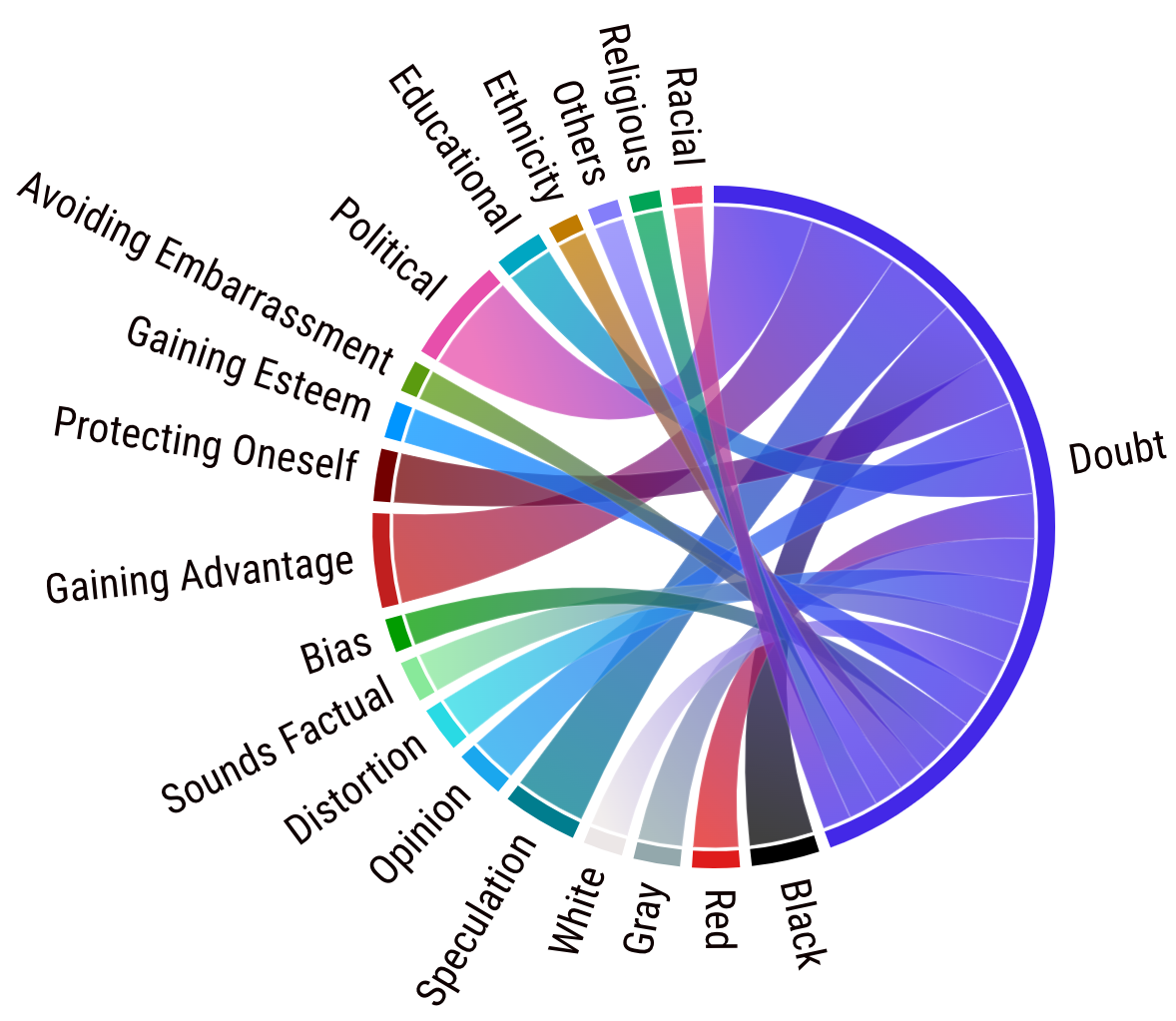}
        \caption{Layers of Deception-Doubt}
    \end{subfigure}
    \begin{subfigure}[b]{0.50\textwidth}
    \centering
        \includegraphics[width=1\textwidth]{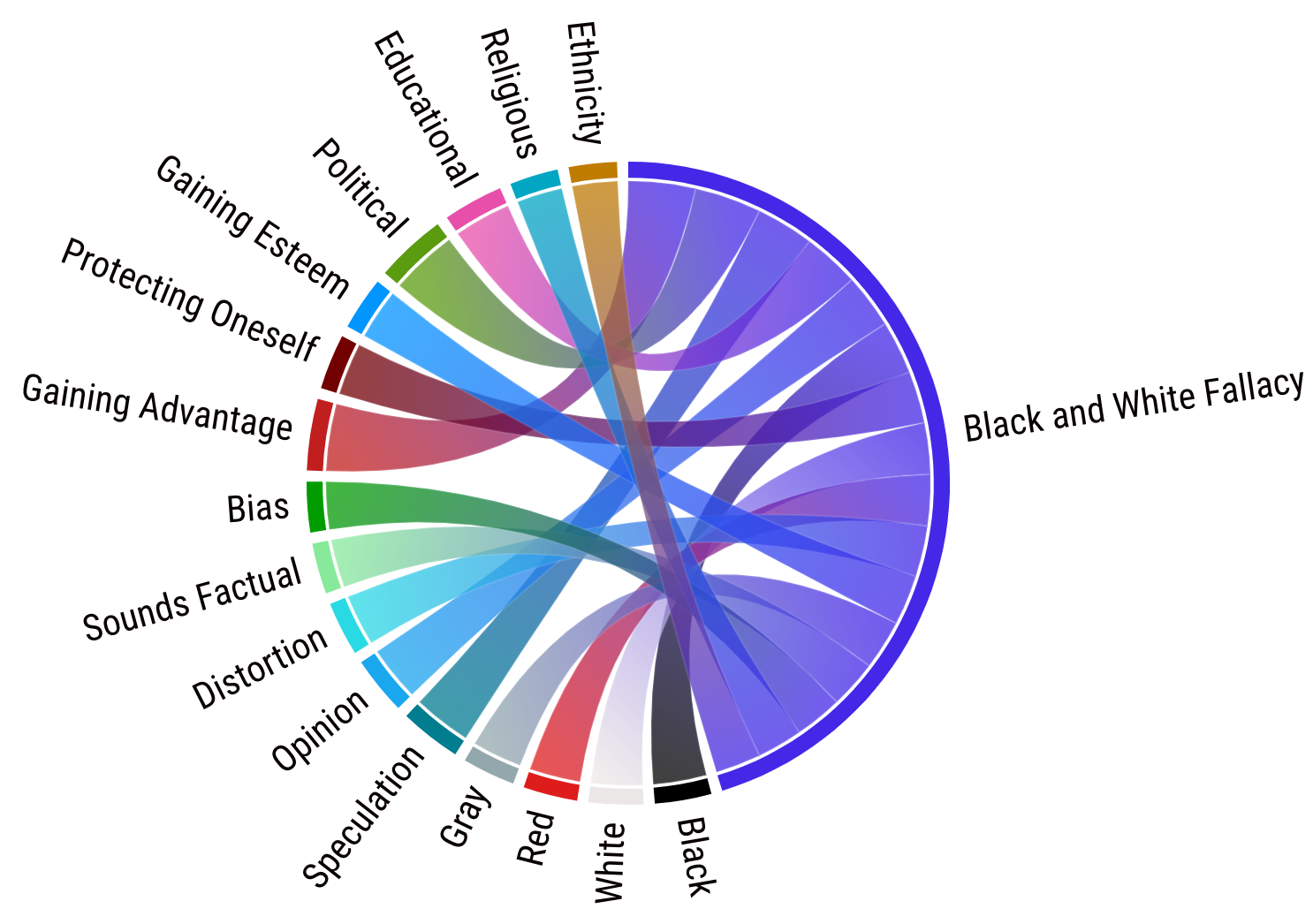}
        \caption{Layers of Deception-Black and white Fallacy}
    \end{subfigure}
    \begin{subfigure}[b]{0.50\textwidth}
    \centering
        \includegraphics[width=0.99\textwidth]{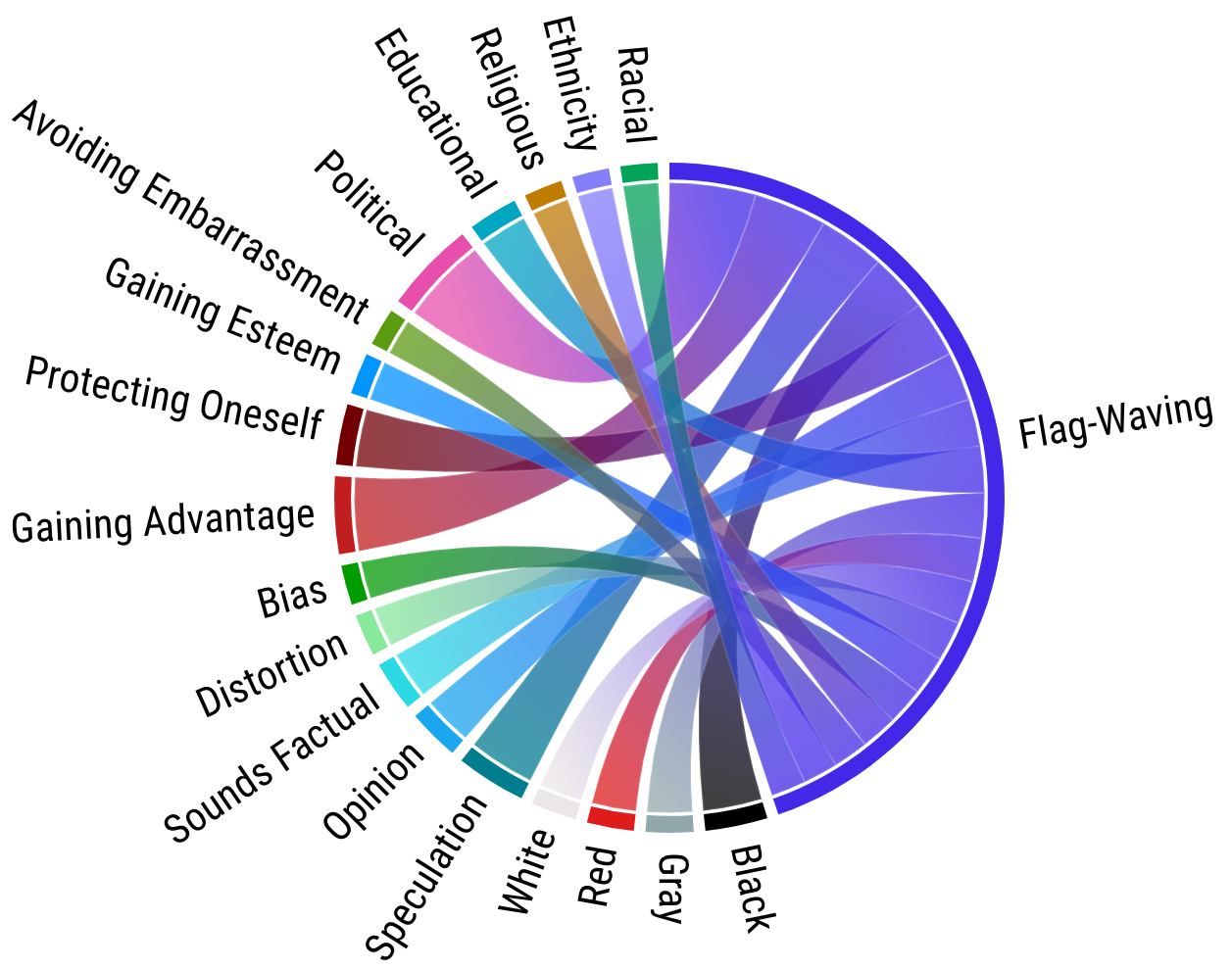}
        \caption{Layers of Deception-Flag Waving}
    \end{subfigure}
\end{figure*}    
\begin{figure*}[htbp]    
    \begin{subfigure}[b]{0.50\textwidth}
    \centering
        \includegraphics[width=0.9\textwidth]{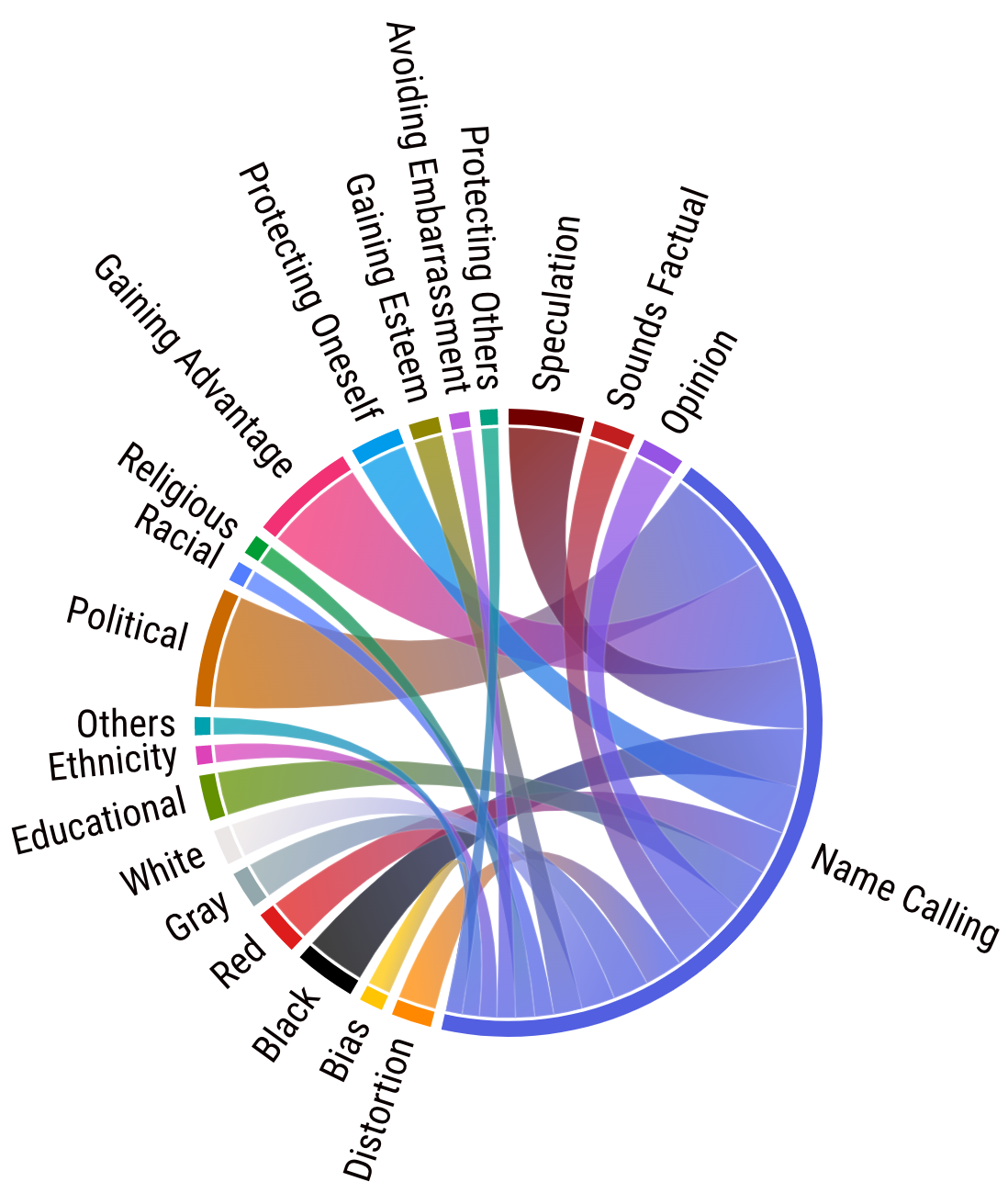}
        \caption{Layers of Deception-Name Calling}
    \end{subfigure}
    \begin{subfigure}[b]{0.50\textwidth}
    \centering
        \includegraphics[width=0.9\textwidth]{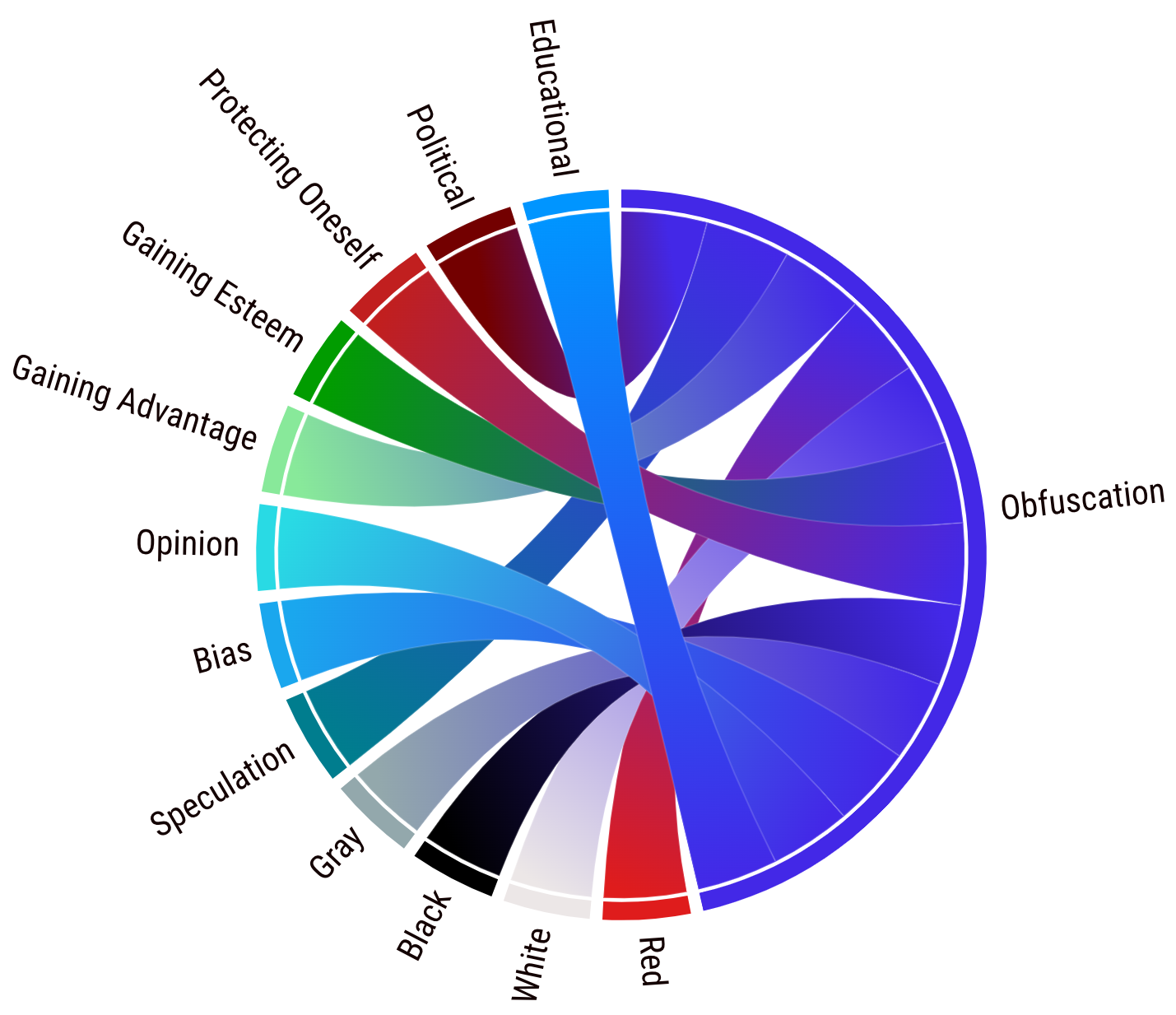}
        \caption{Layers of Deception-Obfuscation}
    \end{subfigure}
    \begin{subfigure}[b]{0.50\textwidth}
    \centering
        \includegraphics[width=0.9\textwidth]{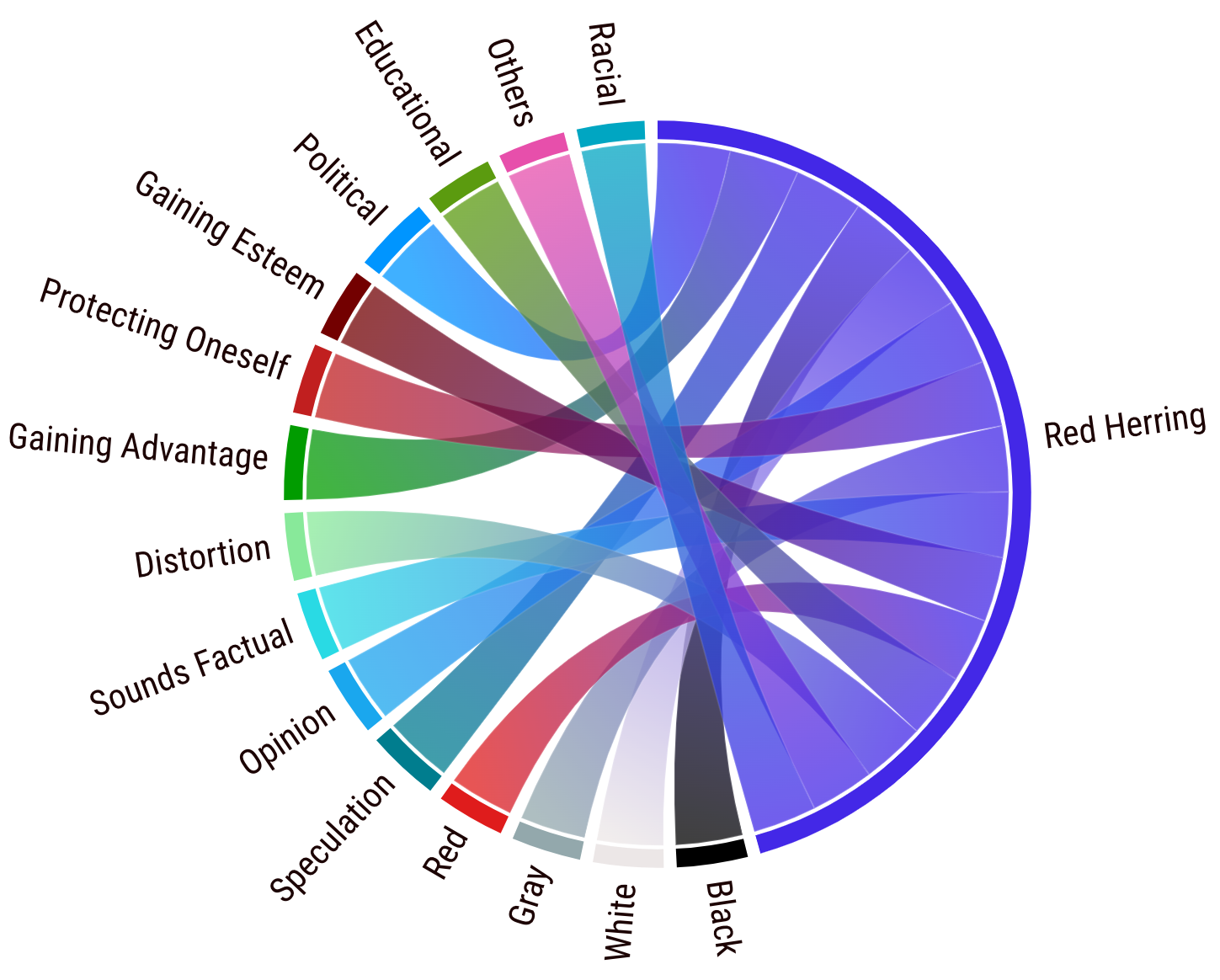}
        \caption{Layers of Deception-Red Herring}
    \end{subfigure}
    \begin{subfigure}[b]{0.50\textwidth}
    \centering
        \includegraphics[width=\textwidth]{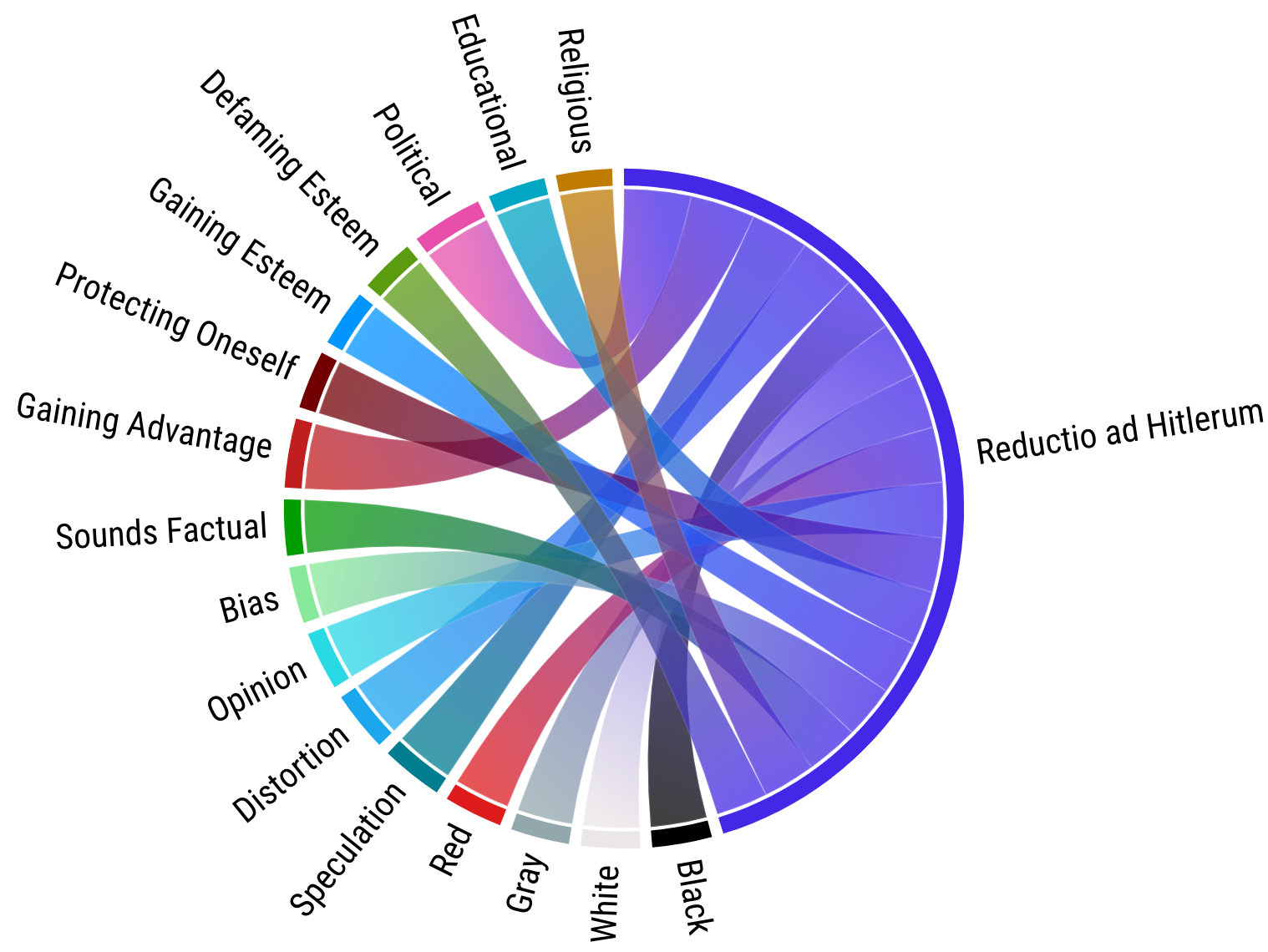}
        \caption{Layers of Deception-Reductio ad Hitlerum}
    \end{subfigure}
    \begin{subfigure}[b]{0.50\textwidth}
    \centering
        \includegraphics[width=\textwidth]{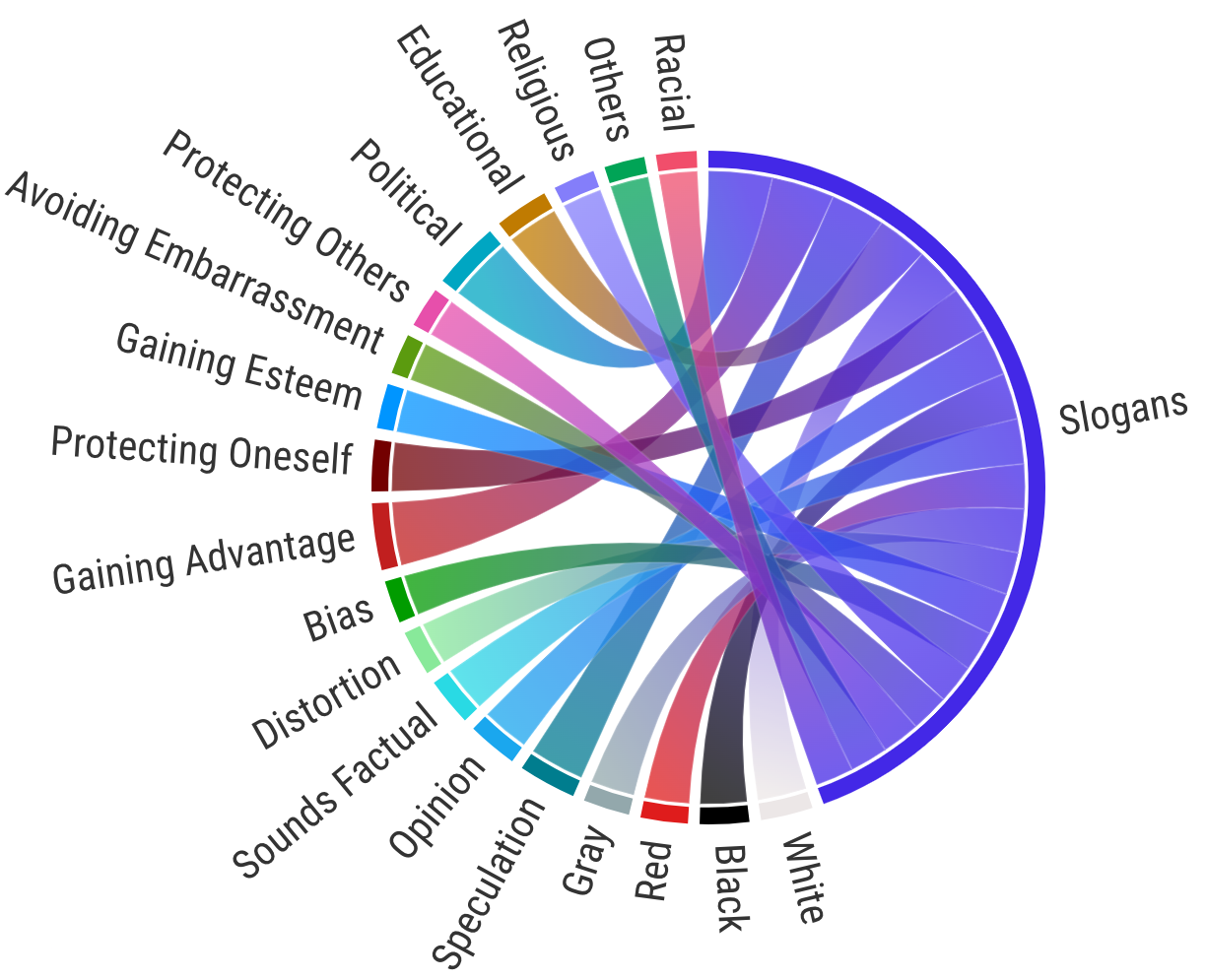}
        \caption{Layers of Deception-Slogans}
    \end{subfigure}    
    \begin{subfigure}[b]{0.50\textwidth}
    \centering
        \includegraphics[width=1.3\textwidth]{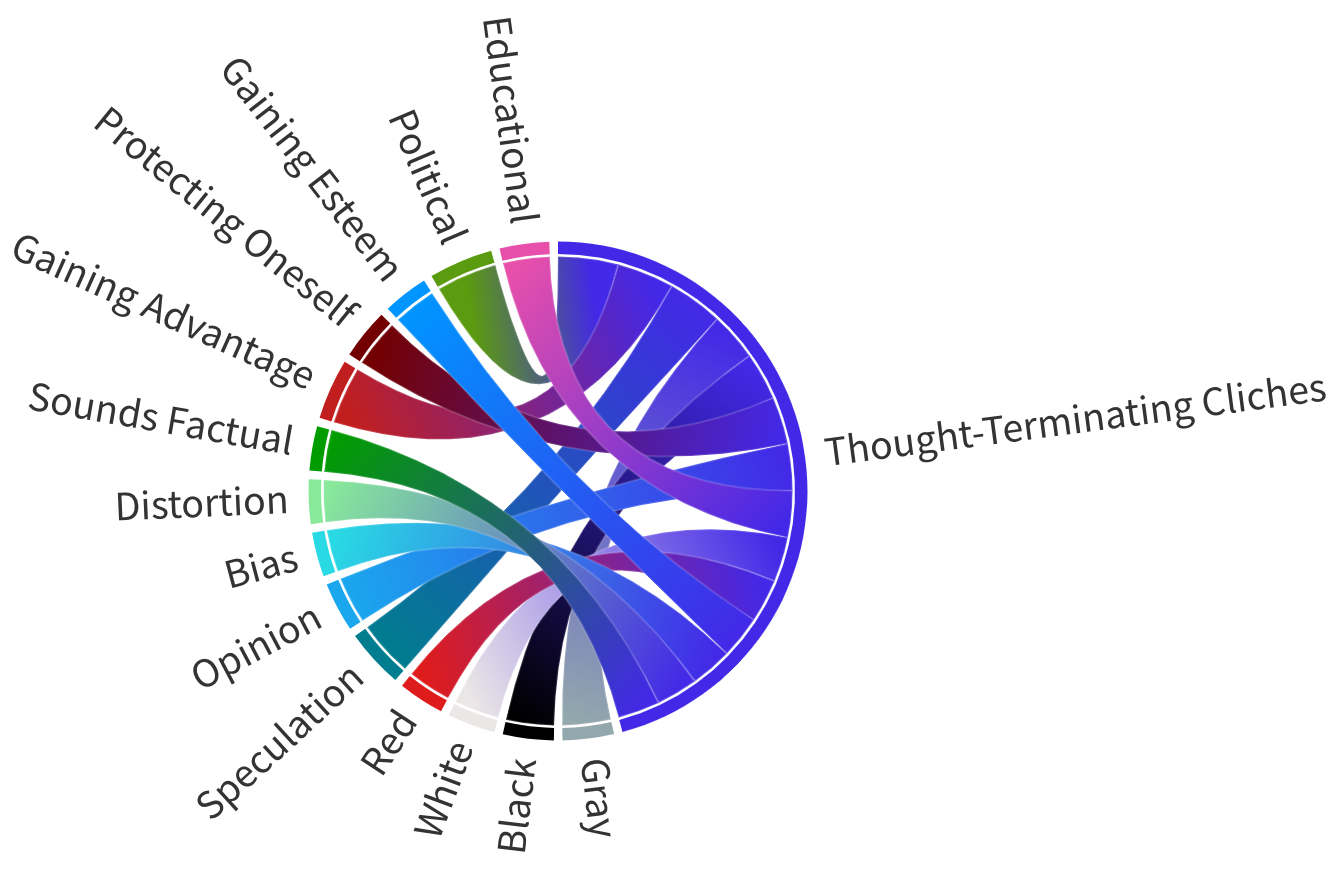}
        \caption{Layers of Deception-Thought terminating cliches}
    \end{subfigure}
\end{figure*}

\begin{figure*}[htbp]
    \begin{subfigure}[b]{0.50\textwidth}
    \centering
        \includegraphics[width=\textwidth]{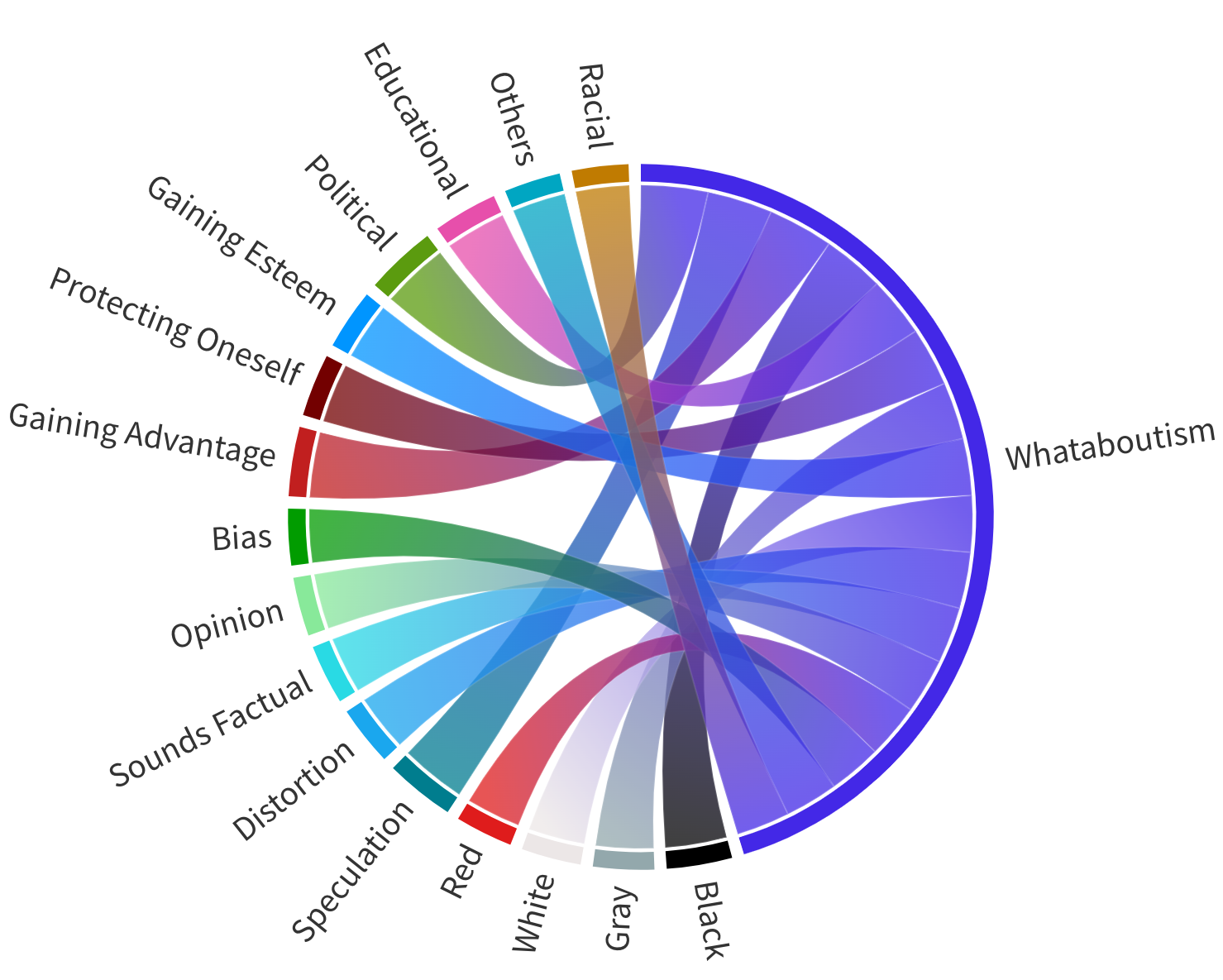}
        \caption{Layers of Deception-Whataboutism}
    \end{subfigure}
    \begin{subfigure}[b]{0.50\textwidth}
    \centering
        \includegraphics[width=\textwidth]{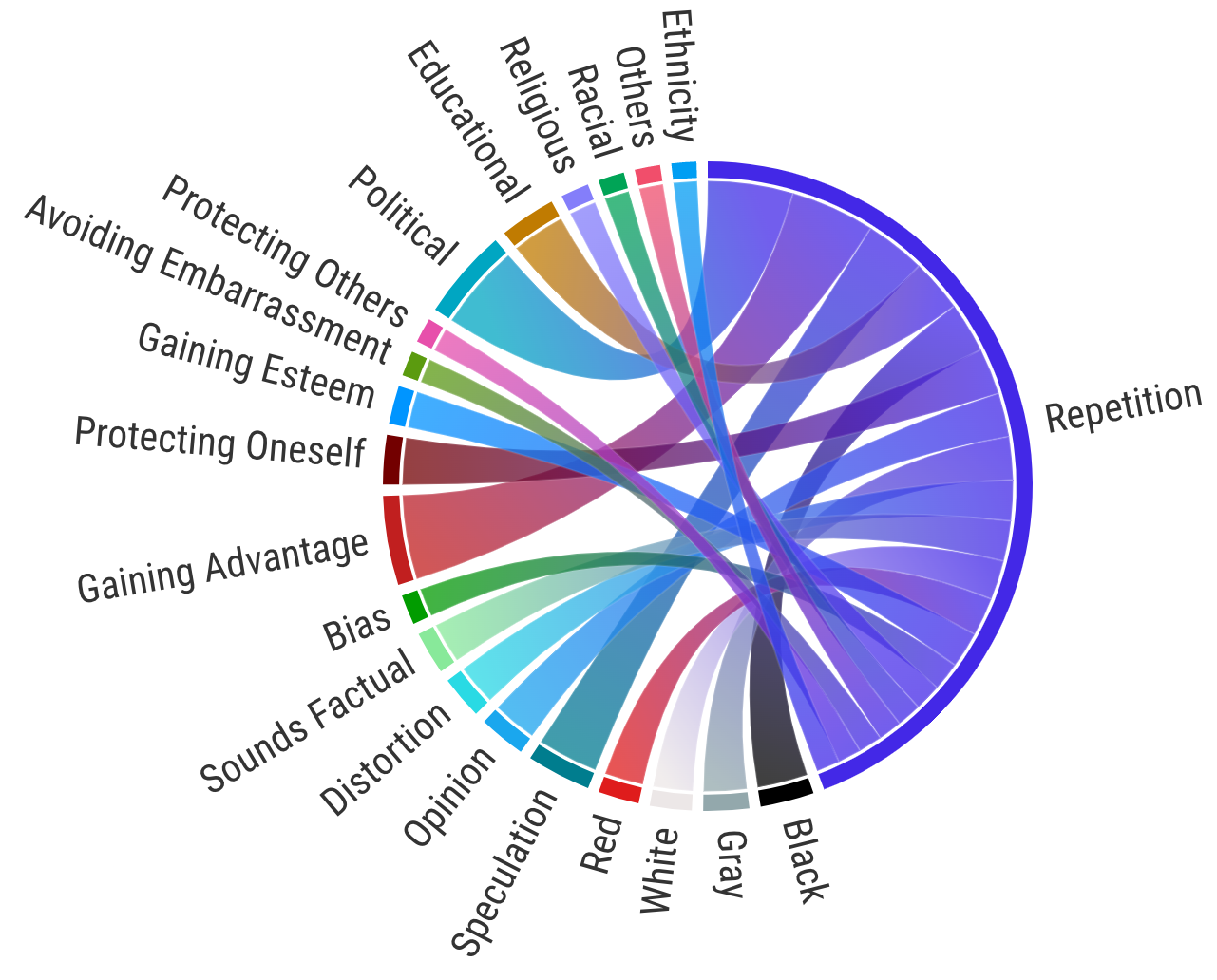}
        \caption{Layers of Deception-Repetition}
    \end{subfigure}
    \begin{subfigure}[b]{0.50\textwidth}
    \centering
        \includegraphics[width=1\textwidth]{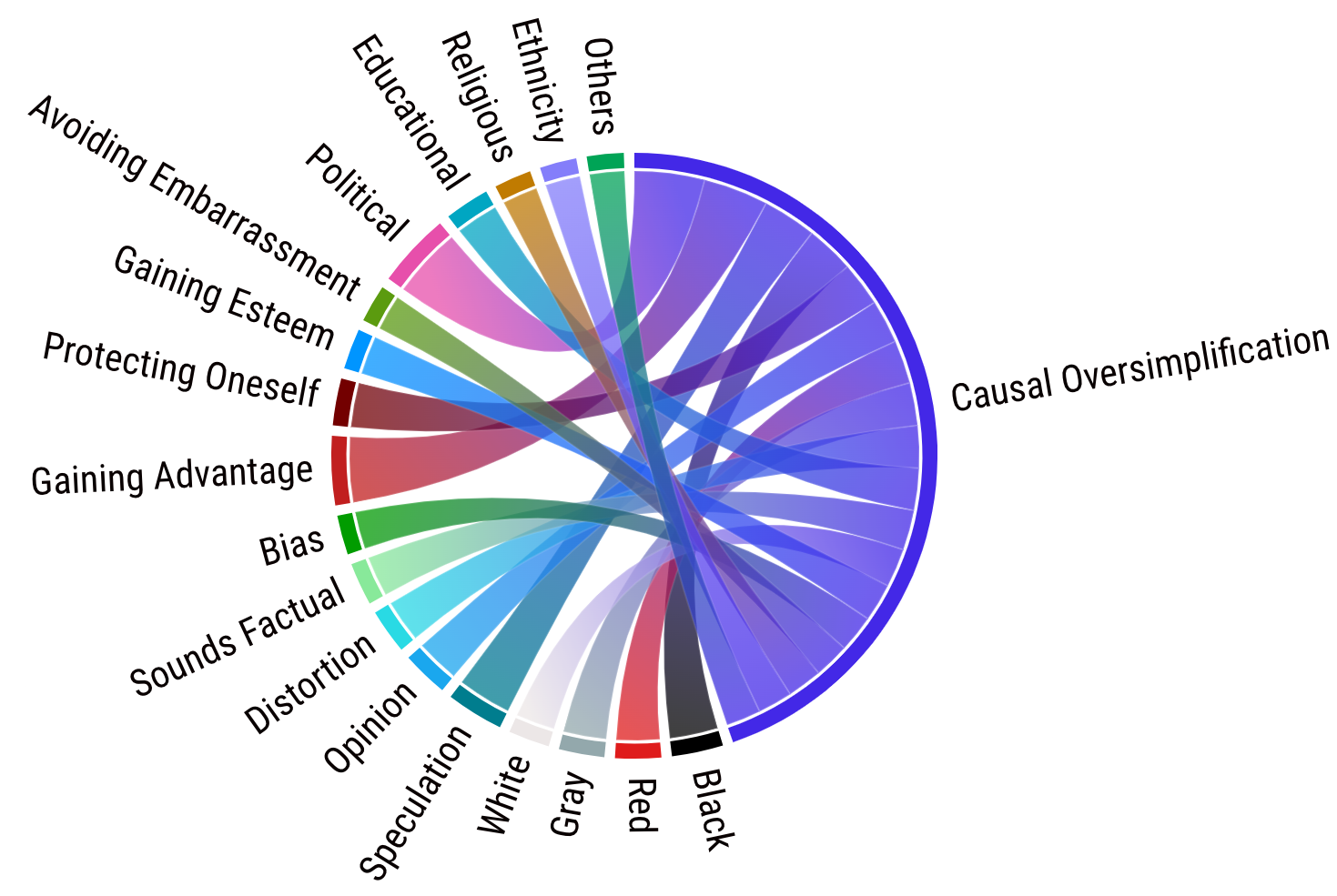}
        \caption{Layers of Deception-Casual Oversimplification}
    \end{subfigure}
    \begin{subfigure}[b]{0.50\textwidth}
    \centering
        \includegraphics[width=0.7\textwidth]{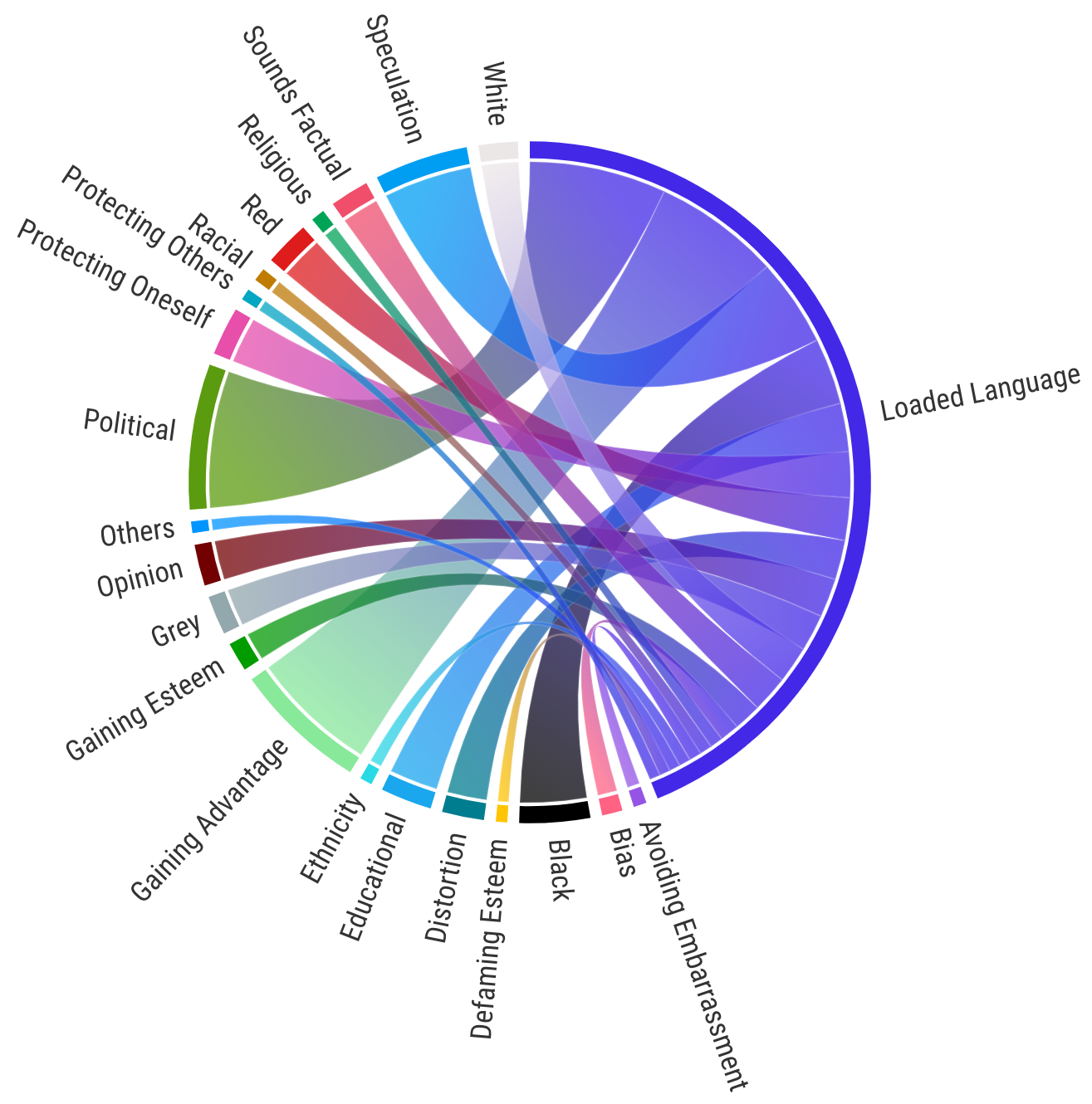}
        \caption{Layers of Deception-Loaded Language}
    \end{subfigure}
    \begin{subfigure}[b]{0.50\textwidth}
    \centering
        \includegraphics[width=0.95\textwidth]{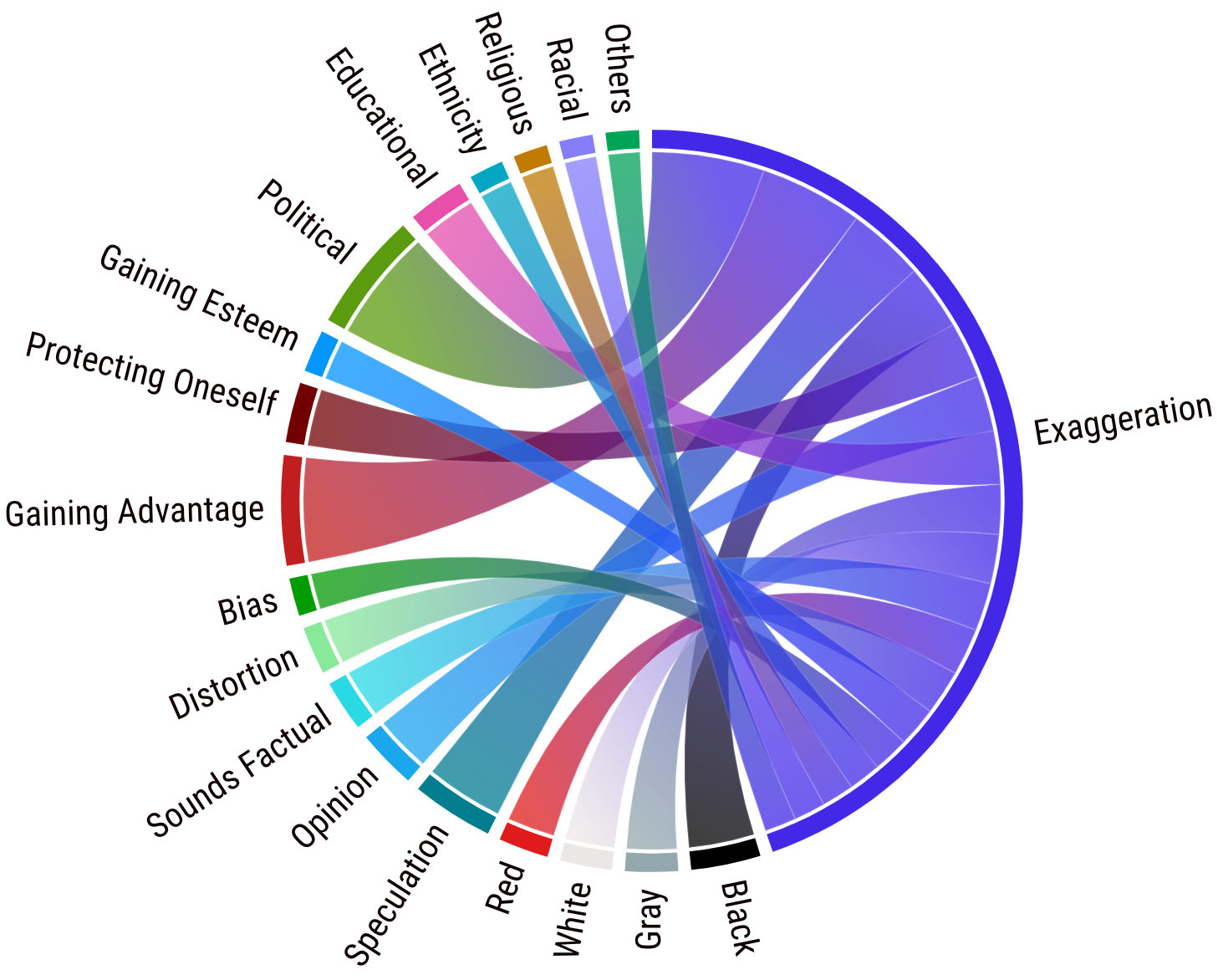}
        \caption{Layers of Deception-Exaggeration}
    \end{subfigure}
    \begin{subfigure}[b]{0.50\textwidth}
    \centering
        \includegraphics[width=1.1\textwidth]{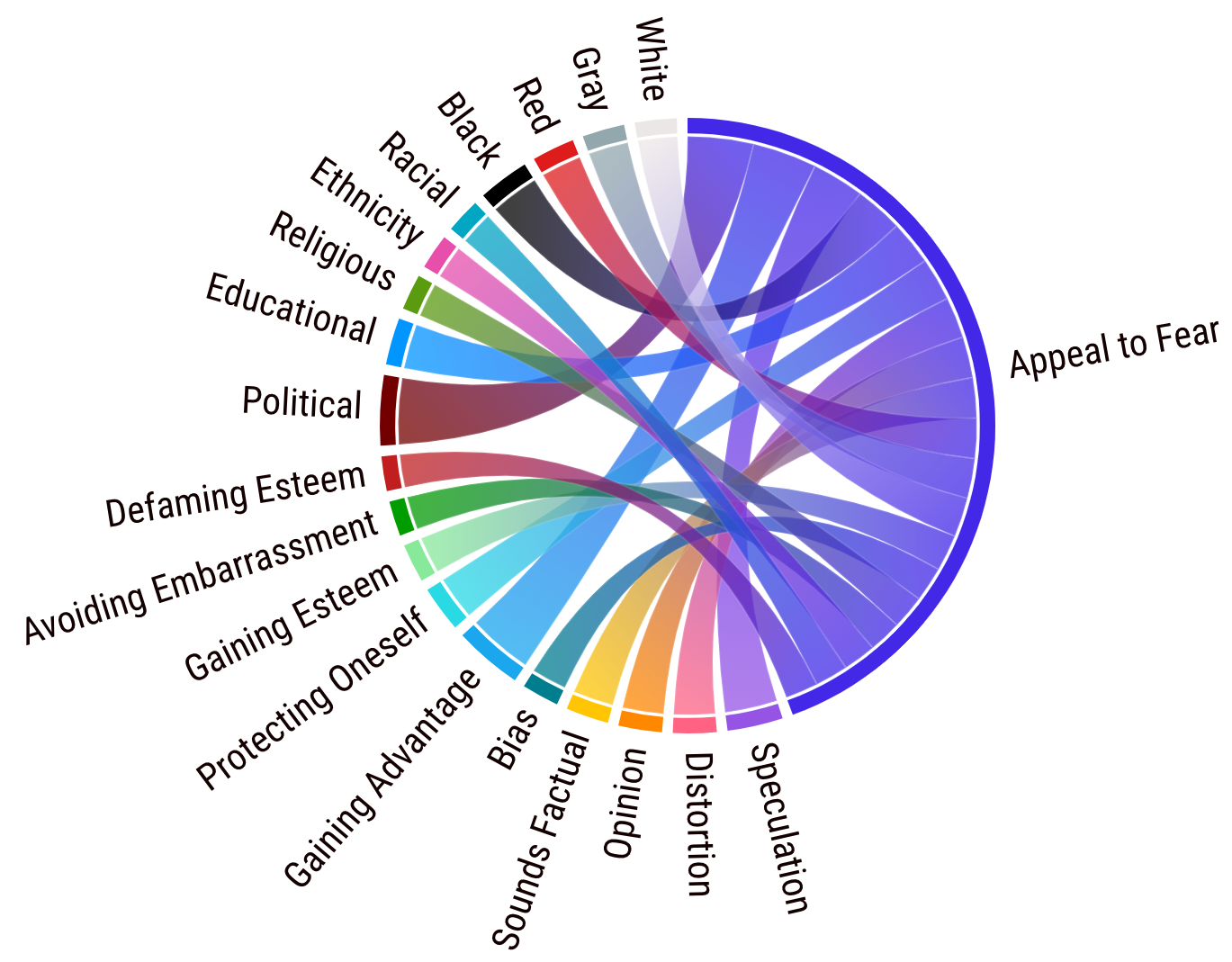}
        \caption{Layers of Deception-Appeal to fear}
    \end{subfigure}
\end{figure*}